%% file: main.tex
\documentclass[10pt,twocolumn,letterpaper]{article}

\usepackage[pagenumbers]{cvpr} 
\usepackage{listings}
\usepackage{multirow}
\input{preamble}

\usepackage{times}
\usepackage{epsfig}
\usepackage{graphicx}
\usepackage{amsmath}
\usepackage{amssymb}
\usepackage{listings}
\usepackage{enumerate}
\usepackage{pbox}
\usepackage{url}
\usepackage{mathtools}
\usepackage{amsmath}
\usepackage{amssymb}

\usepackage{blindtext}
\usepackage{algorithm}
\usepackage{algpseudocode}

\usepackage{epsfig}
\usepackage{graphicx}
\usepackage{float}
\usepackage{subcaption}
\usepackage{caption}
\usepackage{makecell}

\usepackage{epsfig}
\usepackage{graphicx}
\usepackage{subcaption}
\usepackage{enumerate}

\usepackage{booktabs}
\usepackage{pifont}
\newcommand{\cmark}{\ding{51}}%
\newcommand{\xmark}{\ding{55}}%

\usepackage{multirow}

\usepackage[misc]{ifsym}

\newcommand{\comment}[1]{}

\definecolor{cvprblue}{rgb}{0.21,0.49,0.74}
\usepackage[pagebackref,breaklinks,colorlinks,citecolor=cvprblue]{hyperref}

\newcommand{\dalle}{DALL$\cdot$E3\xspace}

\newcommand\paperurl[1]{{\footnotesize{\color{blue}{\url{#1}}}}}

\definecolor{codegreen}{rgb}{0,0.6,0}
\definecolor{codegray}{rgb}{0.5,0.5,0.5}
\definecolor{codepurple}{rgb}{0.58,0,0.82}
\definecolor{backcolour}{rgb}{0.95,0.95,0.92}

\lstdefinestyle{mystyle}{
    backgroundcolor=\color{backcolour},   
    commentstyle=\color{codegreen},
    keywordstyle=\color{magenta},
    numberstyle=\tiny\color{codegray},
    stringstyle=\color{codepurple},
    basicstyle=\sffamily\scriptsize,
    breakatwhitespace=false,         
    breaklines=true,                 
    captionpos=b,                    
    keepspaces=true,                 
    numbers=left,                    
    numbersep=5pt,                  
    showspaces=false,                
    showstringspaces=false,
    showtabs=false,                  
    tabsize=2
}

\usepackage{mathtools}

\newcommand{\fI}{\mathsf{I}}

\newcommand{\fT}{\mathsf{T}}

\newlength\savewidth\newcommand\shline{\noalign{\global\savewidth\arrayrulewidth
  \global\arrayrulewidth 1pt}\hline\noalign{\global\arrayrulewidth\savewidth}}
\newcommand{\tablestyle}[2]{\setlength{\tabcolsep}{#1}\renewcommand{\arraystretch}{#2}\centering\footnotesize}

\definecolor{cvprblue}{rgb}{0.21,0.49,0.74}

\newcommand{\redtext}[1]{\noindent \textcolor{red}{{#1}}}

\definecolor{light-light-gray}{gray}{0.90} 

\usepackage{wrapfig} 

\usepackage{amssymb}
\usepackage{pifont}

\RequirePackage[breakable,skins]{tcolorbox}
\RequirePackage{xcolor}

\newcommand\codeurl[1]{{{\color{blue}{\url{#1}}}}}

\makeatletter
\def\@fnsymbol#1{\ensuremath{\ifcase#1\or \ddagger\or \dagger\or
\mathsection\or \mathparagraph\or \|\or **\or \ddagger\ddagger
\or \dagger\dagger \else\@ctrerr\fi}}
\makeatother

\title{Glyph-ByT5: A Customized Text Encoder for Accurate Visual Text Rendering}

\author{
{\normalsize \quad Zeyu Liu$^{\dagger}{\thanks{Equal contribution. \Letter\; \texttt{yuhui.yuan@microsoft.com}}}$
 \quad Weicong Liang$^{\dagger}$ \quad Zhanhao Liang$^{\dagger}$ \quad Chong Luo \quad Ji Li \quad Gao Huang \quad Yuhui Yuan$^{\ddagger\sharp}$}\\[0mm]
{\small  $^\dagger$interns at microsoft \qquad  $^\ddagger$core contribution \qquad $^\sharp$project lead}\\[1mm]
\normalsize{Microsoft Research Asia\quad\quad Tsinghua University \quad\quad  Peking University \quad\quad The Australian National University}\\
{\footnotesize\codeurl{{https://glyph-byt5.github.io}}}\vspace{-4mm}}

\begin{document}

\twocolumn[{%
\renewcommand\twocolumn[1][]{#1}%
\maketitle
\begin{center}
\begin{minipage}[t]{1\linewidth}
\centering
\begin{minipage}{0.24\textwidth}
{\includegraphics[width=\textwidth]{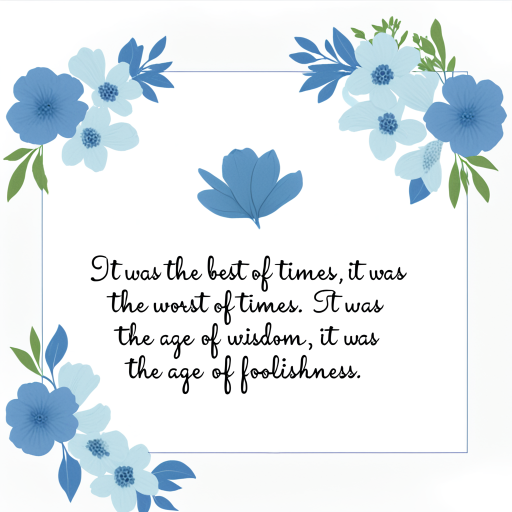}}
\vspace{-3mm}
\end{minipage}
\begin{minipage}{0.24\textwidth}
{\includegraphics[width=\textwidth]{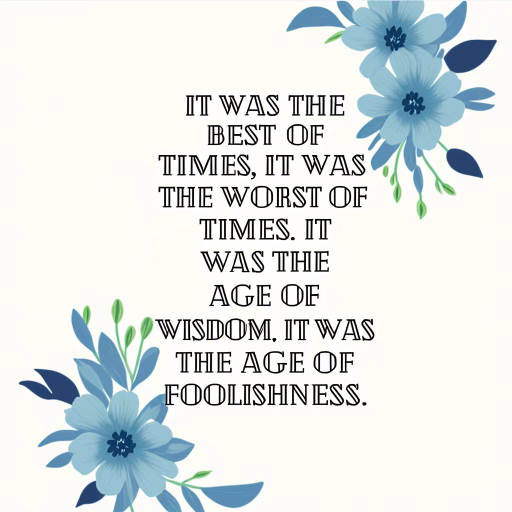}}
\vspace{-3mm}
\end{minipage}
\begin{minipage}{0.24\textwidth}
{\includegraphics[width=\textwidth]{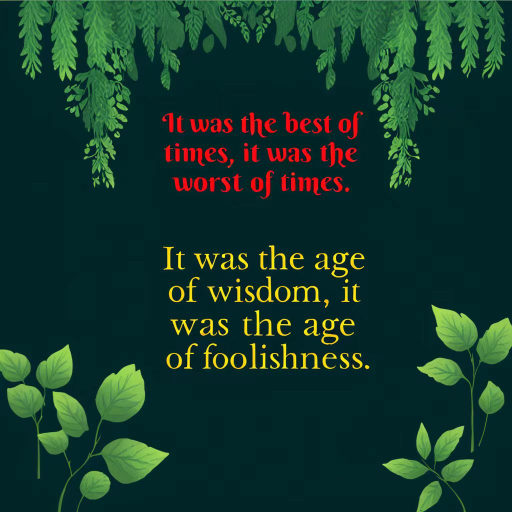}}
\vspace{-3mm}
\end{minipage}
\begin{minipage}{0.24\textwidth}
{\includegraphics[width=\textwidth]{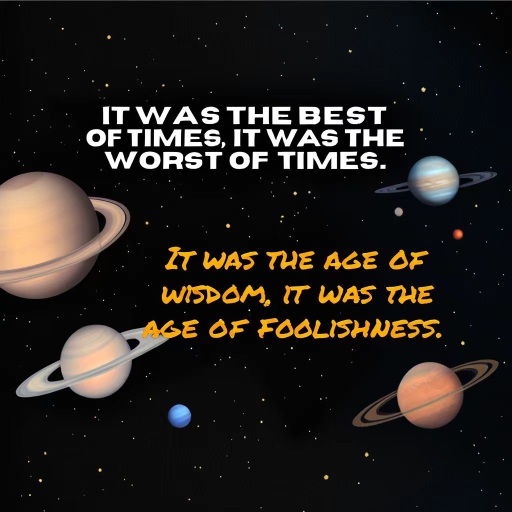}}
\vspace{-3mm}
\end{minipage}\\
\begin{minipage}{0.24\textwidth}
{\includegraphics[width=\textwidth]{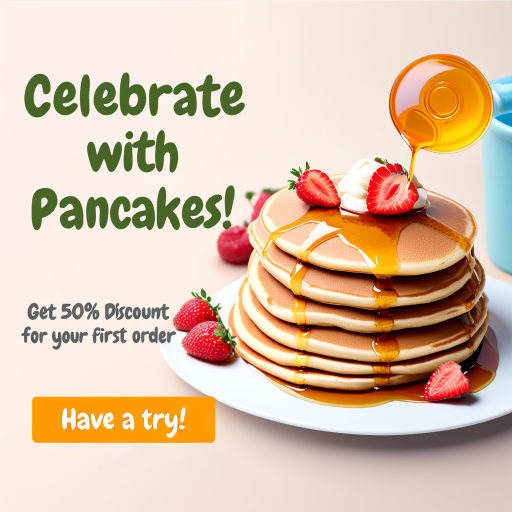}}
\vspace{-3mm}
\end{minipage}
\begin{minipage}{0.24\textwidth}
{\includegraphics[width=\textwidth]{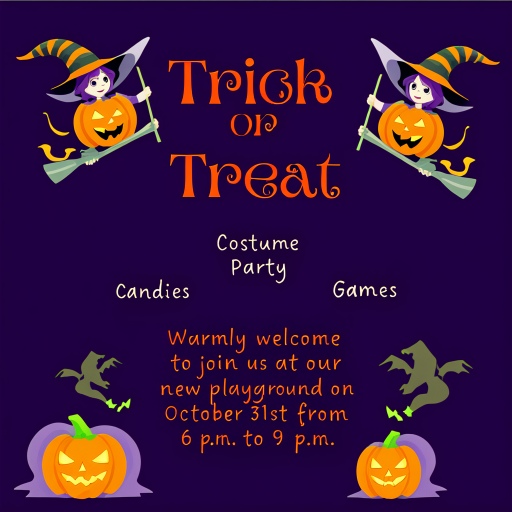}}
\vspace{-3mm}
\end{minipage}
\begin{minipage}{0.24\textwidth}
{\includegraphics[width=\textwidth]{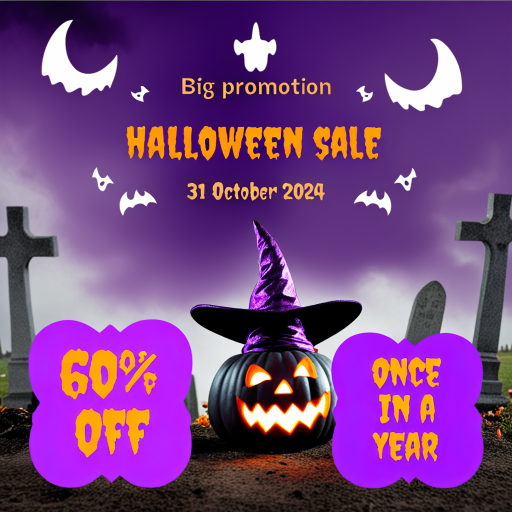}}
\vspace{-3mm}
\end{minipage}
\begin{minipage}{0.24\textwidth}
{\includegraphics[width=\textwidth]{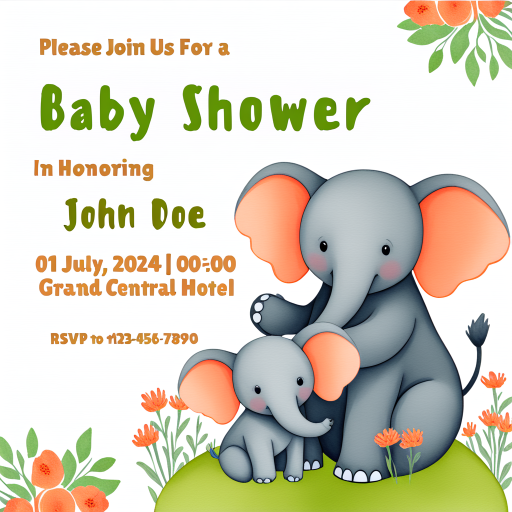}}
\vspace{-3mm}
\end{minipage}\\
\begin{minipage}{0.24\textwidth}
{\includegraphics[width=\textwidth]{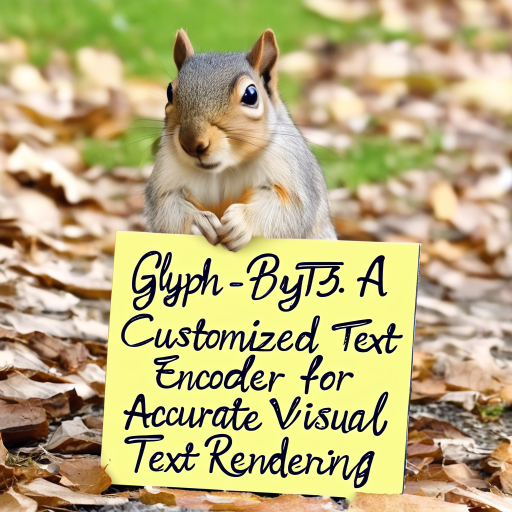}}
\vspace{-3mm}
\end{minipage}
\begin{minipage}{0.24\textwidth}
{\includegraphics[width=\textwidth]{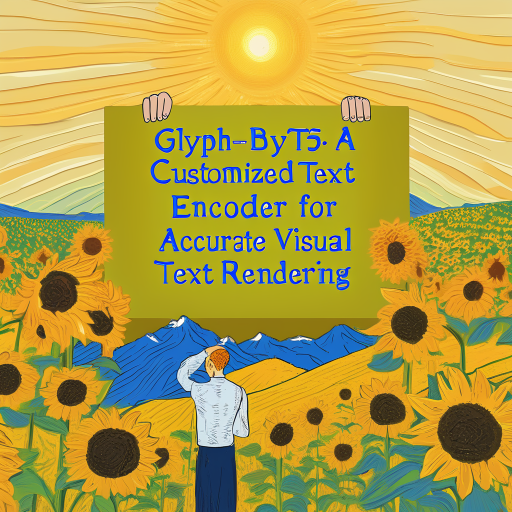}}
\vspace{-3mm}
\end{minipage}
\begin{minipage}{0.24\textwidth}
{\includegraphics[width=\textwidth]{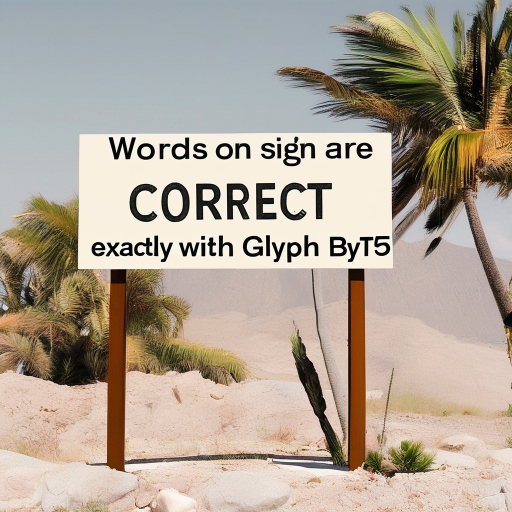}}
\vspace{-3mm}
\end{minipage}
\begin{minipage}{0.24\textwidth}
{\includegraphics[width=\textwidth]{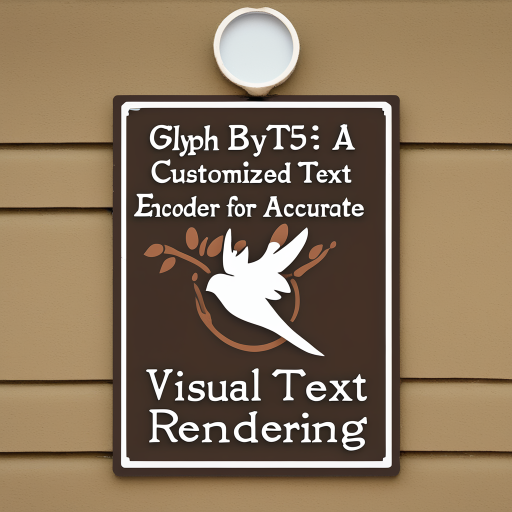}}
\vspace{-3mm}
\end{minipage}
\vspace{-3mm}
\captionof{figure}{\small{Illustrating the paragraph rendering capabilities with automatic multi-line layout planning ($1^\text{st}$ row), text-rich design images ($2^\text{nd}$ row), and open-domain images with scene text ($3^\text{rd}$ row), generated with our approach. }}
\label{fig:teaser}
\end{minipage}
\end{center}
}
]

\input{sec/0_abstract}    
\input{sec/1_intro}

\input{sec/2_related_work}
\input{sec/3_approach}
\input{sec/4_experiment}
\input{sec/5_conclusion}
{
    \small
    \bibliographystyle{ieeenat_fullname}
    \bibliography{main}
}

\input{sec/6_appendix}

\end{document}

%% file: preamble.tex
\usepackage[dvipsnames]{xcolor}


%% file: sec/0_abstract.tex
\begin{abstract}
Visual text rendering poses a fundamental challenge for contemporary text-to-image generation models, with the core problem lying in text encoder deficiencies. To achieve accurate text rendering, we identify two crucial requirements for text encoders: character awareness and alignment with glyphs. Our solution involves crafting a series of customized text encoder, Glyph-ByT5, by fine-tuning the character-aware ByT5 encoder using a meticulously curated paired glyph-text dataset. We present an effective method for integrating Glyph-ByT5 with SDXL, resulting in the creation of the Glyph-SDXL model for design image generation. This significantly enhances text rendering accuracy, improving it from less than $20\%$ to nearly $90\%$ on our design image benchmark. Noteworthy is Glyph-SDXL's newfound ability for text paragraph rendering, achieving high spelling accuracy for tens to hundreds of characters with automated multi-line layouts. Finally, through fine-tuning Glyph-SDXL with a small set of high-quality, photorealistic images featuring visual text, we showcase a substantial improvement in scene text rendering capabilities in open-domain real images. These compelling outcomes aim to encourage further exploration in designing customized text encoders for diverse and challenging tasks.
\end{abstract}

%% file: sec/1_intro.tex
\vspace{-3mm}
\section{Introduction}
\label{sec:intro}

Diffusion models have emerged as the predominant approach for image generation. Noteworthy contributions, like \dalle \cite{dalle3paper,dalle3system} and Stable Diffusion series \cite{rombach2022high,podell2023sdxl}, showcase remarkable proficiency in generating high-quality images in response to user prompts. However, a significant limitation persists in their ability to accurately render visual text, which is a critical element in various image generation applications. These applications range from producing design images for posters, cards, and brochures to synthesizing real-world images featuring scene text found in road signs, billboards, or text-laden T-shirts. The challenge of achieving precise text rendering accuracy has hindered the practical deployment of image generation models in these important domains.

We posit that the primary challenge hindering visual text rendering performance lies in the limitations of text encoders. The widely used CLIP text encoder, trained to align with visual signals, primarily focuses on grasping image concepts rather than delving into image details. Conversely, the commonly adopted T5 text encoder, designed for a comprehensive understanding of language, lacks alignment with visual signals. We argue that a text encoder capable of encoding character-level information and aligning with visual text signals, or glyphs, is essential for achieving high accuracy in visual text rendering. Drawing inspiration from the character-aware ByT5 encoder~\cite{Liu2022CharacterAwareMI}, our approach aims to customize it to better align with visual text or glyphs.

To construct the desired character-aware and glyph-aligned text encoder, we employ a fine-tuning approach based on the ByT5 model using paired text-glyph data. The main challenge arises from the scarcity of high-quality paired text-glyph data, which we overcome by establishing a scalable pipeline capable of generating virtually unlimited paired data based on graphic rendering. Additionally, we incorporate a glyph augmentation strategy to enhance the character awareness of the text encoder, addressing various error types commonly encountered in visual text rendering, as discussed in~\cite{Liu2022CharacterAwareMI}. Leveraging our meticulously crafted dataset and employing an innovative box-level contrastive loss, we efficiently fine-tune ByT5 into a series of customized text encoder for glyph generation, named Glyph-ByT5.

\begin{table}[!t]
\begin{minipage}[t]{1\linewidth}
\centering
\tablestyle{1pt}{1.5}
\resizebox{1.0\linewidth}{!}
{
\begin{tabular}{l|c|c|c|cccc}
\multirow{2}{*}{Method} & \multirow{2}{*}{\#Params} & \multirow{2}{*}{Char-aware}  & \multirow{2}{*}{Glyph-align} & \multicolumn{4}{c}{Precision ($\%$)} \\\cline{5-8}
& & & & $\le$20 chars & $\le$20-50 chars & $\le$50-100 chars & $\ge$100 chars  \\
\shline
$\textrm{SDXL}_{\;\small{\text{(CLIP \& OpenCLIP)}}}$ & $817$M & \xmark  & \xmark  & $21.72$ & $20.98$ & $18.23$ & $19.17$ \\
+ T5-L & + $394$M & \xmark  & \xmark & $48.46$ & $44.89$ & $34.59$ & $26.09$ \\
+ ByT5-S  & + $292$M & \cmark & \xmark  & $60.52$ & $52.79$ & $50.11$  & $42.05$ \\
+ Glyph-ByT5-S & + $292$M & \cmark  & \cmark & $92.58$  & $90.38$  & $87.16$  & $83.17$ \\
+ $\textrm{Glyph-ByT5-S}^{1\textrm{M}}$ & + $292$M & \cmark  & \cmark  & $\bf{93.89}$ & $\bf{93.67}$ & $\bf{91.45}$ & $\bf{89.17}$ \\\hline
$\textrm{DeepFloyd-IF}_{\;\small{\text{(T5-XXL)}}}$ & $4.3$B & \xmark  & \xmark  & $17.63$   & $17.17$  & $16.42$  & $13.05$ \\
\dalle & Unknown & \xmark  & \xmark   & $23.23$  & $21.59$ & $20.1$ & $15.81$  \\
\end{tabular}
}
\vspace{2mm}
\caption{
\small{Illustrating the improved results achieved with our approach based on SDXL across a varying number of characters, we choose the encoder of T5-Large and ByT5-Small for a relatively fair comparison. We only display the number of parameters for the text encoder components in the second column. Performance is demonstrated through evaluating the word-level precision of each model on different text length ranges. \textit{Char-aware}: using character-aware text encoder. \textit{Glyph-align}: glyph-alignment pre-training. We also report the performance of DeepFloyd-IF and \dalle in our benchmark, which comprises 1,000 prompts, with 250 prompts within each range of character numbers. By default, we compute all precision scores at the word level. The superscript `1M' indicates the use of 1 million training pairs, whereas the preceding four rows use 500K by default.}}
\label{tab:teaser_table}
\end{minipage}
\end{table}

Upon thorough training, Glyph-ByT5 is seamlessly integrated into the SDXL model using an efficient region-wise cross-attention mechanism, significantly enhancing the text rendering performance of the original diffusion model. The resultant Glyph-SDXL model showcases exceptional spelling accuracy, outperforming other state-of-the-art models in the generation of text-rich design images, as illustrated in Table~\ref{tab:teaser_table}. Furthermore, we fine-tuned Glyph-SDXL using a limited set of scene-text images, significantly bolstering its proficiency in generating scene-text images. The examples featured in Fig. \ref{fig:teaser} demonstrate that the refined model adeptly renders text paragraphs as scene text without perceptible degradation in the image generation capabilities of the original model. 

Our investigation reveals that, through the training of a customized text encoder and the implementation of a suitable information injection mechanism, we can transform an open-domain image generator into an outstanding visual text renderer. When presented with a textual paragraph ranging from tens to hundreds of characters, our fine-tuned diffusion model achieves high spelling accuracy for rendering within the designated region, with fully automated handling of multi-line layouts. In essence, this work contributes in three distinct yet complementary ways. First, we train a character-aware, glyph-aligned text encoder, Glyph-ByT5, as the key solution to the accurate visual text rendering problem. Second, we elaborate on the architecture and training of Glyph-SDXL, a robust design image generator that integrates Glyph-ByT5 into SDXL through an efficient region-wise cross-attention mechanism. Lastly, we showcase the potential of fine-tuning Glyph-SDXL into a scene-text image generator, laying the groundwork for the development of a comprehensive, open-domain image generator equipped with exceptional visual text rendering capabilities.

%% file: sec/2_related_work.tex
\section{Related Work}
\label{sec:related_work}

\subsection{Visual Text Rendering}
Rendering legible and visually coherent text poses a well-known limitation and a significant challenge for diffusion-based image generation models. It is worth noting that certain contemporary open-domain image generation models, such as Stable Diffusion 3~\cite{esser2024scaling} and Ideogram 1.0\footnote{\url{https://about.ideogram.ai/1.0}}, have dedicated considerable effort to enhance visual text rendering performance. However, the spelling accuracy of the rendered text remains unsatisfactory.
Conversely, there have been endeavors focused on visual text rendering, such as GlyphControl, GlyphDraw, and the TextDiffuser series \cite{yang2023glyphcontrol,ma2023glyphdraw,Liu2022CharacterAwareMI,chen2023textdiffuser,chen2023textdiffuser2}. While these efforts have shown substantial improvements in spelling accuracy, it is disappointing to note that they are still focusing on rendering single words or text lines with fewer than approximately 20 characters.
In this study, we aim to tackle the precise visual text rendering problem, particularly when dealing with textual content longer than a hundred characters, setting forth an ambitious goal in this domain.

\subsection{Customized Text Encoder}
Several recent efforts~\cite{ji2023improving,chendiffute,zhao2023udifftext} have been made to train text-oriented diffusion models and replace or augment the original CLIP encoders with customized text encoders in different manners. 
However, these methods, like their predecessors, are limited to handling text sequences of a certain length, with UDiffText~\cite{zhao2023udifftext} supporting sequences of no more than 12 characters. In contrast, our methodology distinguishes itself by its ability to generate text sequences of more than 100 characters while achieving exceptionally high accuracy, reaching nearly $90\%$ word-level accuracy. This significant progress addresses the shortcomings of previous methods, providing wider applicability and improved performance in text generation tasks. Another closely related work is Counting-aware CLIP~\cite{paiss2023teaching}, which enhances the original CLIP text encoder with a specialized image-text counting dataset and a counting-focused loss function. However, a significant limitation of their approach is the lack of scalability in their dataset. They choose to replace the original text encoders and train diffusion models from scratch, whereas our data construction pipeline is scalable, and we prioritize integrating GlyphByT5 with the original text encoders to improve efficiency.

\vspace{2mm}
\noindent\textbf{Our Contribution} Our work aligns with the insights of the previously mentioned studies, identifying that one critical limitation in most current text-to-image generation models resides in the text encoder. The primary contribution of our work lies in presenting an effective strategy for systematically addressing the glyph rendering task. We first demonstrate that leveraging graphic rendering to create scalable and accurate glyph-text data is crucial for training a high-quality, glyph-aligned, character-aware text encoder. Then, we introduce a simple yet powerful method to integrate our Glyph-ByT5 text encoder with the original CLIP text encoder used in SDXL. Additionally, we illustrate how our approach can be applied to scene-text generation by performing design-to-scene alignment fine-tuning.
We anticipate that training the customized text encoder on scalable, high-quality data represents a promising avenue for overcoming fundamental limitations, such as spatial awareness and numeracy.

%% file: sec/3_approach.tex
\section{Our Approach}
\label{sec:approach}

We begin by illustrating the details of our customized glyph-aligned, character-aware text encoder, Glyph-ByT5, which is trained using a substantial dataset of paired glyph images and textual instructions. Subsequently, we demonstrate how Glyph-ByT5 significantly enhances the visual text rendering accuracy when integrated with the SDXL models for the design-text rendering task. Finally, we introduce a straightforward yet effective approach for design-to-scene alignment, enabling the adaptation of Glyph-SDXL for precise scene-text generation.

\subsection{Glyph-ByT5: Customized Glyph-Aligned Character-Aware Text Encoder for Design-text Generation}
A key factor contributing to inaccuracies in text rendering is the inherent limitations of text encoders in modern diffusion models, especially regarding their interpretation of glyph images. The original CLIP text encoder, for example, is tailored for broad visual-language semantic alignment at the conceptual level, while the T5/ByT5 text encoder focuses on deep language understanding. However, neither is explicitly fine-tuned for glyph image interpretation although the recent works show that T5/ByT5 text encoder is favorable for visual text rendering task. This lack of customized text encoder design can result in less accurate text rendering in various applications.

\begin{figure*}[t]
\begin{minipage}[t]{1\linewidth}
\centering
\begin{subfigure}[b]{0.115\textwidth}
\includegraphics[width=\textwidth]{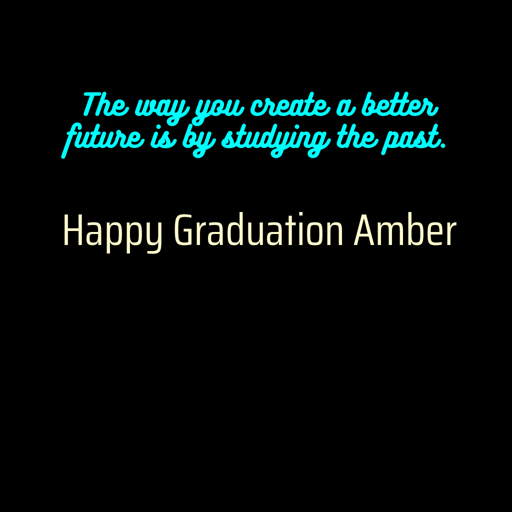}
\vspace{-3mm}
\caption{\scriptsize{}}
\end{subfigure}
\begin{subfigure}[b]{0.115\textwidth}
\includegraphics[width=\textwidth]{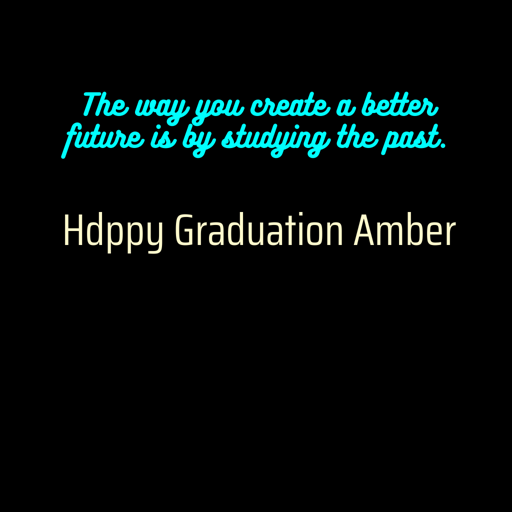}
\vspace{-3mm}
\caption{\scriptsize{}}
\end{subfigure}
\begin{subfigure}[b]{0.115\textwidth}
\includegraphics[width=\textwidth]{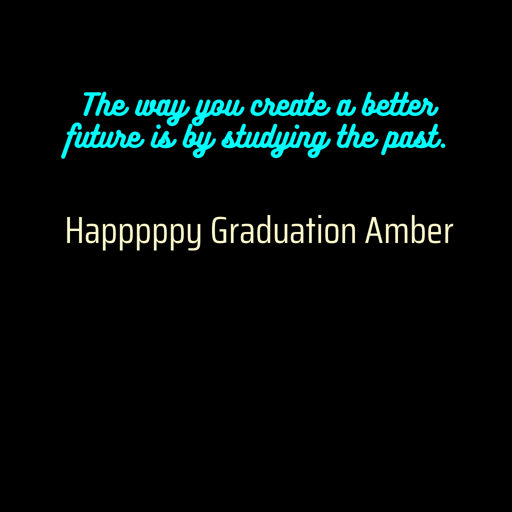}
\vspace{-3mm}
\caption{\scriptsize{}}
\end{subfigure}
\begin{subfigure}[b]{0.115\textwidth}
\includegraphics[width=\textwidth]{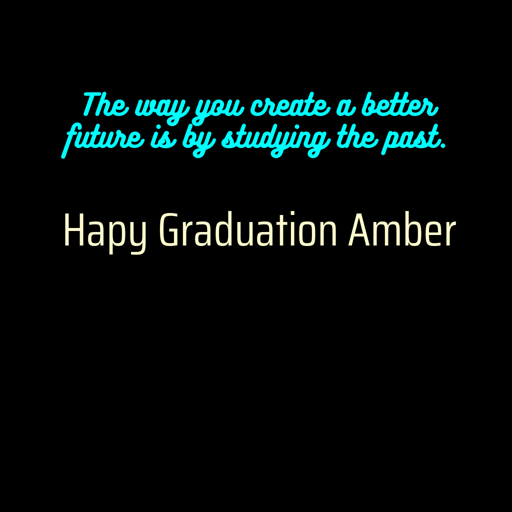}
\vspace{-3mm}
\caption{\scriptsize{}}
\end{subfigure}
\begin{subfigure}[b]{0.115\textwidth}
\includegraphics[width=\textwidth]{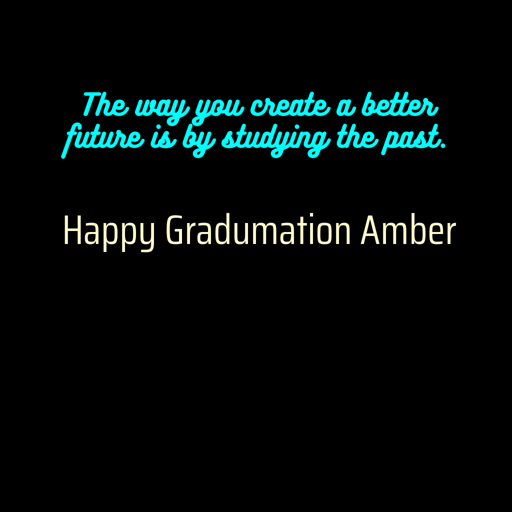}
\vspace{-3mm}
\caption{\scriptsize{}}
\end{subfigure}
\begin{subfigure}[b]{0.115\textwidth}
\includegraphics[width=\textwidth]{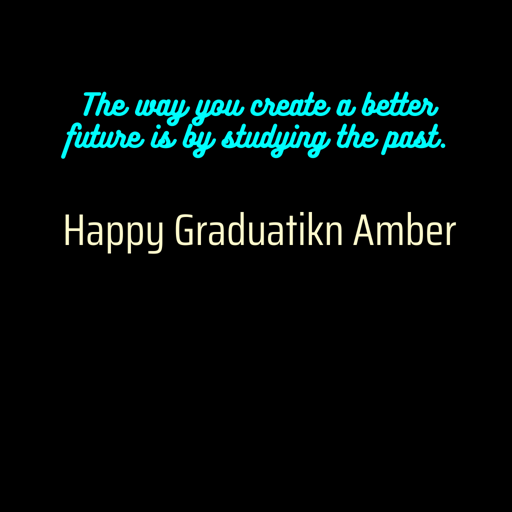}
\vspace{-3mm}
\caption{\scriptsize{}}
\end{subfigure}
\begin{subfigure}[b]{0.115\textwidth}
\includegraphics[width=\textwidth]{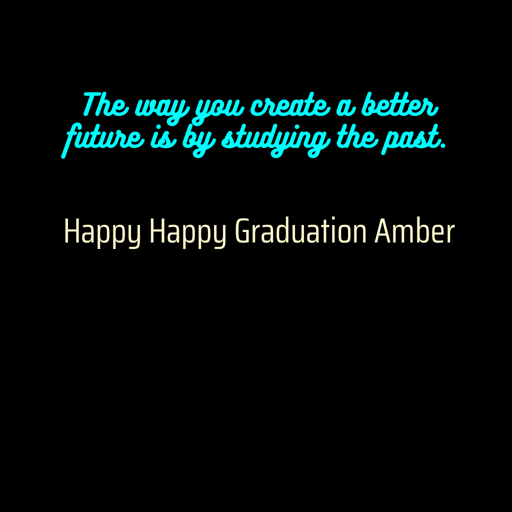}
\vspace{-3mm}
\caption{\scriptsize{}}
\end{subfigure}
\begin{subfigure}[b]{0.115\textwidth}
\includegraphics[width=\textwidth]{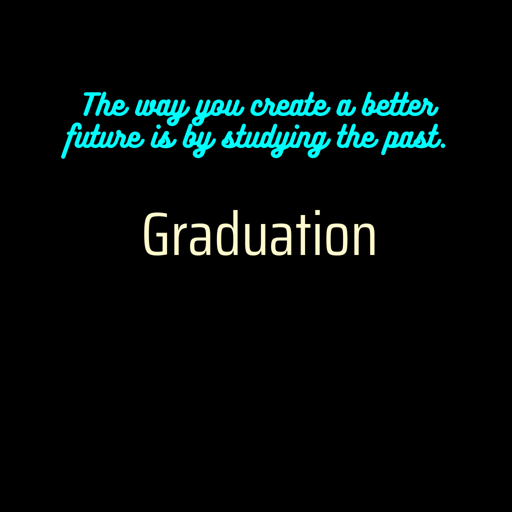}
\vspace{-3mm}
\caption{\scriptsize{}}
\end{subfigure}
\caption{\small{Illustrating the scheme of glyph augmentation. (a) original glyph. (b) character replacement (Happy $\to$ Hdppy). (c) character repeat (Happy $\to$ Happpppy). (d) character drop (Happy $\to$ Hapy). (e) character add (Graduation $\to$ Gradumation). (f) word replacement (Graduation $\to$ Gauatikn). (g) word repeat (Happy $\to$ Happy Happy). (h) word drop (Happy Graduation Amber $\to$ Graduation).\\}}
\label{fig:glyph_aug}
\end{minipage}
\vspace{-5mm}
\begin{minipage}[t]{1\linewidth}
\vspace{3mm}
\centering
\begin{subfigure}[b]{0.115\textwidth}
\includegraphics[width=\textwidth]{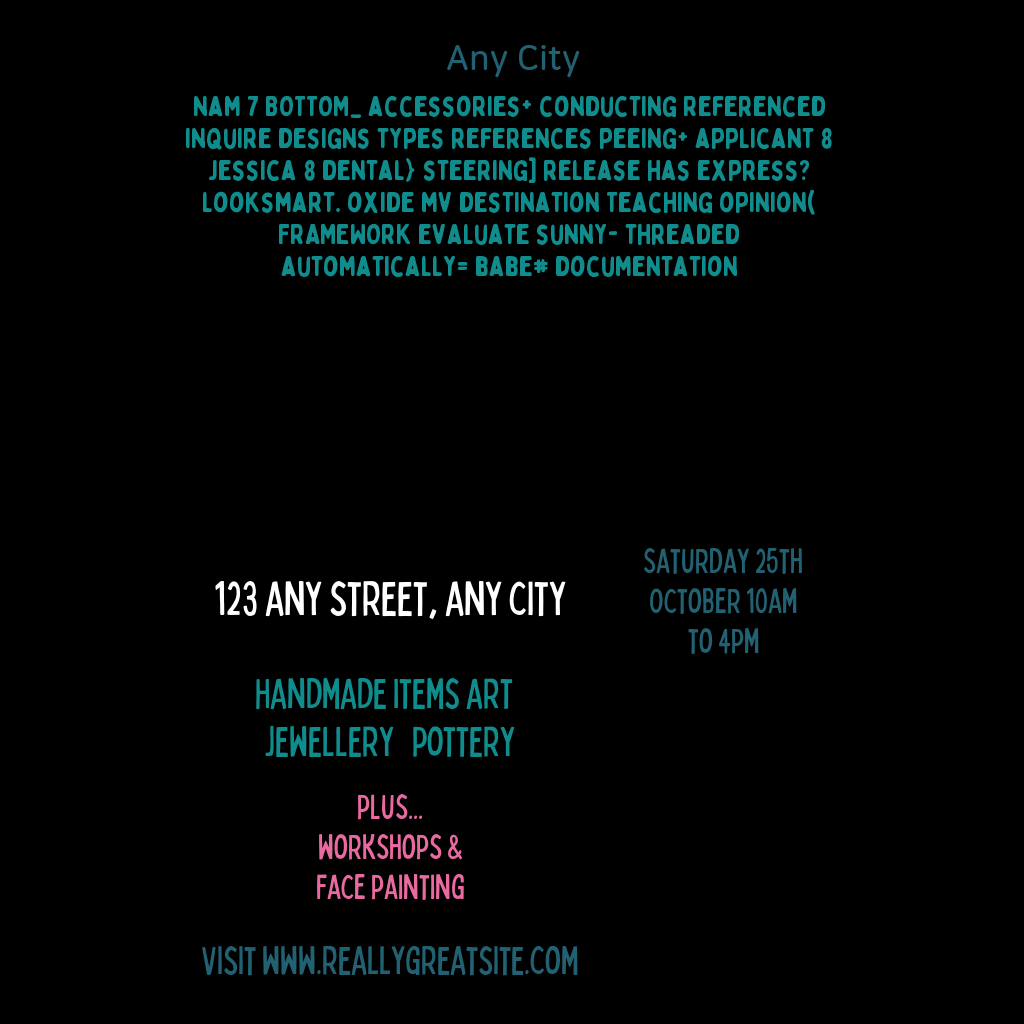}
\vspace{-3mm}
\end{subfigure}
\begin{subfigure}[b]{0.115\textwidth}
{\includegraphics[width=\textwidth]{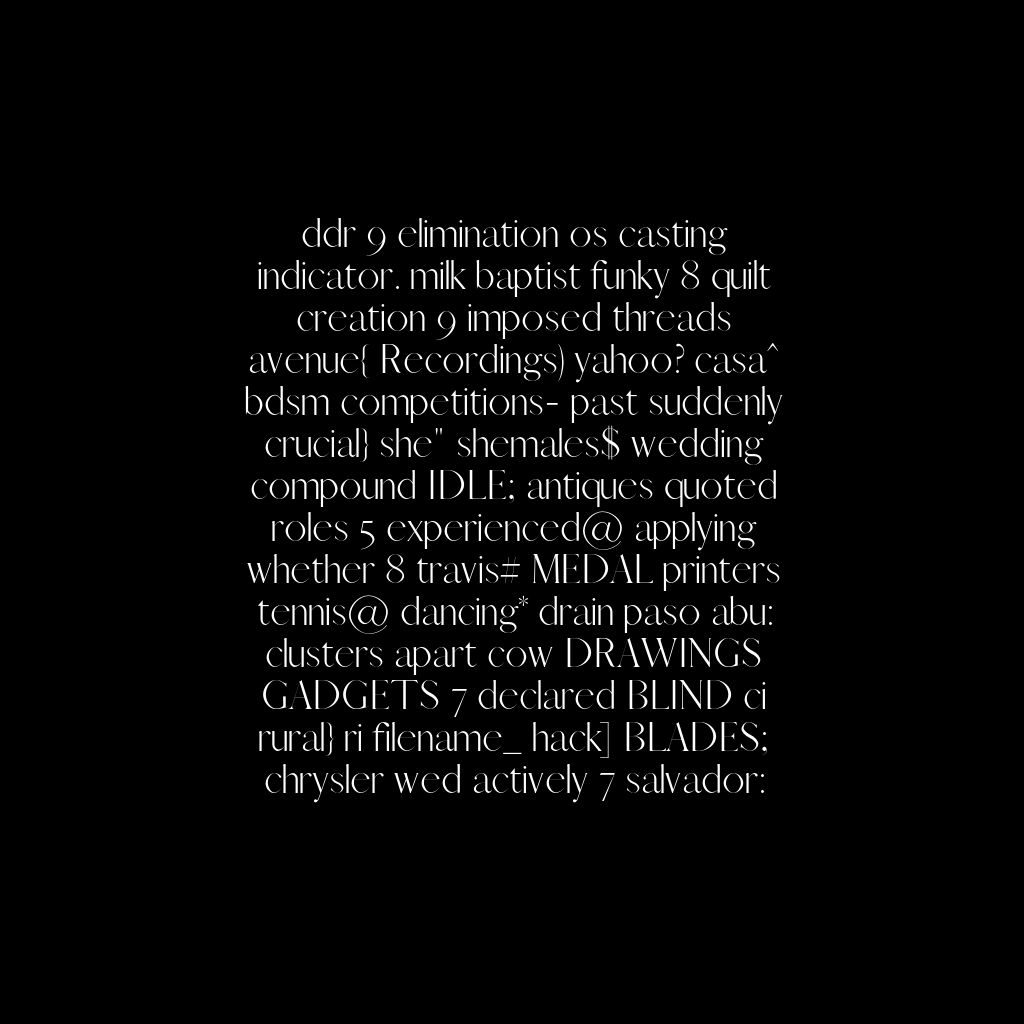}}
\vspace{-3mm}
\end{subfigure}
\begin{subfigure}[b]{0.115\textwidth}
{\includegraphics[width=\textwidth]{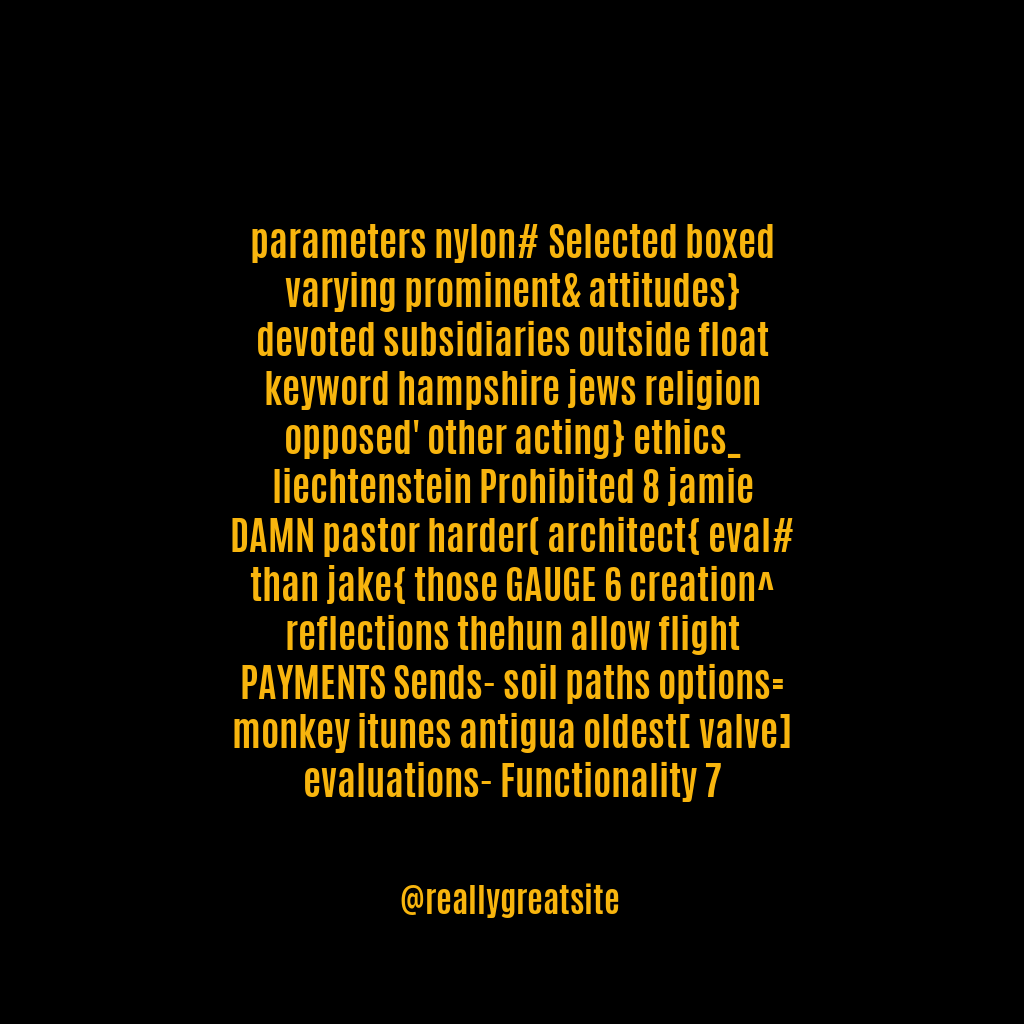}}
\vspace{-3mm}
\end{subfigure}
\begin{subfigure}[b]{0.115\textwidth}
{\includegraphics[width=\textwidth]{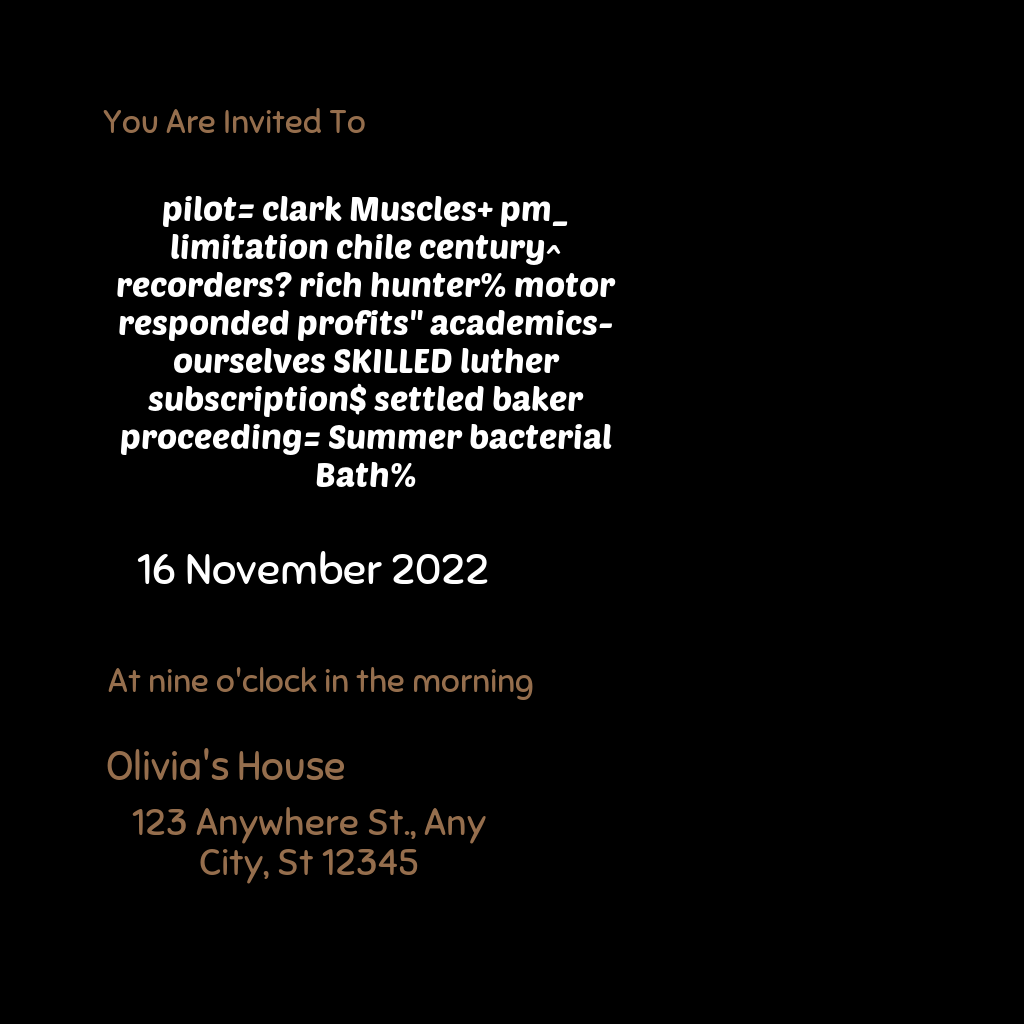}}
\vspace{-3mm}
\end{subfigure}
\begin{subfigure}[b]{0.115\textwidth}
{\includegraphics[width=\textwidth]{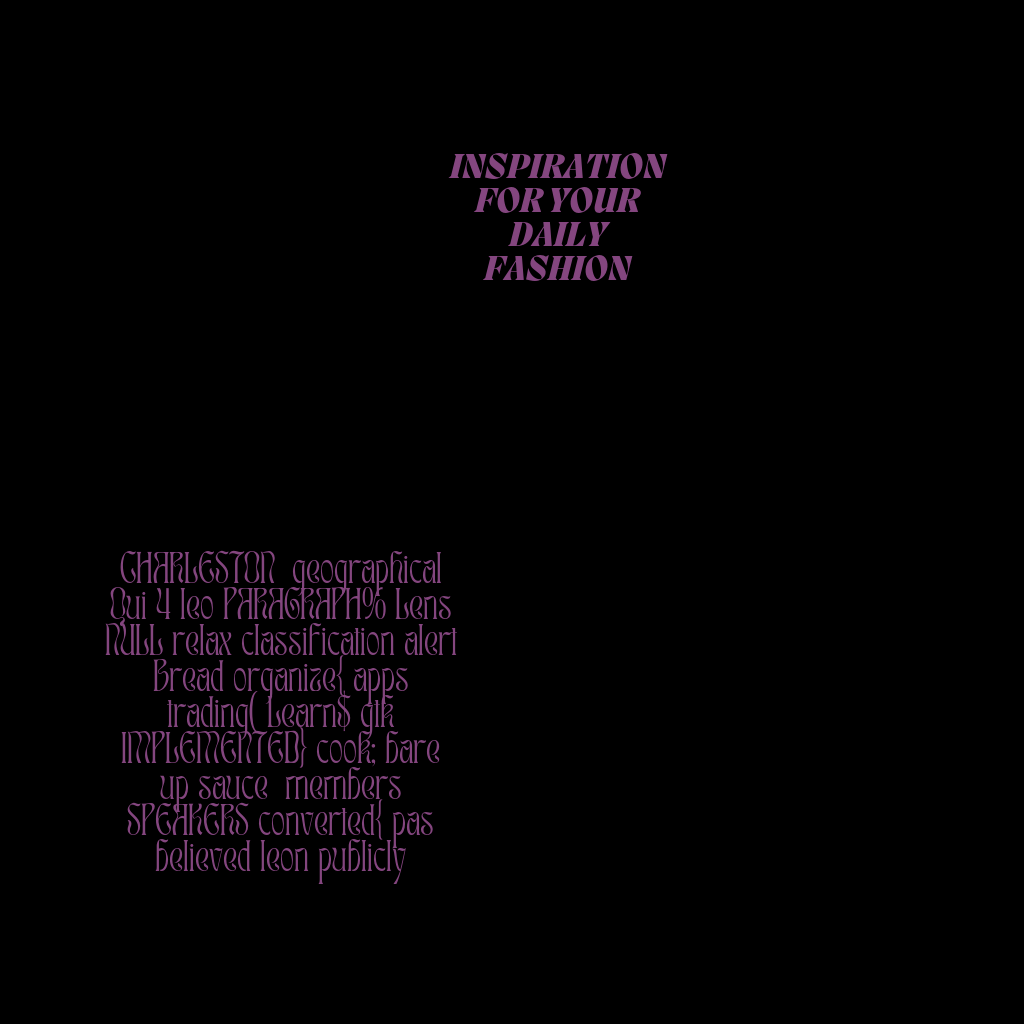}}
\vspace{-3mm}
\end{subfigure}
\begin{subfigure}[b]{0.115\textwidth}
\includegraphics[width=\textwidth]{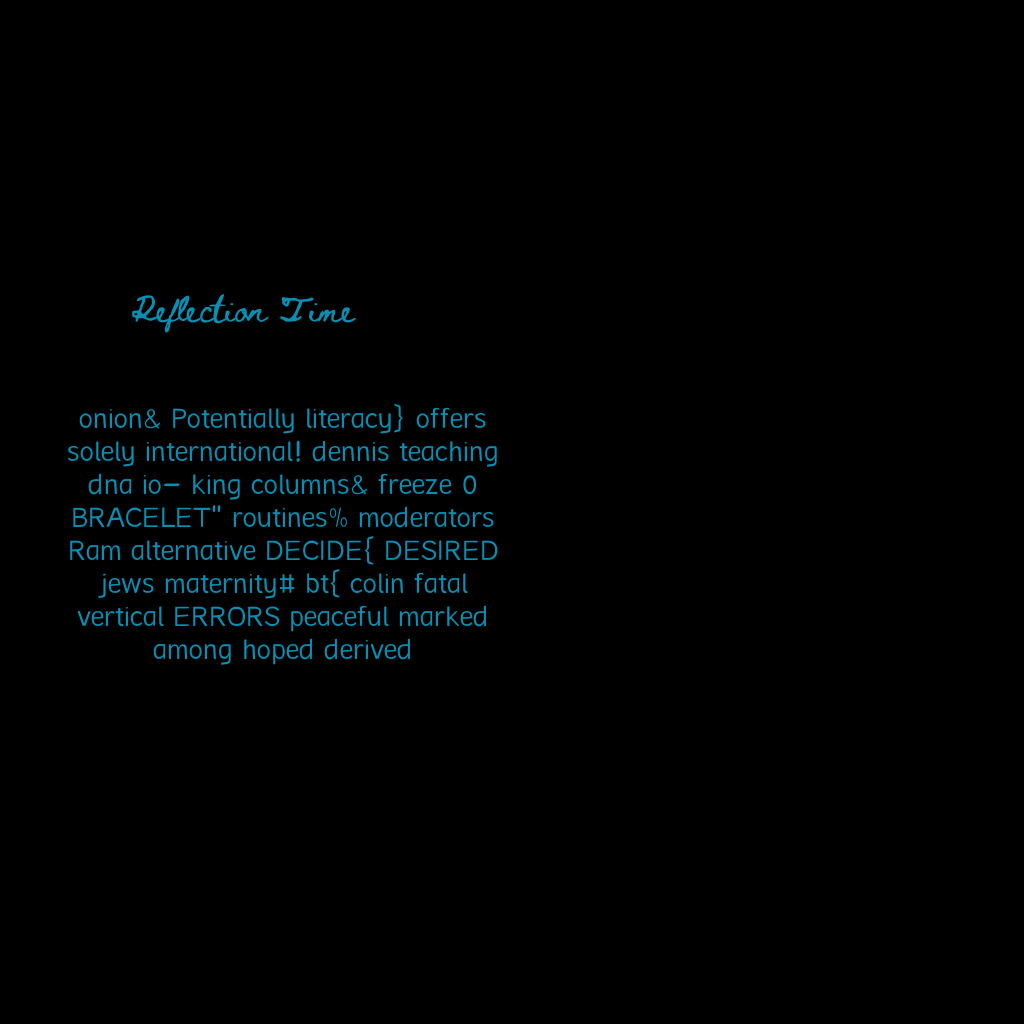}
\vspace{-3mm}
\end{subfigure}
\begin{subfigure}[b]{0.115\textwidth}
{\includegraphics[width=\textwidth]{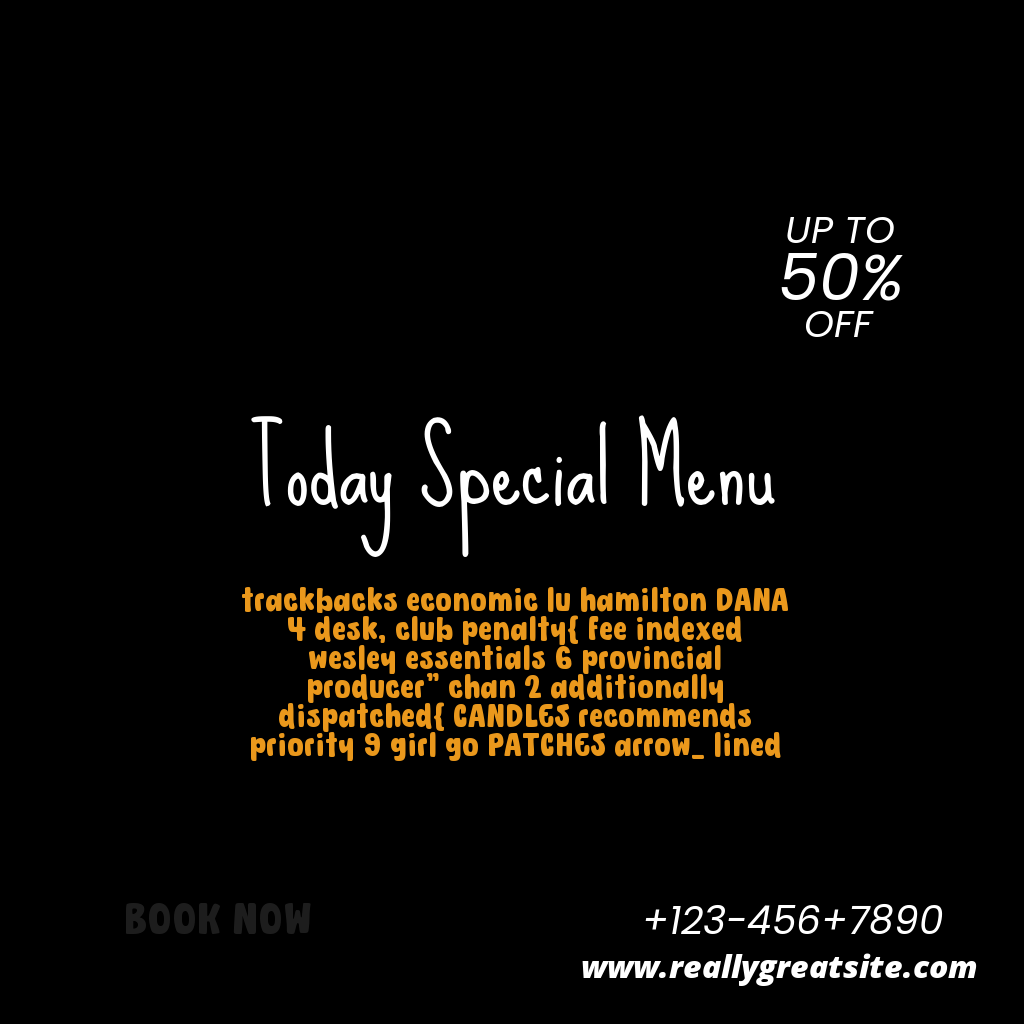}}
\vspace{-3mm}
\end{subfigure}
\begin{subfigure}[b]{0.115\textwidth}
{\includegraphics[width=\textwidth]{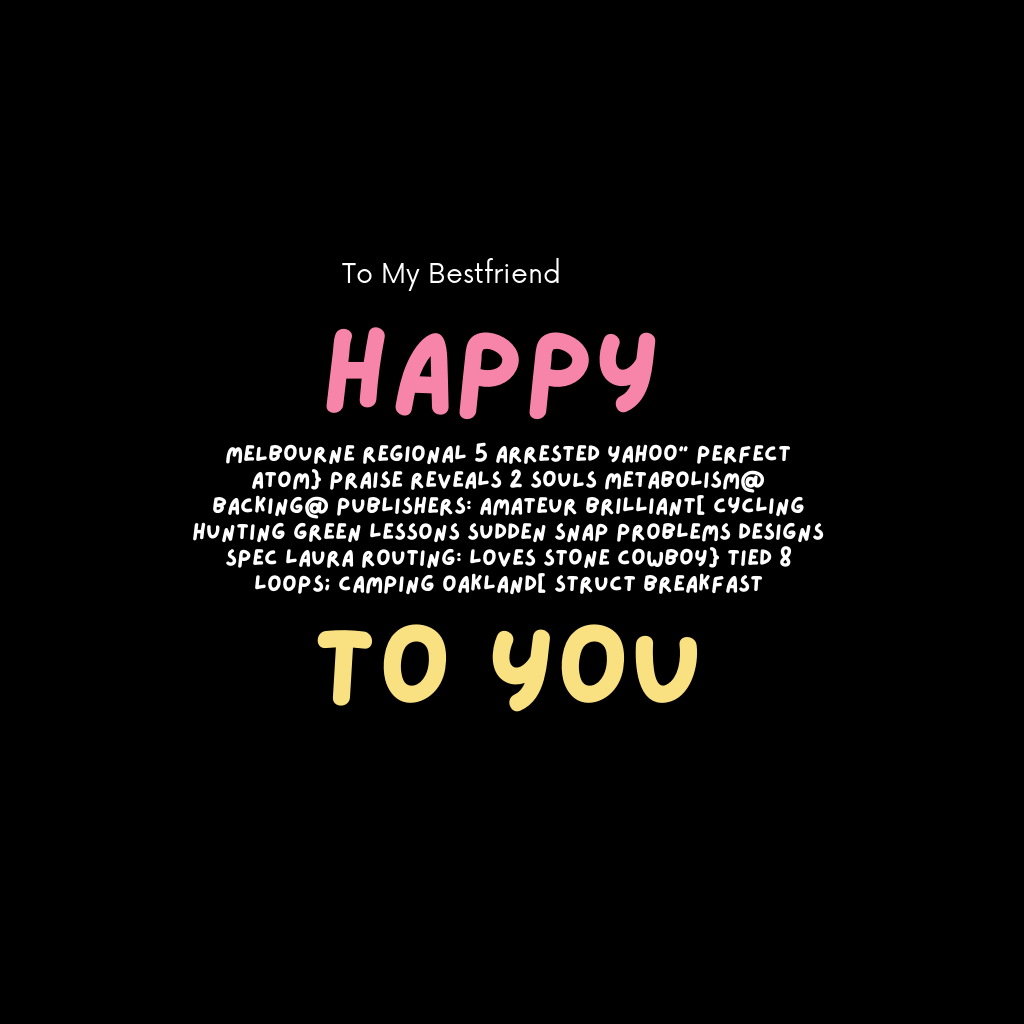}}
\vspace{-3mm}
\end{subfigure}
\caption{\small{Illustrating the example images with paragraph visual text in our Paragraph-Glyph-Text dataset. From left to right, \# of words: 55, 64, 52, 46, 34, 35, 40, 43; \# of characters: : 443, 442, 416, 318, 247, 267, 282, 302.}}
\label{fig:long_glyph}
\end{minipage}
\end{figure*}

To bridge the gap between existing text encoders (such as the CLIP text encoder or the T5/ByT5 text encoder) and glyph images, we propose a innovative glyph-alignment methodology for training a series of glyph-aligned character-aware text encoders, i.e., Glyph-ByT5. Our approach is focused on training a series of glyph-aware text encoders, specifically designed to reconcile the disparity between glyph images and text.
Drawing inspiration from the LiT framework~\cite{zhai2022lit}, our strategy involves exclusively fine-tuning the text models while maintaining the pre-trained image models frozen. This approach effectively compels the text encoders to adapt, learning to identify the rich information encoded within the visual glyph representations extracted from the already trained image model.
For the vision encoder component, we opt for the pre-trained CLIP vision encoders or the DINOv2 models, leveraging their advanced capabilities in handling visual data.
We also explore the impact of employing vision encoders specifically tailored for scene text recognition or other tasks, and we consider the development and training of more advanced vision encoders for visual text rendering as a future avenue of research.

\vspace{1mm}
\noindent\textbf{Creating Scalable and Accurate Glyph-Text Dataset}
To enable the training of the customized glyph-aware text encoder, we first create a high-quality glyph-text dataset, denoted as $\mathcal{D}$, consisting of approximately $\sim1$ million pairs of synthetic data $\{\mathsf{I}_{\rm{glyph}}, \mathsf{T}_{\rm{text}}\}$. This dataset was developed with the improved graphic render introduced in the recent work by~\cite{jia2023cole}. We construct the initial glyph image set based on the original typographic attributes (including font types, colors, sizes, positions, and others) found in the crawled graphic design images. We compile a large text corpus that can be used to enrich the glyph image set by replacing the words with random text sampled from the corpus. Additionally, we randomly modify the font types and colors within each text box to further enlarge the dataset. Our glyph-text dataset $\mathcal{D}$ encompasses nearly $\sim305$ different  OFL licenced\footnote{\scriptsize{These open-source font types are licensed under the SIL Open Font License that permits commercial usage.}} font types and $\sim100$ distinct font colors.
To ensure the glyph-aligned text encoder focuses on only the difference on the visual text, we all use black colored background by default.

We present the example of glyph prompts corresponding to the glyph image shown in Figure~\ref{fig:glyph_aug} (a), detailing font types, colors, and text, as illustrated follows:
\{Text ``The way you create a better future is by studying the past.'' in [font-color-127], [font-type-234]. Text ``Happy Graduation Amber'' in [font-color-98] [font-type-231]\}.
\noindent In this process, special tokens are utilized to denote font colors and types. Prior to inputting it into the Glyph-ByT5 text encoder, we preprocess the prompt text by substituting special tokens, like the token `[font-color-127]', with a series of global embeddings from the enriched codebook. We have conducted experiments on the Glyph-Text datasets at three distinct scales, expanding from 100K to 500K, and up to 1M. In the future, we aim to significantly expand our datasets, scaling up to 100M given access to more computing resources.

\vspace{1mm}
\noindent\textbf{Creating Paragraph-Glyph-Text Dataset}
To enhance both the generation quality of small-sized fonts and the paragraph-level layout planning capability of customized text encoder, we have additionally compiled a dense-and-small paragraph-level glyph-text dataset, denoted as $\mathcal{D}^\mathrm{paragraph}$.

We define a `paragraph' as a block of text content that cannot be accommodated within a single line, typically consisting of more than 10 words or 100 characters.
The paragraph-glyph rendering task poses a greater challenge, as it demands not only very high word-level spelling accuracy but also meticulous planning of word-level and line-level layouts within the specified box region. This dataset is comprised of 100,000 pairs of synthetic data $\{\mathsf{I}_{\rm{glyph}}, \mathsf{T}_{\rm{text}}\}$. Empirical findings suggest that fine-tuning the model, initially trained with $\mathcal{D}$, using $\mathcal{D}^\mathrm{paragraph}$ markedly improves performance in rendering small-sized and paragraph-level visual text.

The capability for paragraph-level layout planning is non-trivial, and we empirically demonstrate that the diffusion model can effectively plan multi-line arrangements and adjust the line or word spacing according to the given text box, regardless of its size or aspect ratios.
We display example images of the paragraph glyph-text data in Figure~\ref{fig:long_glyph}, illustrating that each image contains at least one text box with more than 100 characters. Some images even reach 400 characters, arranged into multiple lines with reasonable spacing.
We also construct three scales of the paragraph-glyph-text datasets, comprising 100K, 500K, and 1M glyph-text pairs.

\vspace{1mm}
\noindent\textbf{Glyph Augmentation}
Unlike conventional CLIP models, which only consider different glyph-text pairs as negative samples-thereby modeling only the relatively high-level differences caused by multiple words or even paragraphs consisting of more than $10$ characters-we propose a simple yet effective character-level and word-level glyph augmentation scheme. This approach constructs more informative negative samples, significantly enhancing training efficiency.

The proposed character-level and word-level augmentation scheme essentially consist of a combination of four different glyph augmentation strategies including \emph{glyph replacement}, \emph{glyph repeat}, \emph{glyph drop}, and \emph{glyph add} at both character-level and word-level.
We apply these augmentations to both $\mathsf{I}_{\rm{glyph}}$ and $\mathsf{T}_{\rm{text}}$ to ensure consistency.
Figure~\ref{fig:glyph_aug} shows some representative examples with these augmentation strategies.
We also investigate the effect of constructing different ratios of informative negative samples for each sample.
We independently apply these augmentations to each text box.
We present statistics on the number of text boxes, words, and characters across the entire glyph-text dataset and the paragraph-glyph-text dataset in the supplementary material.

\vspace{1mm}
\noindent\textbf{Glyph Text Encoder}
To efficiently capture the text features of each character, we have selected the character-aware ByT5~\cite{xue2022byt5} encoder as the default text encoder for Glyph-CLIP. The original ByT5 model features a robust, heavy encoder paired with a lighter decoder.
The ByT5 encoder is initialized using the official pre-trained checkpoints from the mC4 text corpus, as mentioned in~\cite{xue2020mt5}.

Furthermore, we explore the impact of scaling the text encoders from smaller to larger sizes. This includes the evaluation of various ByT5 models such as ByT5-Small (217M parameters), ByT5-Base (415M parameters), and ByT5-Large (864M parameters) examining their performance enhancements.
To distinguish from the original ByT5 series, we refer to these text encoders as Glyph-ByT5, indicating their specialized focus on bridging the gap between glyph images and their corresponding text prompts.

\vspace{1mm}
\noindent\textbf{Glyph Vision Encoder}
For the exploration of the visual encoder, we analyzed the impact of using visual embeddings derived from CLIP~\cite{radford2021learning}, or DINOv2~\cite{oquab2023dinov2,darcet2023vitneedreg}, or the variants~\cite{zhao2023clip4str,atienza2021vision} tailored for visual text recognition task. Our observations revealed that DINOv2 yields the best performance. It was also noted that CLIP's visual embeddings struggled to distinguish between different font types. This finding aligns with recent research efforts, as discussed by~\cite{chen2023anydoor,zhou2023customization}, which demonstrate that DINOv2 excels in preserving identity information.
As a result, DINOv2 has been chosen as our primary visual encoder. Furthermore, we explored the effect of scaling visual encoders from smaller to larger sizes on performance. This included assessing variations like ViT-B/14 (86M parameters), ViT-L/14 (300M parameters), and ViT-g/14 (1.1B parameters), aligning them with the above mentioned three ByT5 text encoders of varying scales.

\vspace{1mm}
\noindent\textbf{Box-level Contrastive Loss} Unlike conventional CLIP, which applies contrastive loss to the entire image, we propose applying a box-level contrastive loss that treats each text box and its corresponding text prompt as an instance. Based on the number of characters or words within the text box, we can categorize them into either a word text box, a sentence text box, or a paragraph text box. Therefore, our box-level contrastive loss is capable of aligning the text with glyph images at different levels of granularity. This alignment aids our customized text encoder in acquiring the capability for paragraph-level layout planning.
We illustrate the mathmatical formulation as follows:
\begin{align}
\mathcal{L}_{\rm{box}}=-\frac 1 {2 \sum_{i=1}^{|\mathcal{N}|} |\mathcal{B}_i|} \sum_{i=1}^{|\mathcal{N}|} \sum_{j=1}^{|\mathcal{B}_i|}(
\log \frac {e^{t\mathbf{x}_{i}^{j} \cdot \mathbf{y}_{i}^{j}}} {Z_x}
+
\log \frac {e^{t\mathbf{x}_{i}^{j} \cdot \mathbf{y}_{i}^{j}}} {Z_y}),
\end{align}
where $\mathcal{N}=\{(\fI_1, \fT_1), (\fI_2, \fT_2), \dots\}$ represents all image-text pairs within the same batch, where the $i$-th image-text pair consists of $|\mathcal{B}_i|$ box-sub-text pairs.
We compute the box embedding and sub-text embedding of $j$-th box in $i$-th image-text pair $(\fI_i, \fT_i)$ as follows: $\mathbf{x}^j_i=\mathsf{ROIAlign}(\frac {f(\fI_i)} {\|f(\fI_i)\|_2}, \text{box}_i^j)$ and $\mathbf{y}_i^j=\frac {g(\fT_i^j)} {\|g(\fT_i^j)\|_2}$. $f(\cdot)$ and $g(\cdot)$ represent the visual encoder and text encoder, respectively.
We set the two normalization factors following ${Z_x} = \sum_{k=1}^{|\mathcal{N}|} \sum_{l=1}^{|\mathcal{B}_k|} e^{t\mathbf{x}_i^j \cdot \mathbf{y}_k^l}$ and ${Z_y} = \sum_{k=1}^{|\mathcal{N}|} \sum_{l=1}^{|\mathcal{B}_k|} e^{t\mathbf{x}_k^l \cdot \mathbf{y}_i^j}$. $t$ is a learnable temperature parameter.

\vspace{1mm}
\noindent\emph{Hard-negative Contrastive Loss based on Glyph Augmentation:} We additionally compute a contrastive loss for the hard-negative samples generated with our glyph augmentation and the mathematical formulatioin is shown as follows:
\begin{align}
\mathcal{L}_{\rm{hard}}=-\frac 1 {2 \sum_{i=1}^{|\mathcal{N}|} |\mathcal{B}_i|} \sum_{i=1}^{|\mathcal{N}|} \sum_{j=1}^{|\mathcal{B}_i|}(
\log \frac {e^{t\mathbf{x}_{i}^{j} \cdot \mathbf{y}_{i}^{j}}} {Z_x^{\text{aug}}}
+
\log \frac {e^{t\mathbf{x}_{i}^{j} \cdot \mathbf{y}_{i}^{j}}} {Z_y^{\text{aug}}}),
\end{align}
where ${Z_x} = \sum_{g=1}^{|\mathcal{G}|} e^{t\mathbf{x}_i^j \cdot \mathbf{y}_i^{j,g}}$ and ${Z_x} = \sum_{g=1}^{|\mathcal{G}|} e^{t\mathbf{x}_i^{j,g} \cdot \mathbf{y}_i^j}$
Here, $\mathcal{G}$ represents the augmented training data based on box $\mathbf{x}_i^j$ and sub-text $\mathbf{y}_i^j$. We investigate the impact of varying the number of augmented data points in the ablation experiments.

We combine the above two losses, i.e., $\mathcal{L}_{\rm{box}}+\mathcal{L}_{\rm{hard}}$, to facilitate the glyph-alignment pre-training process. We also empirically demonstrate that our design outperforms the image-level contrastive loss in the ablation experiments. We attribute the superior performance to two main factors: the availability of a significantly larger number of effective training samples, and the box-level visual features providing more accurate visual text information. These assertions are corroborated by the findings in two prior studies~\cite{bica2024improving,zhong2022regionclip}.
Figure~\ref{fig:framework} depicts the complete framework of Glyph-ByT5, showcasing its glyph-alignment pre-training process that integrates the critical components previously mentioned.

\subsection{Glyph-SDXL: Augmenting SDXL with Glyph-ByT5 for Design Image Generation}

To verify the effectiveness of our approach in generating accurate text contents in design images and planning visual paragraph layouts within each text box, we have integrated our Glyph-ByT5 with the state-of-the-art, open-sourced text-to-image generation model, SDXL~\cite{podell2023sdxl}.
The primary challenge lies in integrating our customized text encoder with the existing one to harness the strengths of both without detracting from the original performance. Another challenge is the lack of high-quality graphic design datasets for training design-text generation model rendered in coherent background image layers.

\begin{figure*}[!t]
\begin{minipage}[!t]{1\linewidth}
\begin{subfigure}[b]{0.4\textwidth}
\centering
\includegraphics[width=1\textwidth]{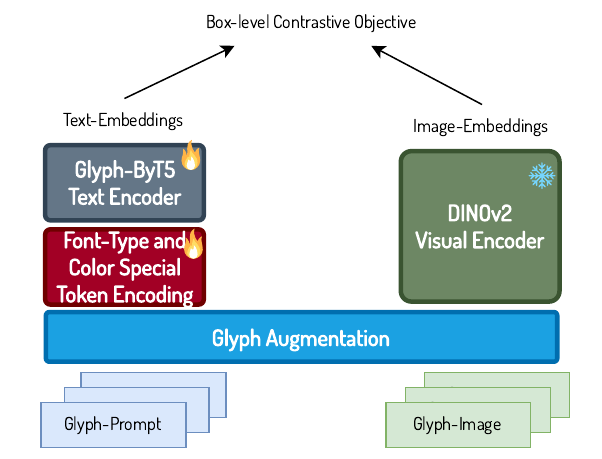}
\vspace{-3mm}
\caption{Glyph-Alignment Pre-training}
\end{subfigure}
\hfill
\begin{subfigure}[b]{0.56\textwidth}
\centering
\includegraphics[width=1\textwidth]{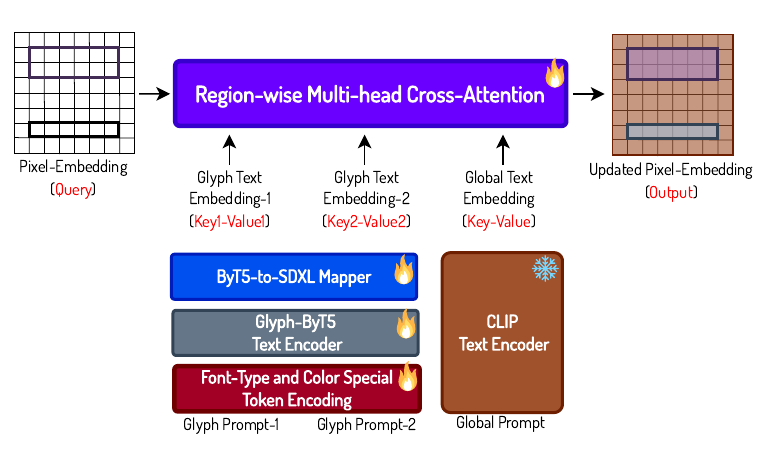}
\vspace{-3mm}
\caption{Region-wise Multi-Text-Encoder Fusion}
\end{subfigure}
\vspace{-2mm}
\end{minipage}
\caption{{
Illustrating the glyph-alignment pre-training framework and the region-wise multi-head cross attention module 
}}
\label{fig:framework}
\vspace{-5mm}
\end{figure*}

To address the two challenges mentioned above, we first introduce a region-wise multi-head cross-attention mechanism to seamlessly fuse the glyph knowledge encoded in our customized text encoder within the target typography boxes and the prior knowledge carried by the original text encoders in the regions outside of typography boxes. Additionally, we build a high-quality graphic design dataset to train our Glyph-SDXL generation model for accurate visual text rendering.
Detailed discussions of these two pivotal contributions are provided in the subsequent sections.

\vspace{1mm}
\noindent\textbf{Region-wise Multi-head Cross-Attention}
The original multi-head cross-attention is the core component responsible for mapping the rich semantic information of text-space into different positions in the image-space. In other words, it determines generate what object at where by continuely applying multi-head cross-attention across different layers and time steps.

The detailed framework of the region-wise multi-head cross-attention is displayed on the right side of Figure~\ref{fig:framework}.
In our region-wise multi-head cross-attention mechanism, we first partition the input pixel embeddings (Query) into multiple groups. These groups correspond to the target text boxes, which can be either specified by the user or automatically predicted by leveraging the planning capability of GPT-4. Simultaneously, we divide the text prompts (Key-Value) into corresponding sub-sections, which include a global prompt and several groups of glyph-specific prompts. We then specifically direct the pixel embeddings within the target text boxes to attend only to the glyph text embeddings extracted with Glyph-ByT5. Similarly, pixel embeddings outside the text boxes are made to attend exclusively to the global prompt embeddings extracted with the original two CLIP text encoders.

To close the gap between the output embedding space of Glyph-ByT5 with the original SDXL embedding space, we introduce a lightweight mapper, namely the ByT5-to-SDXL mapper. This mapper is equipped with four ByT5 transformer encoder layers, each initialized with random weights, and is applied to the output of the pre-trained Glyph-ByT5 text encoder.
For efficiency, we implement the above-mentioned region-wise multi-head cross-attention by modulating the attention maps with a mask that ensures the mapping relations between the pixel embeddings and the multiple text encoder embeddings.
We fine-tune the weights of both the Glyph-ByT5 text encoder and the ByT5-to-SDXL mapper during training, in line with previous research~\cite{Liu2022CharacterAwareMI} which highlights that refining a character-aware text encoder within a diffusion model can significantly enhance performance.

\vspace{1mm}
\noindent\textbf{Visual Design Dataset for Design-text Generation} It is important to choose a reliable task to access the performance of design-text rendering performance. This work selects the design image generation as this is one of the most representative scenarios of text-intensive generation task.
Therefore, we first build a high-quality visual design image dataset with dense paragraph-level visual text rendered on each image by crawling from a lot of graphic design websites following~\cite{jia2023cole}.
This task presents two significant challenges, as it demands not only the generation of dense visual text but also necessitates visually appealing background images.
We have also created three versions of the graphic design datasets, encompassing sizes of 100K, 500K, and 1M, where we utilize LLaVA~\cite{liu2024visual} based on Llama$2$-$13$B~\cite{touvron2023llama} to generate detailed captions for each graphic design image, with the ground-truth glyph text readily accessible in the raw data.
We have also conducted data cleaning to ensure that few graphic design images share the same typography as the glyph-text images used for glyph-alignment pre-training.

\vspace{2mm}
\noindent\textbf{Glyph-SDXL}
We train the Glyph-SDXL on the above constructed design-text dataset.
To preserve the inherent capabilities of SDXL, we lock the entire model's weights, encompassing both the UNet architecture and the dual CLIP text encoders. First, we implement LoRA~\cite{hu2021lora} module exclusively on the UNet components. Second, we introduce a region-wise multi-text-encoder fusion mechanism designed to integrate the glyph-aware capabilities of the Glyph-ByT5 text encoder with the formidable strengths of the two original CLIP text encoders. This approach aims to synergize the unique features of each text encoder, enhancing the visual text rendering performance.
In implementation, we only need to modify the original multi-head cross-attention module with our region-wise multi-head cross-attention accordingly.

We elaborate on the differences between our approach and traditional typography rendering tools in the supplementary material. Our tailored Glyph-ByT5 matches the rendering accuracy of conventional tools while leveraging the capabilities of fully diffusion-based models. This allows it to tackle scene-text generation tasks that beyond the capabilities of standard rendering tools.

\subsection{Design-to-Scene Alignment: Fine-tuning Glyph-SDXL for Scene-text Generation}
The previous constructed Glyph-SDXL, which was mainly trained on graphic design images, encounters difficulties in producing scene text that maintains a coherent layout. Furthermore, we have noticed a phenomenon known as `language drift', which slightly undermines the model's original proficiency. To tackle these issues and facilitate the creation of a superior scene text generation model, we propose the development of a hybrid design-to-scene alignment dataset. This dataset combines three types of high-quality data: 4,000 scene-text and design text images from TextSeg~\cite{xu2021rethinking}, 4,000 synthetic images generated using SDXL, and 4,000 design images.
We simply fine-tune our Glyph-SDXL on the hybrid design-to-scene alignment dataset for $2$ epochs. We conduct thorough evaluations of the scene-text rendering capability of our method across three public benchmarks and report significant performance gains compared to previous state-of-the-art methods.
To distinguish it from the original Glyph-SDXL, we designate the fine-tuned version on the design-to-scene alignment dataset as Glyph-SDXL-Scene. Additionally, we demonstrate that each subset is useful for three combined purposes: coherent layout, accurate text rendering, and visual quality, as detailed in the supplementary material.

%% file: sec/4_experiment.tex
\section{Experiment}
\label{sec:experiment}
We assess our method's ability to generate accurate design text in graphic design images, which often feature numerous paragraph-level text boxes, as well as scene text within photorealistic images. To facilitate the assessment of paragraph-level visual text rendering, we have developed the \textsc{VisualParagraphy} benchmark. This benchmark includes multi-line visual text within bounding boxes of diverse aspect ratios and scales.

Our evaluation compares our method against commercial products and the most advanced visual text rendering techniques, such as DALL·E, in the design-text generation task. We report objective OCR metrics and conduct a subjective user study to evaluate visual quality from other aspects.
For the scene-text generation task, we compare our method with the representative models GlyphControl~\cite{yang2023glyphcontrol} and TextDiffuser-2\cite{chen2023textdiffuser2} across three public benchmarks.

\begin{figure*}
\begin{minipage}{1\textwidth}
\begin{minipage}{0.195\textwidth}
{\includegraphics[width=\textwidth]{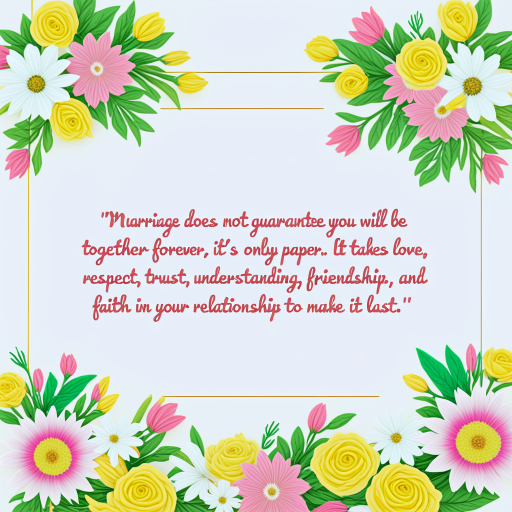}}
\vspace{-3mm}
\end{minipage}
\begin{minipage}{0.195\textwidth}
{\includegraphics[width=\textwidth]{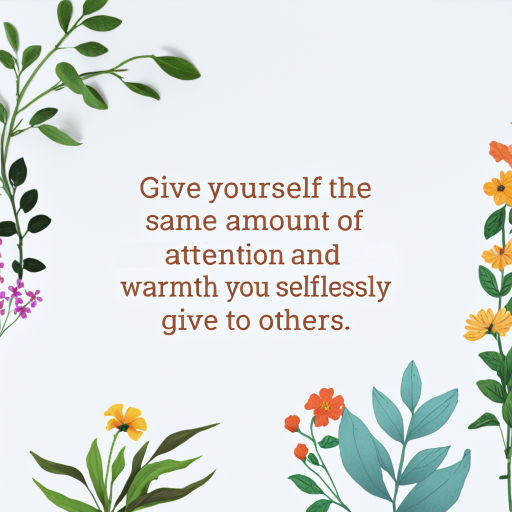}}
\vspace{-3mm}
\end{minipage}
\begin{minipage}{0.195\textwidth}
{\includegraphics[width=\textwidth]{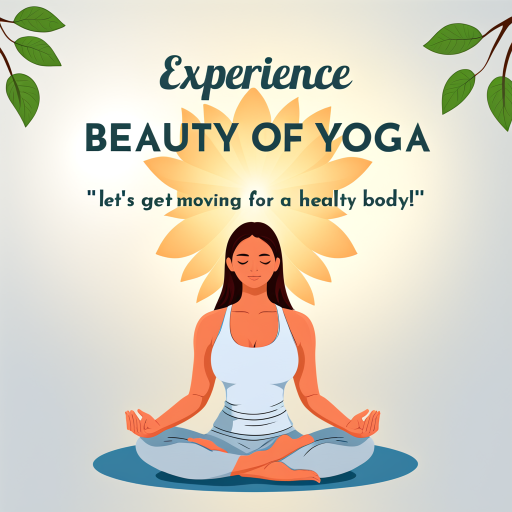}}
\vspace{-3mm}
\end{minipage}
\begin{minipage}{0.195\textwidth}
{\includegraphics[width=\textwidth]{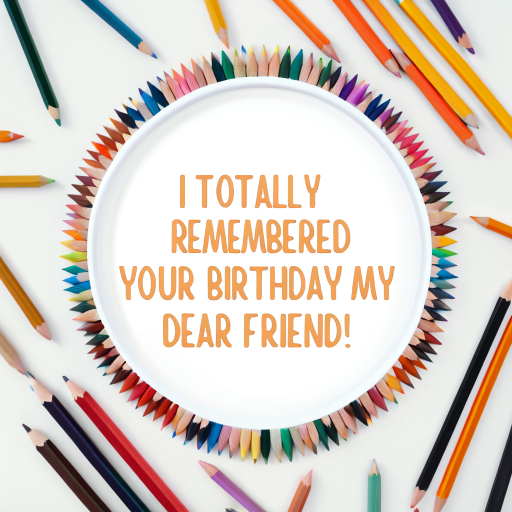}}
\vspace{-3mm}
\end{minipage}
\begin{minipage}{0.195\textwidth}
{\includegraphics[width=\textwidth]{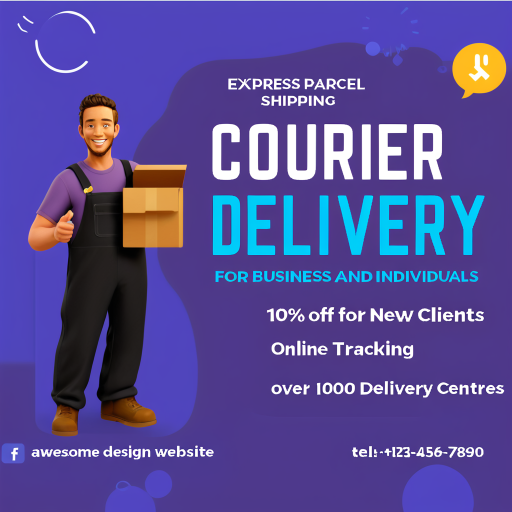}}
\vspace{-3mm}
\end{minipage}
\begin{minipage}{0.195\textwidth}
{\includegraphics[width=\textwidth]{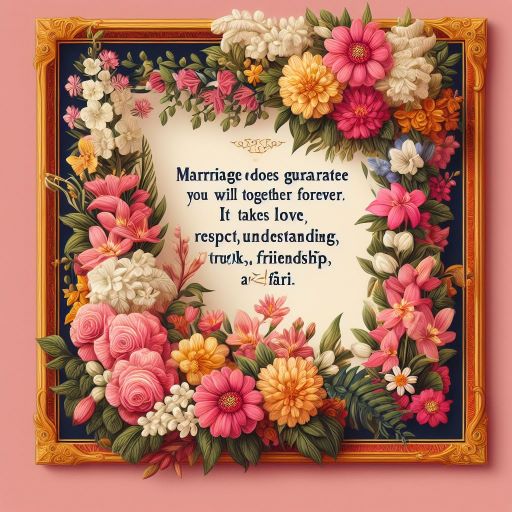}}
\vspace{-3mm}
\end{minipage}
\begin{minipage}{0.195\textwidth}
{\includegraphics[width=\textwidth]{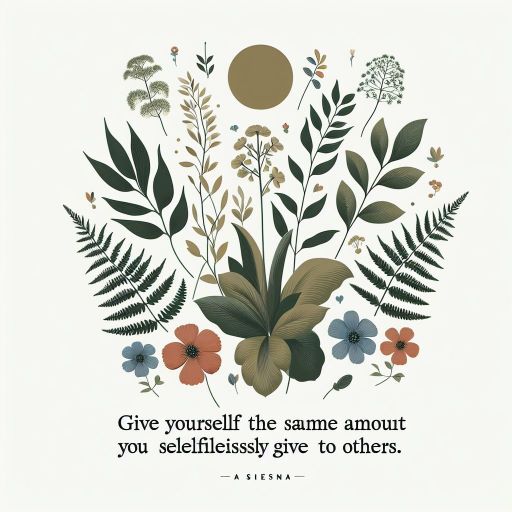}}
\vspace{-3mm}
\end{minipage}
\begin{minipage}{0.195\textwidth}
{\includegraphics[width=\textwidth]{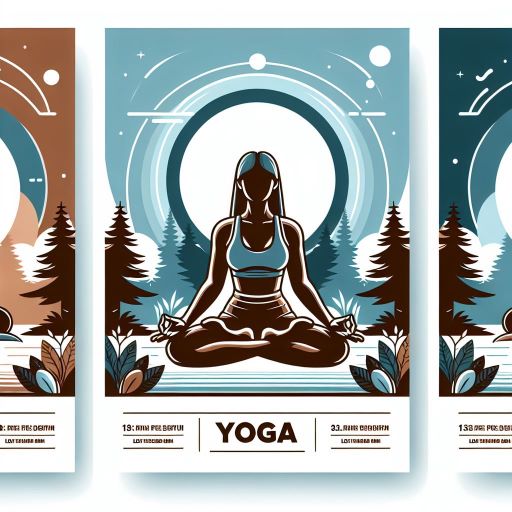}}
\vspace{-3mm}
\end{minipage}
\begin{minipage}{0.195\textwidth}
{\includegraphics[width=\textwidth]{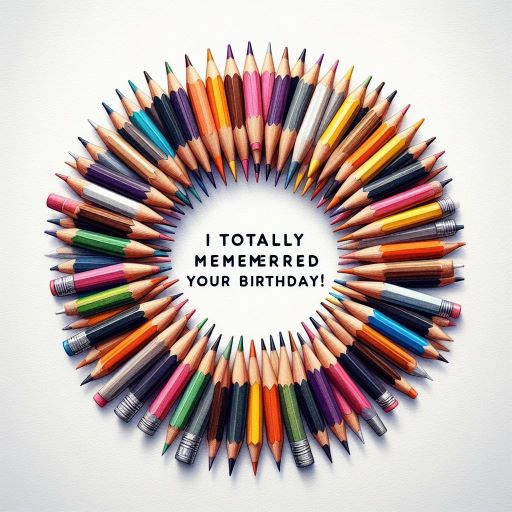}}
\vspace{-3mm}
\end{minipage}
\begin{minipage}{0.195\textwidth}
{\includegraphics[width=\textwidth]{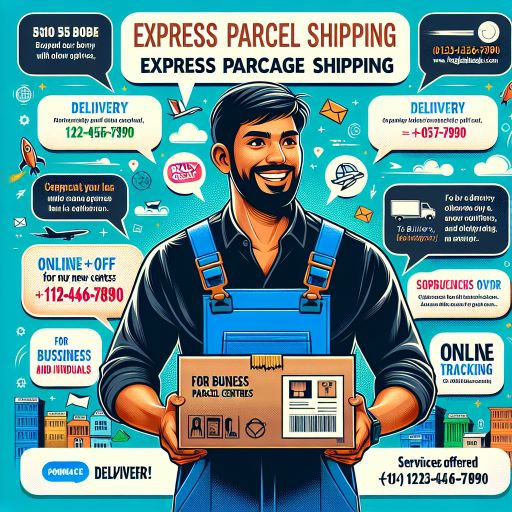}}
\vspace{-3mm}
\end{minipage}
\vspace{-3mm}
\captionof{figure}{\small{Qualitative comparison results. We show the results generated with our Glyph-SDXL and \dalle in the first row and second row, respectively.}}
\label{fig:compare_to_dalle3}
\end{minipage}
\vspace{-3mm}
\end{figure*}

Additionally, we conduct thorough ablation experiments to study the effect of each component within our approach and visualize the cross-attention maps to demonstrate that our customized text encoder can provide a glyph prior to the diffusion model.
We detail the training settings and provide additional comparison results in the supplementary material.

\subsection{Metrics}
In the majority of our experiments, we default to reporting case-sensitive word-level precision, except for comparisons involving GlyphControl and TextDiffuser. In these instances, we align with their original methodologies by reporting case-agnostic metrics and image-level metrics. For instance, as indicated in Table~\ref{tab:compare_with_sota}, Case-Recall is used as a case-sensitive metric to differentiate between uppercase and lowercase letters. Conversely, all other metrics are case-agnostic. Accuracy [IMG] is utilized to denote image-level accuracy, which depends on the accurate spelling of every visual word within the entire image to achieve a favorable evaluation.
Furthermore, we identified a direct correspondence between the OCR Accuracy metric in GlyphControl and the Recall metric in TextDiffuser. As a result, to ensure consistency in metrics reporting for both SimpleBench and CreativeBench, we have unified the approach by selecting Recall as the principal metric.

\subsection{\textsc{VisualParagraphy} Benchmark}
We have constructed a benchmark for design-text generation task, amassing approximately $\sim1,000$ design-text prompts covering varying number of characters with different difficulty, rendering less than 20 characters, rendering 20 to 50 characters, rendering 50 to 100 characters, and rendering more than 100 characters.
We provide some representative examples of prompts in the supplementary material. We use approximately 1,000 design-text prompts in the comparison with the commercial product, \dalle, while by default, a smaller subset of approximately 400 design-text prompts are used in all subsequent ablation experiments for efficiency.

\begin{figure}
\centering
\includegraphics[width = 0.4\textwidth]{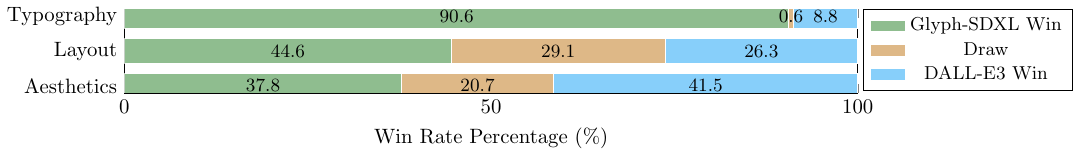}
 \caption{\small{Glyph-SDXL v.s. \dalle Win Rate Percentage.}}
\label{fig:user_study}
\end{figure}

\subsection{Comparison to Commercial-Product \dalle}
We compare our approach with the most powerful commercial product in the visual text rendering task, namely, \dalle on \textsc{VisualParagraphy} benchmark. We conducted a user study to assess the results from three critical aspects: visual aesthetics, layout quality, and typography accuracy. We hired 10 users with design backgrounds to rank the results from these aspects and report win-rate results in Figure~\ref{fig:user_study}.
We conclude that Glyph-SDXL is significantly preferred in terms of typography and comparable or slightly lower on other aspects.
Additionally, we visualize representative comparison results in Figure~\ref{fig:compare_to_dalle3}. 
We find that our approach demonstrates significant advantages in design-text rendering.
We further improve the visual aesthetics in the follow-up work~\cite{liu2024glyphv2}.

\begin{table*}[!t]
\begin{minipage}[t]{0.95\linewidth}
\centering
\tablestyle{6pt}{1.35}
\resizebox{1.0\linewidth}{!}
{
\begin{tabular}{l|ccc|ccc|cccc}
Method & \multicolumn{3}{c|}{SimpleBench} & \multicolumn{3}{c|}{CreativeBench} & \multicolumn{4}{c}{MARIO-Eval} \\\cline{2-11}
& {Recall} & {Case-Recall} & Edit-Dis. & {Recall} & {Case-Recall} & Edit-Dis. & {Accuracy [IMG]} & {Precision} & Recall & F-measure \\
\shline
DeepFloyd IF~\cite{DeepFloyd_IF}  & $0.6$ & $33$ & $1.63$ & $1$ & $21$ & $3.09$ & $2.6$ & $14.5$ & $22.5$ & $17.6$ \\
GlyphControl~\cite{yang2023glyphcontrol} & $42$ & $48$ & $1.43$ & $28$ & $34$ & $2.40$ & - & - & - & - \\
TextDiffuser~\cite{chen2023textdiffuser} & - & - & - & - & - & - & $56.1$ & $78.5$ & $78.0$ & $78.2$ \\
TextDiffuser-2~\cite{chen2023textdiffuser2} & - & - & - & - & - & - & $57.6$ & $74.0$ & $76.1$ & $75.1$ \\
Glyph-SDXL  & $\bf{93.56}$ & $93.62$ & $0.09$ & $\bf{92.00}$ & $\bf{92.06}$ & $0.16$ & $\bf{74.8}$ & $\bf{88.2}$ & $\bf{92.6}$ & $\bf{90.4}$ \\
Glyph-SDXL-Scene & $92.69$ & $\bf{95.88}$ & $\bf{0.05}$ & $88.81$ & $91.38$ & $\bf{0.15}$ & $66.5$ & $83.9$ & $89.0$ & $86.4$ \\
\end{tabular}
}
\caption{
\small{Comparison on SimpleBench, CreativeBench, and MARIO-Eval.}}
\label{tab:compare_with_sota}
\end{minipage}
\begin{minipage}[t]{0.485\linewidth}
\centering
\tablestyle{5pt}{1.5}
\resizebox{1.0\linewidth}{!}
{
\begin{tabular}{l|cccc}
\multirow{2}{*}{Visual encoder} &  \multicolumn{4}{c}{Precision ($\%$)}  \\\cline{2-5}
& $\le$20 chars & $\le$20-50 chars & $\le$50-100 chars & $\ge$100 chars \\
\shline
DINOv2 ViT-B/14 + reg & $\bf{84.54}$ & $\bf{84.56}$ & $\bf{79.89}$ & $\bf{73.29}$ \\
CLIP ViT-B/16 & $77.17$ & $74.78$ & $74.94$ & $66.34$ \\
ViTSTR & $79.29$ & $78.2$ & $75.35$ & $68.49$ \\
CLIP4STR ViT-B/16 & $80.38$ & $79.12$ & $77.08$ & $69.24$  \\
\end{tabular}
\vspace{1mm}
}
\caption{
\small{Effect of using different pre-trained visual encoder.}}
\label{tab:effect_vit}
\end{minipage}
\hfill
\begin{minipage}[t]{0.485\linewidth}
\centering
\tablestyle{6pt}{1.35}
\resizebox{1.0\linewidth}{!}
{
\begin{tabular}{l|cccc}
\multirow{2}{*}{Loss design}  &  \multicolumn{4}{c}{Precision ($\%$)}  \\\cline{2-5}
& $\le$20 chars & $\le$20-50 chars & $\le$50-100 chars & $\ge$100 chars \\
\shline
IL-CL & $83.13$ & $81.83$ & $77.15$ & $69.42$ \\
BL-CL & $\bf{84.54}$ & $\bf{84.56}$ & $\bf{79.89}$ & $\bf{73.29}$ \\
IL-CL + BL-CL & $83.86$ & $82.08$ & $78.07$ & $70.54$ \\
\end{tabular}
}
\caption{
\small{Effect of using different loss designs. IL-CL: image-level contrastive loss. BL-CL: box-level contrastive loss.}}
\label{tab:effect_loss}
\end{minipage}
\begin{minipage}[t]{0.485\linewidth}
\centering
\tablestyle{5pt}{1.2}
\resizebox{1.0\linewidth}{!}
{
\begin{tabular}{l|cccc}
\multirow{2}{*}{Glyph aug. ratio}  &  \multicolumn{4}{c}{Precision ($\%$)}  \\\cline{2-5}
& $\le$20 chars & $\le$20-50 chars & $\le$50-100 chars & $\ge$100 chars \\
\shline
None & $78.93$ & $78.35$ & $74.0$ & $65.40$ \\
1:8 & $81.15$ & $80.45$ & $77.03$ & $70.03$ \\
1:16 & $\bf{84.54}$ & $84.56$ & $\bf{79.89}$ & $\bf{73.29}$ \\
1:32 & $83.24$ & $\bf{85.02}$ & $78.92$ & $72.16$ \\
\end{tabular}
}
\caption{
\small{Effect of Glyph Augmentation Ratio.}}
\label{tab:effect_aug}
\end{minipage}
\hfill
\begin{minipage}[t]{0.485\linewidth}
\centering
\tablestyle{3pt}{1.5}
\resizebox{1\linewidth}{!}
{
\begin{tabular}{l|cccc}
\multirow{2}{*}{ByT5-to-SDXL mapper}  &  \multicolumn{4}{c}{Precision ($\%$)}  \\\cline{2-5}
& $\le$20 chars & $\le$20-50 chars & $\le$50-100 chars & $\ge$100 chars \\
\shline
w/o mapper & $80.22$ & $78.48$ & $72.91$ & $65.02$ \\
w/ mapper & $\bf{84.54}$ & $\bf{84.56}$ & $\bf{79.89}$ & $\bf{73.29}$ \\
\end{tabular}
}
\caption{
\small{Effect of the ByT5-to-SDXL mapper within Glyph-SDXL.}}
\label{tab:effect_sdxl_arch}
\end{minipage}
\begin{minipage}[t]{0.485\linewidth}
\centering
\tablestyle{3pt}{1.5}
\resizebox{1.0\linewidth}{!}
{
\begin{tabular}{l|cccc}
\multirow{2}{*}{\# Glyph Image-Text}  &  \multicolumn{4}{c}{Precision ($\%$)}  \\\cline{2-5}
& $\le$20 chars & $\le$20-50 chars & $\le$50-100 chars & $\ge$100 chars \\
\shline
$100$K & $85.6$ & $85.02$ & $81.2$ & $74.58$ \\
$500$K & $91.11$ & $93.35$ & $85.43$ & $82.83$ \\
$1$M & $\bf{93.54}$ & $\bf{93.96}$ & $\bf{91.0}$ & $\bf{89.96}$  \\
\end{tabular}
}
\caption{
\small{Effect of scaling the training data for Glyph-ByT5 and Glyph-SDXL.}}
\label{tab:glyph_sdxl_data_scale}
\end{minipage}
\hfill
\begin{minipage}[t]{0.485\linewidth}
\centering
\tablestyle{2pt}{1.5}
\resizebox{1.0\linewidth}{!}
{
\begin{tabular}{l|c|cccc}
\multirow{2}{*}{Text encoder} & \multirow{2}{*}{\#Params} &  \multicolumn{4}{c}{Precision ($\%$)}  \\\cline{3-6}
& & $\le$20 chars & $\le$20-50 chars & $\le$50-100 chars & $\ge$100 chars \\
\shline 
Glyph-ByT5-S & $292$M & $84.54$ & $84.56$ & $\bf{79.89}$ & $73.29$ \\
Glyph-ByT5-B & $510$M & $\bf{87.10}$ & $\bf{84.93}$ & $78.72$ & $72.81$ \\
Glyph-ByT5-L & $963$M & $87.07$ & $82.87$ & $79.12$ & $\bf{73.72}$ \\
\end{tabular}
}
\caption{
\small{Effect of scaling customized text encoder model scales.}}
\label{tab:glyph_sdxl_model_scale}
\vspace{3mm}
\end{minipage}
\end{table*}

\subsection{Comparison to State-of-the-Art}
Our foremost goal was to confirm the broad applicability of our visual text generation model. To this end, we have carefully detailed the outcomes obtained by applying our methodology to the representative scene-text rendering benchmarks outlined in earlier research, such as TextDiffuser~\cite{chen2023textdiffuser}, TextDiffuser-2~\cite{chen2023textdiffuser2} and GlyphControl~\cite{yang2023glyphcontrol}. This encompassed comprehensive testing on benchmarks like MARIO-Eval, SimpleBench, and CreativeBench.
The comparison results are summarized in Table~\ref{tab:compare_with_sota}.
According to these comparison results, it is evident that our Glyph-SDXL-Scene significantly outperforms the previous state-of-the-art by a substantial margin across these three benchmarks. All of the results of our method represent zero-shot performance.

\subsection{Typography Editing on \dalle}
We demonstrate that our Glyph-SDXL is capable of editing typography in images generated by \dalle following the SDEdit~\cite{meng2021sdedit} in the supplementary material.

\subsection{Ablation Experiments}

We carry out all ablation studies by initially undertaking glyph-alignment pre-training, followed by training the Glyph-SDXL model on our graphic design benchmarks. Furthermore, all ablations are carried out on $100$K glyph image-text pairs for Glyph-ByT5 and Glyph-SDXL models respectively unless specified.

\vspace{1mm}
\noindent\textbf{Pre-trained Visual Encoder Choice}
We study the effect of choosing four different pre-trained visual encoders: CLIP visual encoder~\cite{radford2021learning}, DINOv2 ~\cite{darcet2023vitneedreg}, ViTSTR~\cite{atienza2021vision}, and CLIP4STR visual encoder~\cite{zhao2023clip4str}.
We report the detailed comparison results in Table~\ref{tab:effect_vit}. Notably, we also observe that accurate font type and color controls only occur when using DINOv2 as the pre-trained visual encoder.

\vspace{1mm}
\noindent\textbf{Loss Design}
We study the effect of choosing different training loss designs and report the detailed comparison results in Table~\ref{tab:effect_loss}. It is evident that the proposed box-level contrastive loss achieves favorable performance.

\begin{figure*}[t]
\centering
\begin{subfigure}[b]{0.16\textwidth}
{\includegraphics[width=\textwidth]{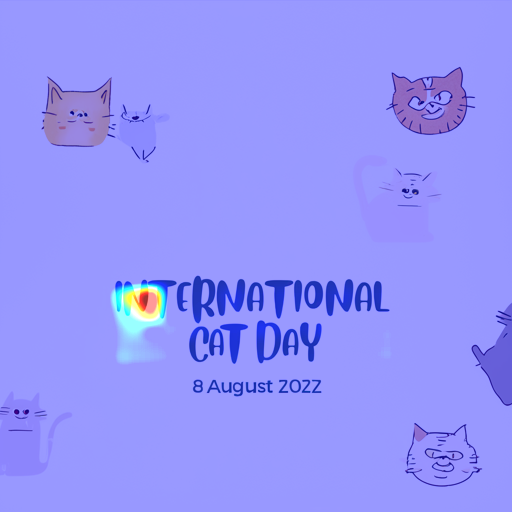}}
\caption*{\scriptsize{I\redtext{N}TERNATIONAL}}
\end{subfigure}
\hspace{5mm}
\begin{subfigure}[b]{0.16\textwidth}
{\includegraphics[width=\textwidth]{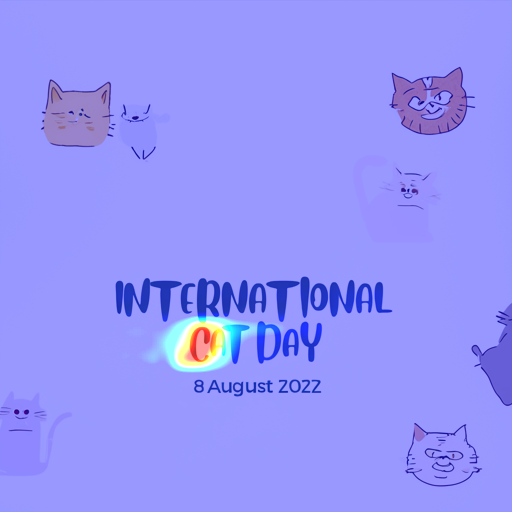}}
\caption*{\scriptsize{\redtext{C}AT}}
\end{subfigure}
\hspace{5mm}
\begin{subfigure}[b]{0.16\textwidth}
{\includegraphics[width=\textwidth]{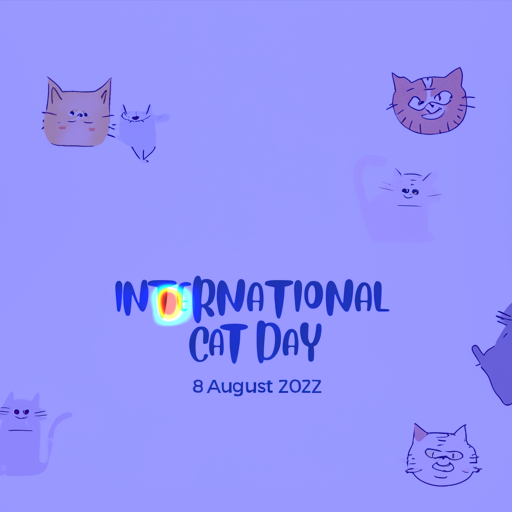}}
\caption*{\scriptsize{IN\redtext{T}ERNATIONAL}}
\end{subfigure}
\hspace{5mm}
\begin{subfigure}[b]{0.16\textwidth}
{\includegraphics[width=\textwidth]{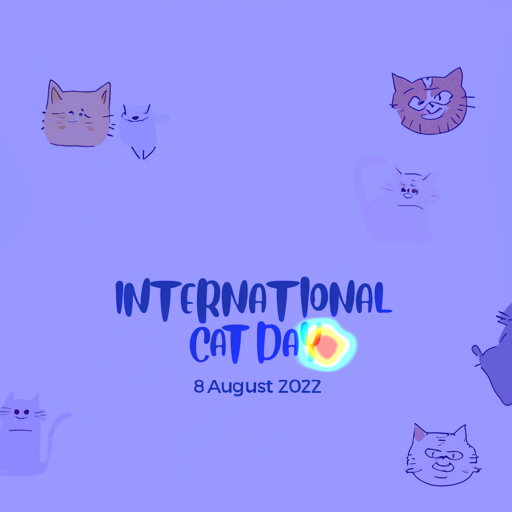}}
\caption*{\scriptsize{DA\redtext{Y}}}
\end{subfigure}
\hspace{5mm}
\begin{subfigure}[b]{0.16\textwidth}
{\includegraphics[width=\textwidth]{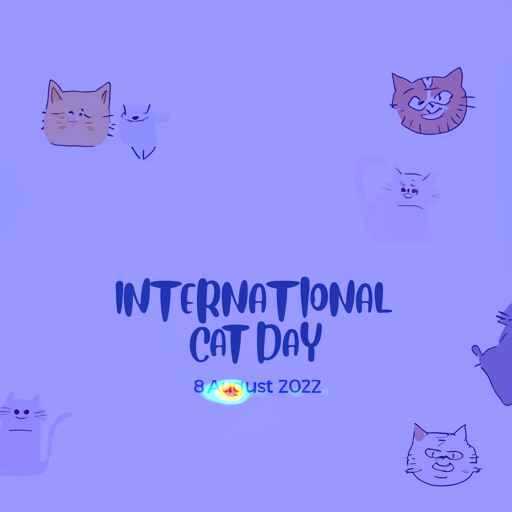}}
\caption*{\scriptsize{A\redtext{u}gust}}
    \end{subfigure}
\captionof{figure}{\small{Visualization of the cross attention maps within our Glyph-SDXL model. We show the heat maps between all pixels and the selected characters from the text box with `INTERNATIONAL CAT DAY' and `8 August 2022'.}}
\vspace{-5mm}
\label{fig:attn_vis}
\end{figure*}

\vspace{1mm}
\noindent\textbf{Glyph Augmentation} We study the effect of glyph augmentation during Glyph-Alignment pretraining. As indicated in Table~\ref{tab:effect_aug}, glyph augmentation provides a notable improvement compared with non-augmented settings, peaking at around 1:16.
Notably, we also observe that font-type and color control only occur when the ratio reaches or exceeds 1:16, also indicating its effectiveness.

\vspace{1mm}
\noindent\textbf{Mapper, Scaling Glyph-Text Dataset and Text Encoder Size, and More}
Table~\ref{tab:effect_sdxl_arch} shows the importance of using the ByT5-to-SDXL mapper to align the gap.
Table~\ref{tab:glyph_sdxl_data_scale} and Table~\ref{tab:glyph_sdxl_model_scale} verify the benefits of scaling up the glyph-text dataset size and text encoder size. We provide more ablations, as well as experiments of Glyph-SDXL-Scene in the appendix.

\vspace{1mm}
\noindent\textbf{Qualitative Analysis}
To gain a deeper understanding of how our Glyph-ByT5 excels at the visual text rendering task, we further visualize the cross-attention maps between glyph text prompts and rendered images, providing an example in Figure \ref{fig:attn_vis}. This visualization confirms that the diffusion model effectively utilizes the glyph-alignment prior encoded within our Glyph-ByT5 text encoder.

%% file: sec/5_conclusion.tex
\section{Conclusion}
\label{sec:conclusion}
This paper presents the design and training of the Glyph-ByT5 text encoder, tailored for accurate visual text rendering with diffusion models. Central to this endeavor are two key developments: the creation of a scalable, high-quality glyph-text dataset and the implementation of pre-training techniques for glyph-text alignment. These critical advancements efficiently bridge the gap between glyph imagery and text prompts, facilitating the generation of accurate text for both text-rich design images and open-domain images with scene text. The compelling performance achieved by our proposed Glyph-SDXL model suggests that the development of specialized text encoders represents a promising avenue for overcoming some of the fundamental challenges associated with diffusion models, indicating a significant trend in the domain.

%% file: sec/6_appendix.tex
\clearpage
\setcounter{page}{1}
\maketitlesupplementary

\subsection*{A. Dataset Statistics} 
\begin{figure*}[!t]
\begin{minipage}[!t]{1\linewidth}
\begin{subfigure}[b]{0.16\textwidth}
\centering
\includegraphics[width=0.99\textwidth]{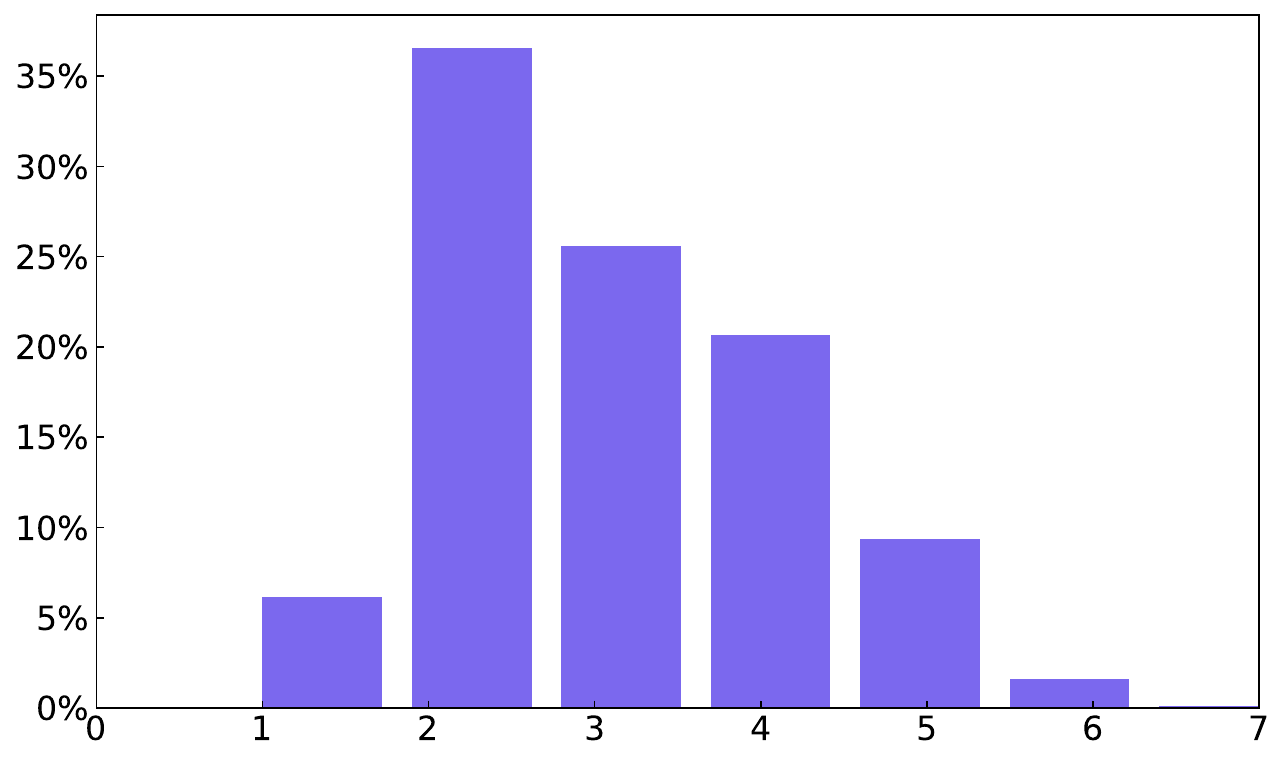}
\vspace{-1mm}
\caption{}
\end{subfigure}
\begin{subfigure}[b]{0.16\textwidth}
\centering
\includegraphics[width=0.99\textwidth]{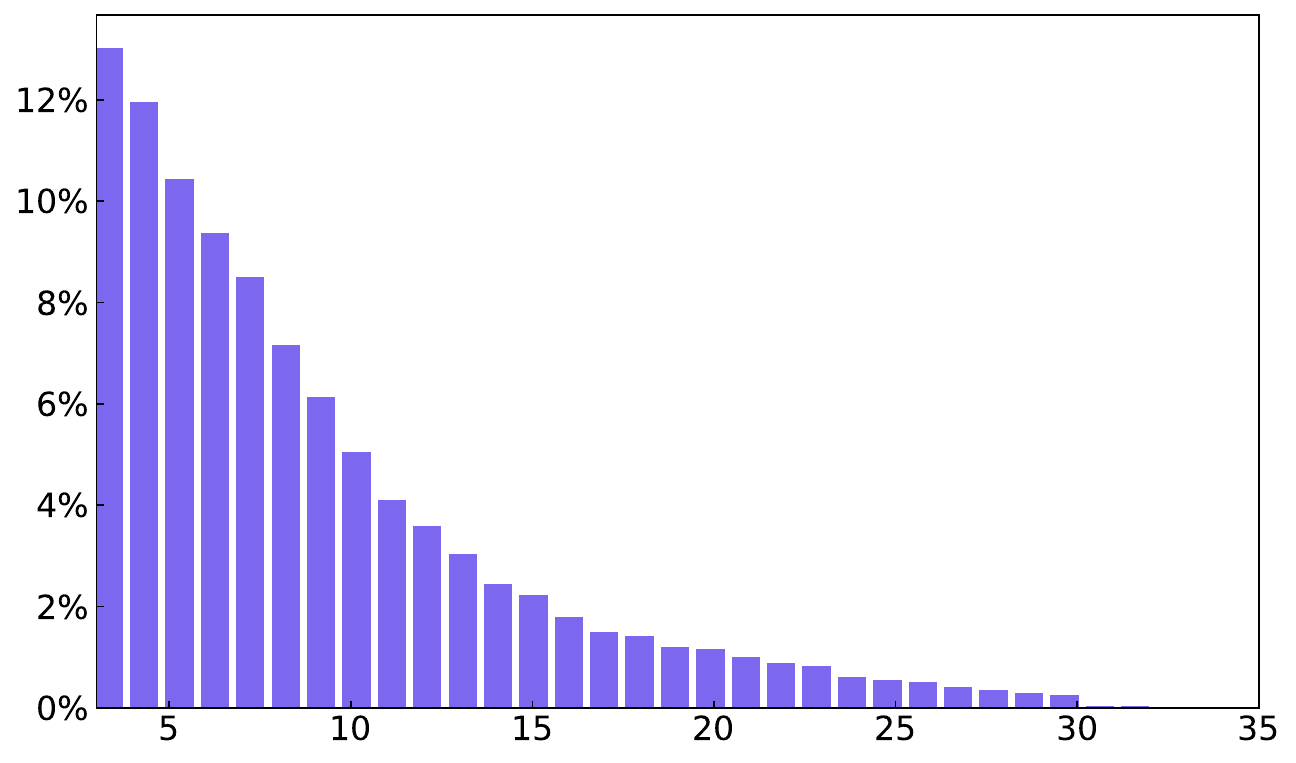}
\vspace{-1mm}
\caption{}
\end{subfigure}
\begin{subfigure}[b]{0.16\textwidth}
\centering
\includegraphics[width=0.99\textwidth]{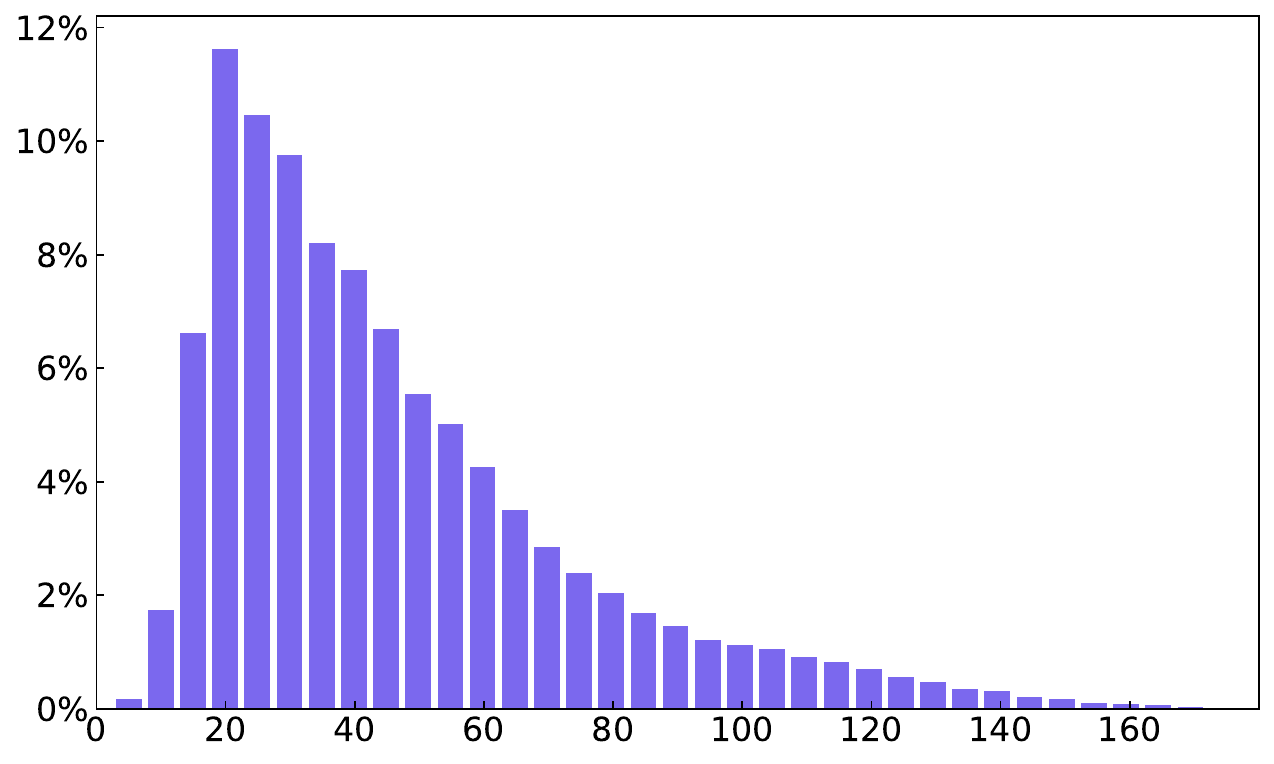}
\vspace{-1mm}
\caption{}
\end{subfigure}
\begin{subfigure}[b]{0.16\textwidth}
\centering
\includegraphics[width=0.99\textwidth]{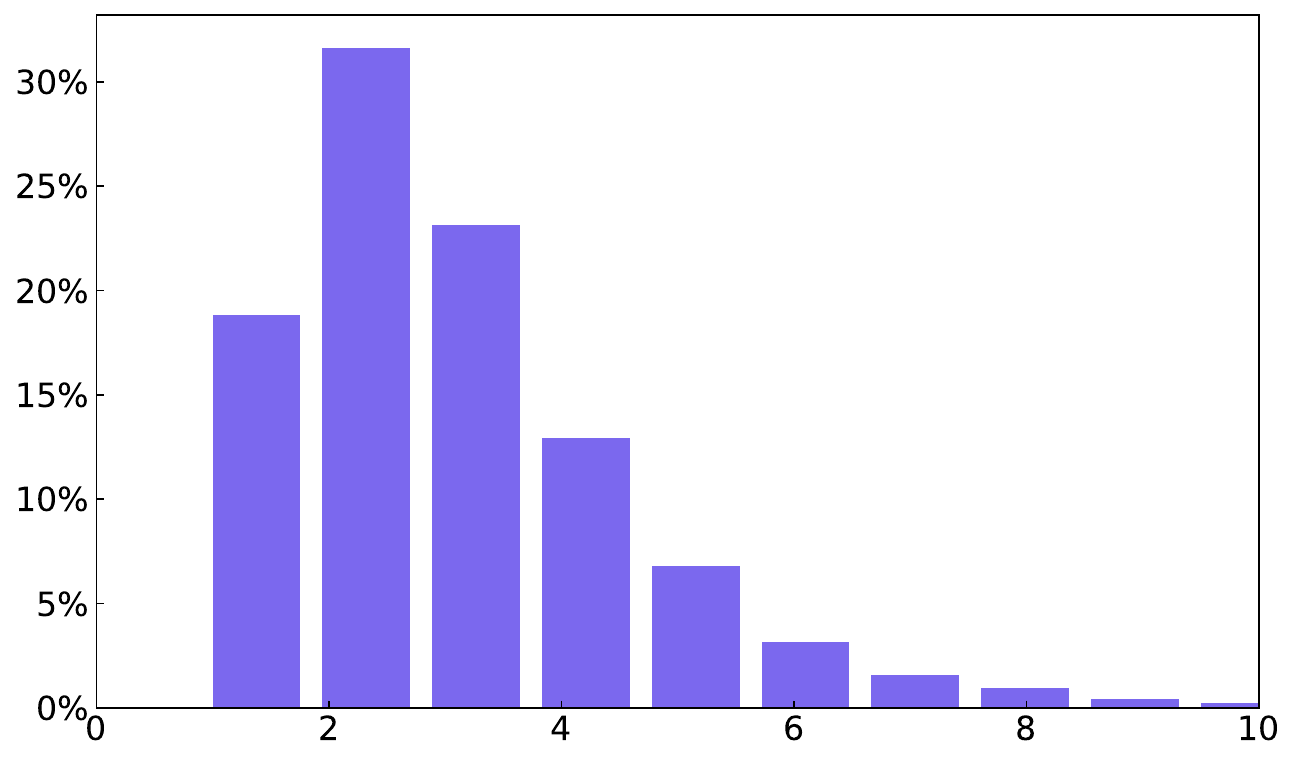}
\vspace{-1mm}
\caption{}
\end{subfigure}
\begin{subfigure}[b]{0.16\textwidth}
\centering
\includegraphics[width=0.99\textwidth]{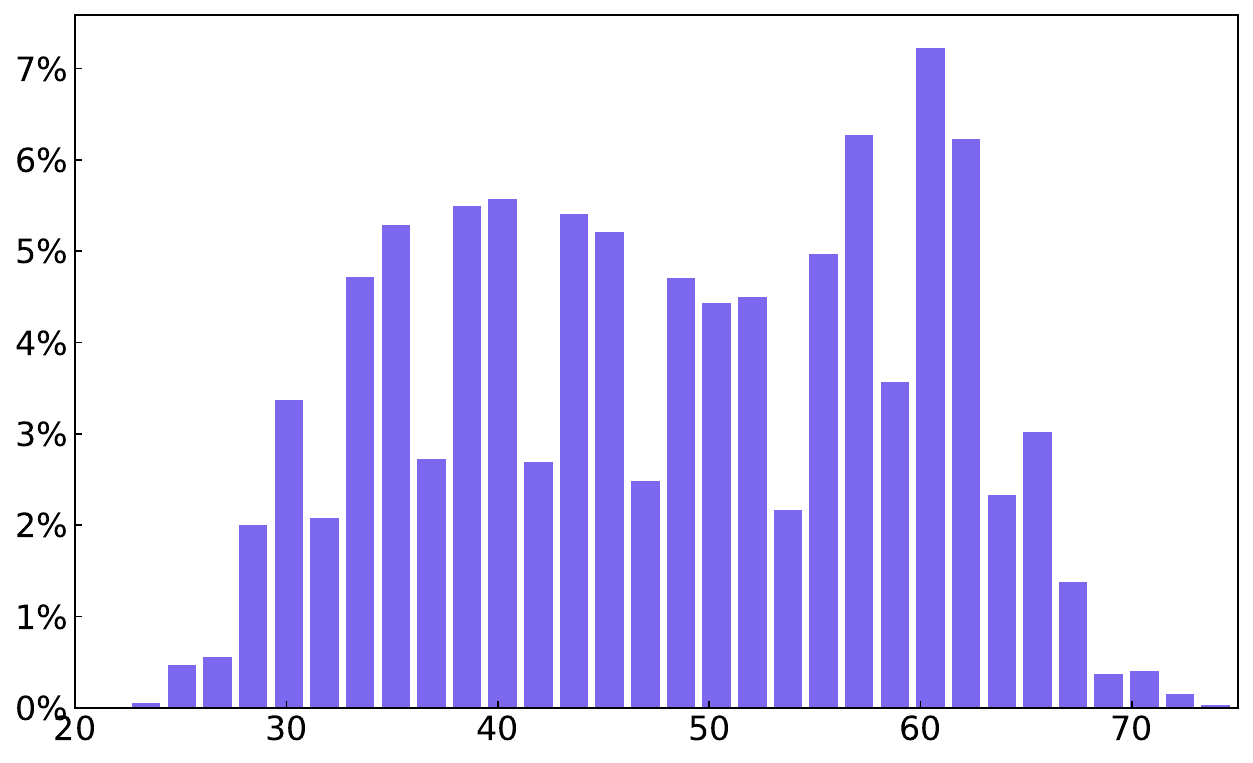}
\vspace{-1mm}
\caption{}
\end{subfigure}
\begin{subfigure}[b]{0.16\textwidth}
\centering
\includegraphics[width=0.99\textwidth]{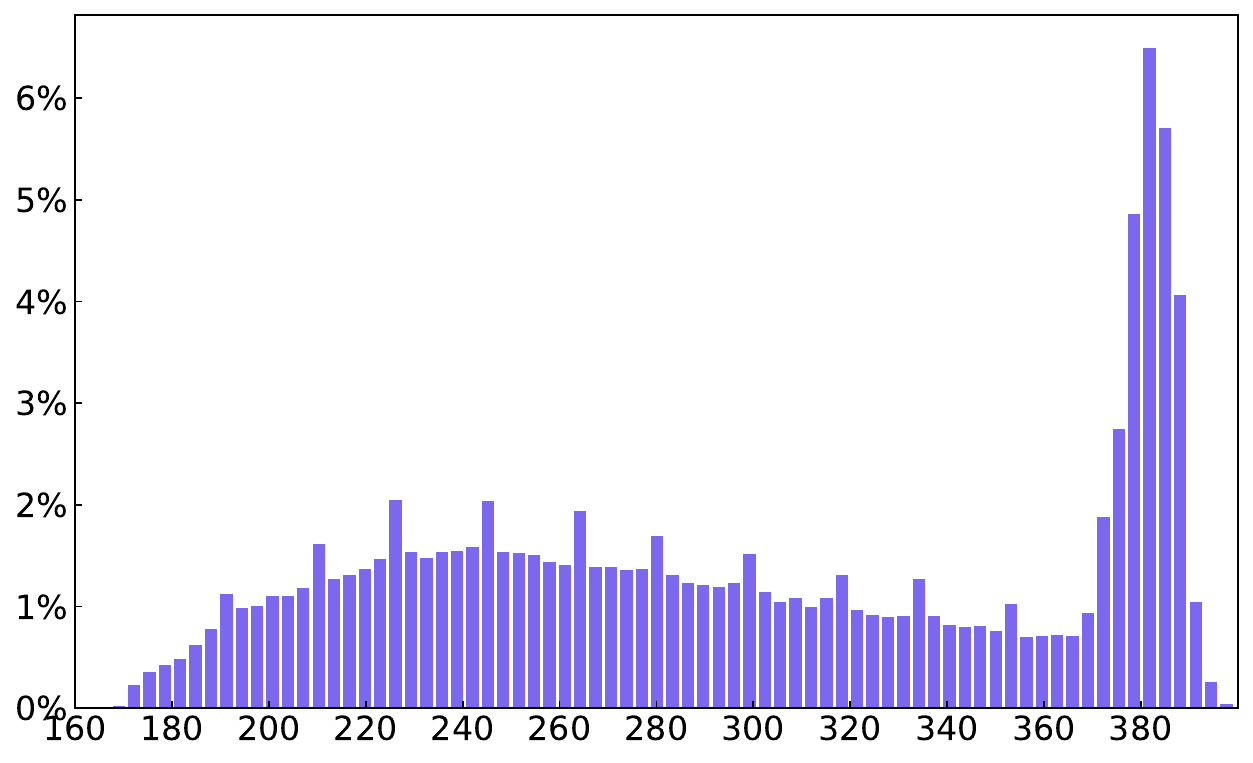}
\vspace{-1mm}
\caption{}
\end{subfigure}
\vspace{-1mm}
\end{minipage}
\vspace{-1mm}
\caption{\small{
Illustrating the statistics of our Glyph-Text Dataset and Paragraph-Glyph-Text Dataset
For Glyph-Text Dataset: (a) \# of text boxes, (b) \# of words, (c) \# of characters.
For Paragraph-Glyph-Text Dataset: (d) \# of text boxes, (e) \# of words, (f) \# of characters.
We can see that the glyph images of Paragraph-Glyph-Text consist of a much larger number of words and characters.
}}
\label{fig:statistics}
\vspace{-5mm}
\end{figure*}

We present statistics on the number of text boxes, words, and characters across the entire glyph-text dataset and the paragraph-glyph-text dataset in Figure~\ref{fig:statistics}.

\subsection*{B. Training Settings}

Table \ref{table:clip_hparam} and Table \ref{table:diffusion_hparam} detail the hyperparameters for training Glyph-CLIP and Glyph-SDXL, respectively. Glyph-CLIP is trained with $4\times$A100 GPUs while Glyph-SDXL is trained with $32 \times$ MI200 GPUs.

\begin{table}[!t]
\begin{minipage}[t]{1\linewidth}
\centering
\tablestyle{1pt}{1.5}
\resizebox{1.0\linewidth}{!}
{
\begin{tabular}{l|c|c|c}
Hyperparameter &            Glyph-CLIP-Small &                Glyph-CLIP-Base &                Glyph-CLIP-Large  \\
\shline
Text Encoder &    ByT5-Small &                ByT5-Base &                ByT5-Large \\
Vision Encoder &     DINOv2 ViT-B/14 &            DINOv2 ViT-L/14 &            DINOv2 ViT-g/14 \\
Peak Learning-rate &    5.00E-04 &            5.00E-04 &            5.00E-04 \\
Batch Size & 1536 &  1024 &   768 \\
Epochs &  5 &   5 &    5 \\
Warmup Iterations &  100 &  100 &  100  \\
Weight Decay &     0.2 &                0.2 &                0.2  \\
Text-Encoder Dropout &      0.1 &      0.1 &         0.1 \\
\end{tabular}
}
\vspace{-2mm}
\caption{\footnotesize Glyph-ByT5 pre-training hyper-parameters.}
\label{table:clip_hparam}
\end{minipage}
\vspace{-2mm}
\end{table}

\begin{table}[!t]
\begin{minipage}[t]{1\linewidth}
\centering
\tablestyle{1pt}{1.5}
\resizebox{1.0\linewidth}{!}
{
\begin{tabular}{l|c|c|c}
Hyperparameter     &        Glyph-SDXL-Small &                Glyph-SDXL-Base &                Glyph-SDXL-Large  \\
\shline
Text Encoder  &    Glyph-ByT5-Small &                Glyph-ByT5-Base &                Glyph-ByT5-Large \\
UNet Learning-rate &            5.00E-05 &            5.00E-05 &            5.00E-05  \\
Text Enoder Learning-rate &            1.00E-04 &            1.00E-04 &            1.00E-04  \\
Batch Size &                256 &                256 &                256 \\
Epochs &                  10 &                  10 &                  10  \\
Weight Decay &                0.01 &                0.01 &                0.01  \\
Text-Encoder Weight Decay &                0.2 &                0.2 &                0.2 \\
Text-Encoder Dropout &                   0.1 &                   0.1 &                   0.1 \\
Gradient Clipping &                   1.0 &                   1.0 &                   1.0 \\
\end{tabular}
}
\vspace{-2mm}
\caption{\footnotesize Glyph-SDXL model training hyper-parameters.}
\label{table:diffusion_hparam}
\end{minipage}

\begin{minipage}[t]{1\linewidth}
\centering
\tablestyle{9pt}{1.5}
\setlength{\tabcolsep}{4pt}
\resizebox{1.0\linewidth}{!}
{
\begin{tabular}{l|cccc}
\multirow{2}{*}{Method} &  \multicolumn{4}{c}{Precision ($\%$)}  \\\cline{2-5}
& $\le$20 chars & $\le$20-50 chars & $\le$50-100 chars & $\ge$100 chars \\
\shline
GlyphControl-SDXL & $65.17$ & $61.04$ & $45.31$ & $32.75$ \\
Glyph-SDXL & $\bf{84.54}$ & $\bf{84.56}$ & $\bf{79.89}$ & $\bf{73.29}$ \\
\end{tabular}
}
\caption{
\small{Comparison with ControlNet style model.}}
\label{tab:compare_controlnet}
\end{minipage}
\vspace{-1mm}
\begin{minipage}[t]{1\linewidth}
\centering
\tablestyle{9pt}{1.5}
\setlength{\tabcolsep}{4pt}
\resizebox{1.0\linewidth}{!}
{
\begin{tabular}{l|cccc}
\multirow{2}{*}{Visual Encoder} &  \multicolumn{4}{c}{Precision ($\%$)}  \\\cline{2-5}
& $\le$20 chars & $\le$20-50 chars & $\le$50-100 chars & $\ge$100 chars \\
\shline 
SDXL-VAE & $70.02$ & $70.57$ & $66.34$ & $56.28$ \\
DINOv2 & $\bf{84.54}$ & $\bf{84.56}$ & $\bf{79.89}$ & $\bf{73.29}$ \\
\end{tabular}
}
\caption{
\small{Comparison with aligning to the latent space of SDXL.}}
\label{tab:compare_vae}
\end{minipage}

\begin{minipage}[t]{1\linewidth}
\centering
\tablestyle{9pt}{1.5}
\setlength{\tabcolsep}{4pt}
\resizebox{1.0\linewidth}{!}
{
\begin{tabular}{l|cccc}
\multirow{2}{*}{Text encoder fusion method}  &  \multicolumn{4}{c}{Precision ($\%$)}  \\\cline{2-5}
& $\le$20 chars & $\le$20-50 chars & $\le$50-100 chars & $\ge$100 chars \\
\shline
concatnate text embeddings & $1.47$ & $2.5$ & $2.77$ & $3.91$ \\
region-wise cross-attention & $\bf{84.54}$ & $\bf{84.56}$ & $\bf{79.89}$ & $\bf{73.29}$ \\
\end{tabular}
\vspace{-1mm}
}
\caption{
\small{Effect of the region-wise multihead cross-attention mechanism.}}
\label{tab:effect_regionwise}
\end{minipage}
\vspace{-2mm}

\end{table}

\subsection*{C. Typography Editing with Region-wise SDEdit}

Inspired by the success of SDEdit \cite{meng2021sdedit} and Blended Latent Diffusion~\cite{avrahami2023blended}, we introduce a region-wise SDEdit scheme, transforming our Glyph-SDXL into a precise and adaptable visual text editor. This enables the refinement of visual text in high-quality images produced by state-of-the-art (SOTA) generation models, such as \dalle.
The typography editing outcomes are displayed in Figure \ref{fig:sdedit}, showcasing our approach's robust capability for precise typography editing.

\begin{figure*}[t]
\begin{minipage}[t]{1\linewidth}
\begin{subfigure}[b]{1\textwidth}
\centering
\vspace{3mm}
\includegraphics[width=1\textwidth]{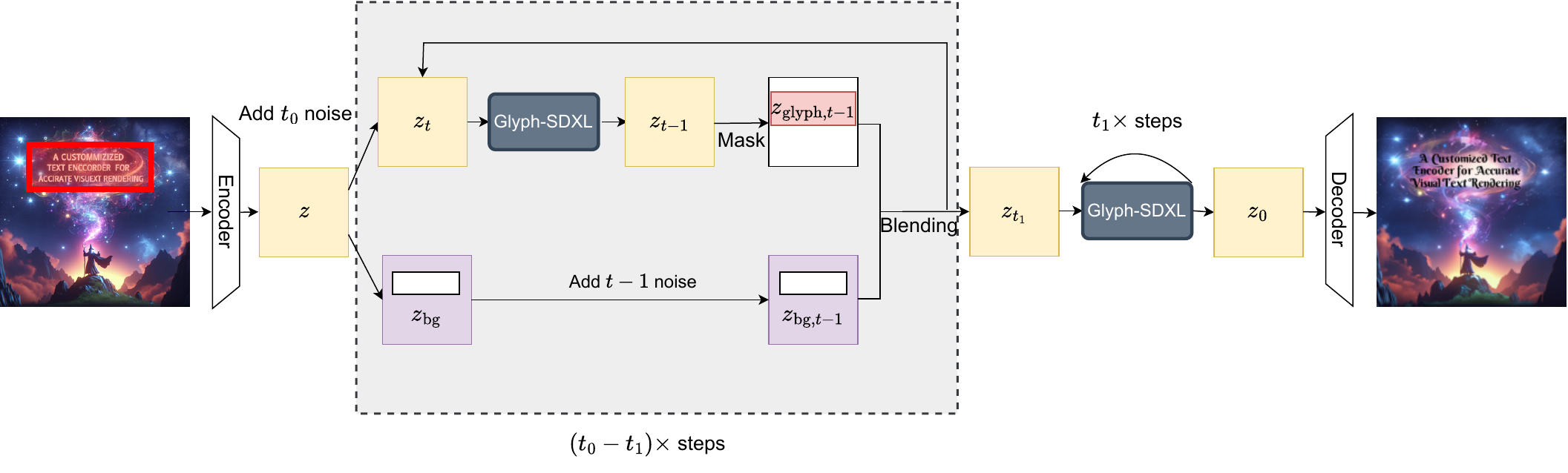}
\vspace{-8mm}
\end{subfigure}
\caption{\small{Illustrating the Region-wise SDEdit pipeline.}}
\label{fig:region_wise_sdedit}
\end{minipage}
\end{figure*}

\begin{figure*}[t]
\begin{minipage}[t]{1\linewidth}
\centering
\begin{subfigure}[b]{0.16\textwidth}
\includegraphics[width=\textwidth]{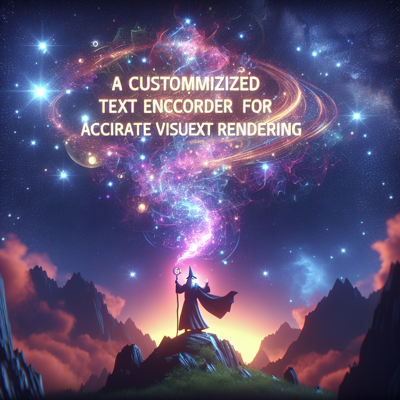}
\vspace{-3mm}
\end{subfigure}
\begin{subfigure}[b]{0.16\textwidth}
{\includegraphics[width=\textwidth]{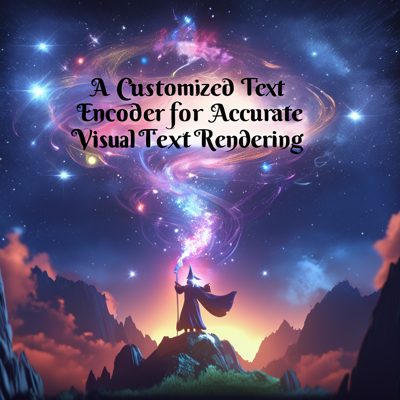}}
\vspace{-3mm}
\end{subfigure}
\begin{subfigure}[b]{0.16\textwidth}
{\includegraphics[width=\textwidth]{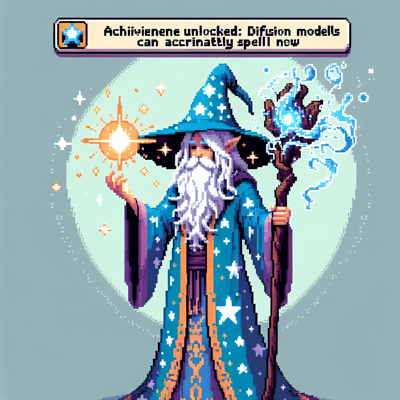}}
\vspace{-3mm}
\end{subfigure}
\begin{subfigure}[b]{0.16\textwidth}
{\includegraphics[width=\textwidth]{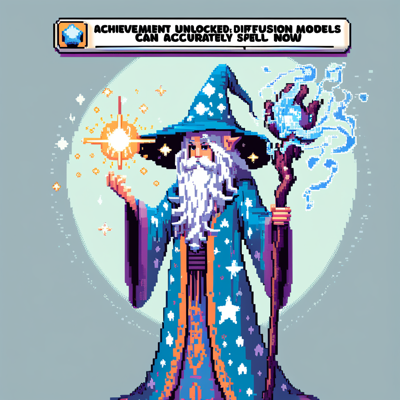}}
\vspace{-3mm}
\end{subfigure}
\begin{subfigure}[b]{0.16\textwidth}
{\includegraphics[width=\textwidth]{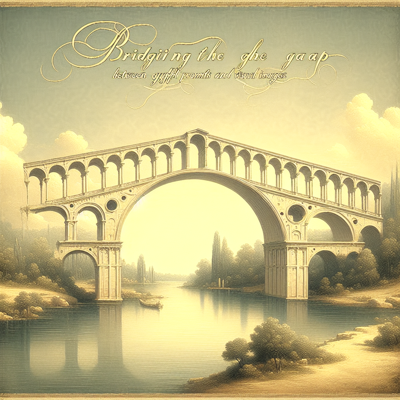}}
\vspace{-3mm}
\end{subfigure}
\begin{subfigure}[b]{0.16\textwidth}
{\includegraphics[width=\textwidth]{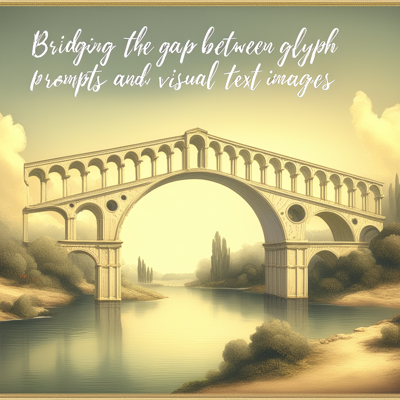}}
\vspace{-3mm}
\end{subfigure} \\
\begin{subfigure}[b]{0.16\textwidth}
\includegraphics[width=\textwidth]{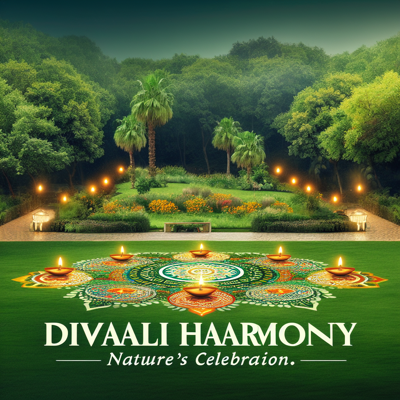}
\vspace{-3mm}
\end{subfigure}
\begin{subfigure}[b]{0.16\textwidth}
{\includegraphics[width=\textwidth]{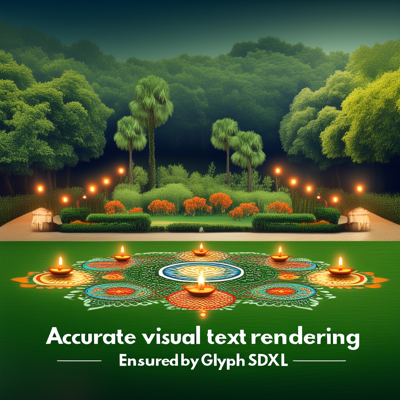}}
\vspace{-3mm}
\end{subfigure}
\begin{subfigure}[b]{0.16\textwidth}
{\includegraphics[width=\textwidth]{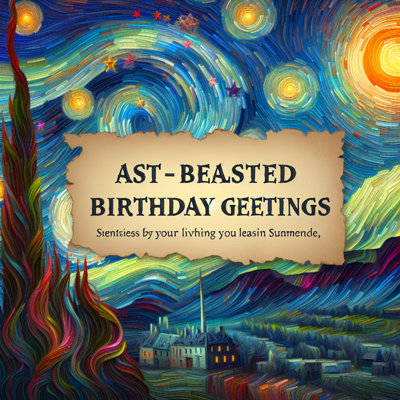}}
\vspace{-3mm}
\end{subfigure}
\begin{subfigure}[b]{0.16\textwidth}
{\includegraphics[width=\textwidth]{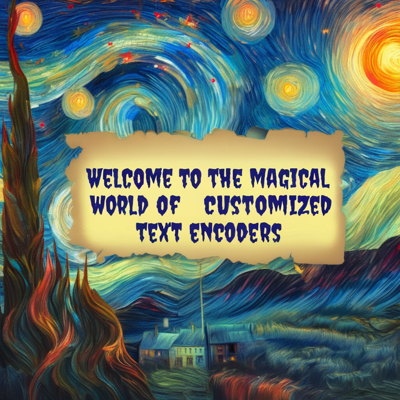}}
\vspace{-3mm}
\end{subfigure}
\begin{subfigure}[b]{0.16\textwidth}
{\includegraphics[width=\textwidth]{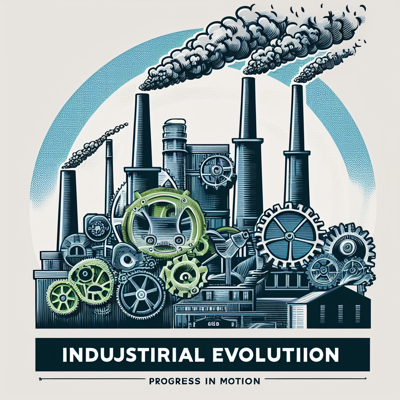}}
\vspace{-3mm}
\end{subfigure}
\begin{subfigure}[b]{0.16\textwidth}
{\includegraphics[width=\textwidth]{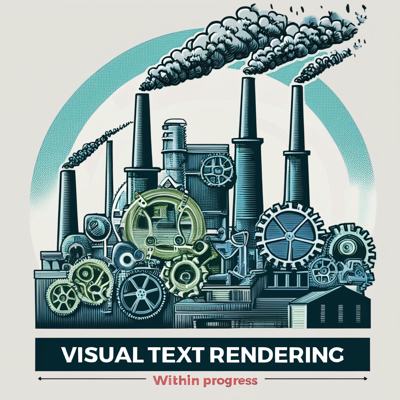}}
\vspace{-3mm}
\end{subfigure}
\caption{\small{\textbf{Illustrating typography editing results based on our Glyph-SDXL}. Original images generated by \dalle and images edited by Glyph-SDXL are illustrated in the 1, 3, 5 rows, and 2, 4, 6 rows respectively.}}
\label{fig:sdedit}
\end{minipage}

\begin{minipage}[t]{1\linewidth}
\centering
\begin{subfigure}[b]{0.16\textwidth}
\includegraphics[width=\textwidth]{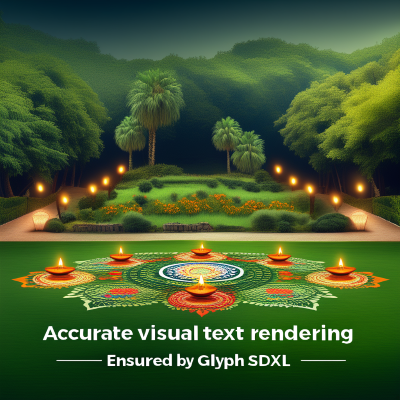}
\vspace{-3mm}
\end{subfigure}
\begin{subfigure}[b]{0.16\textwidth}
{\includegraphics[width=\textwidth]{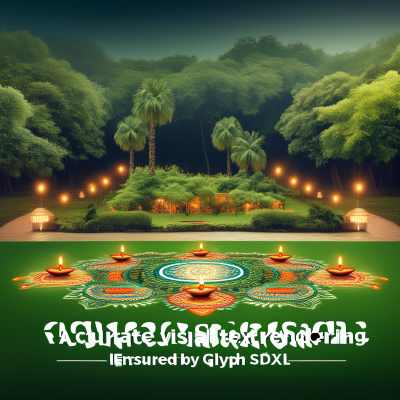}}
\vspace{-3mm}
\end{subfigure}
\begin{subfigure}[b]{0.16\textwidth}
{\includegraphics[width=\textwidth]{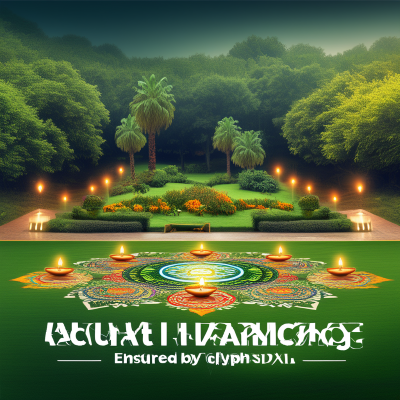}}
\vspace{-3mm}
\end{subfigure}
\begin{subfigure}[b]{0.16\textwidth}
{\includegraphics[width=\textwidth]{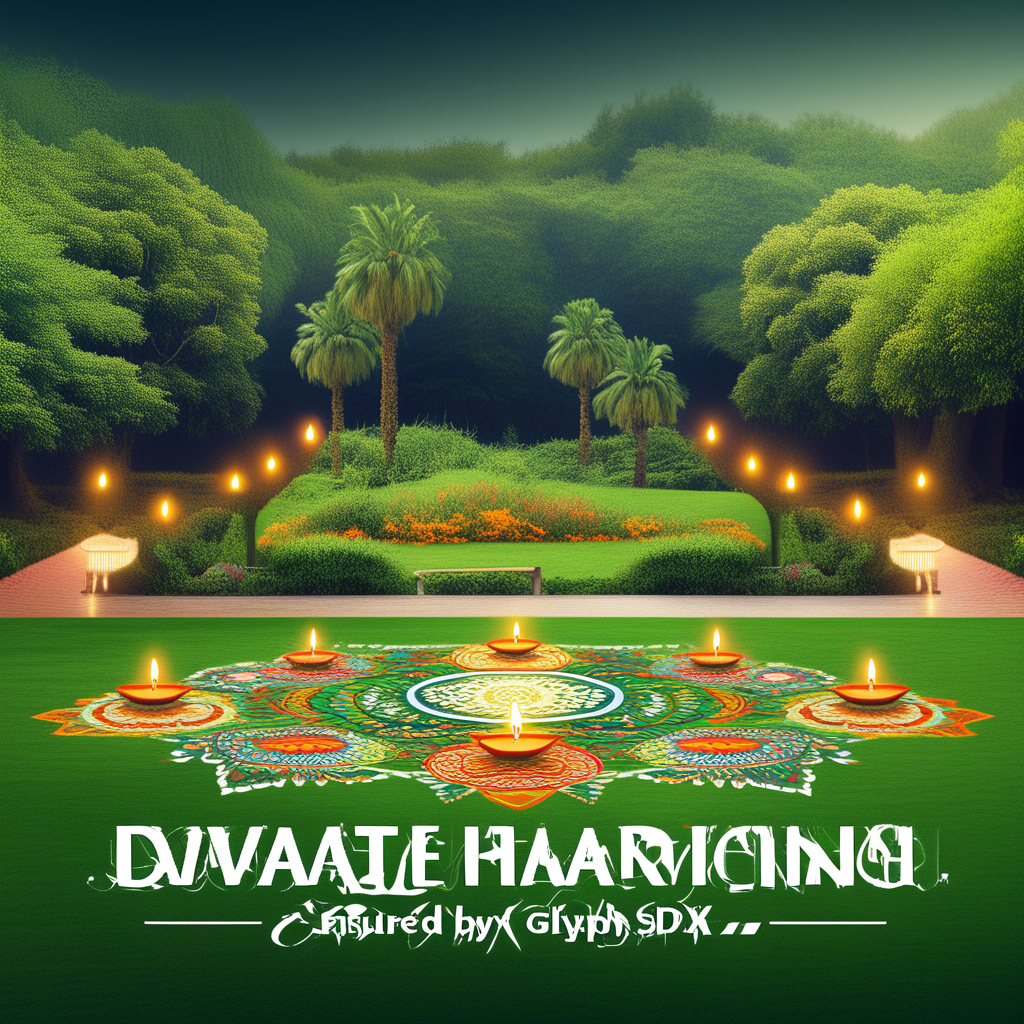}}
\vspace{-3mm}
\end{subfigure}
\begin{subfigure}[b]{0.16\textwidth}
{\includegraphics[width=\textwidth]{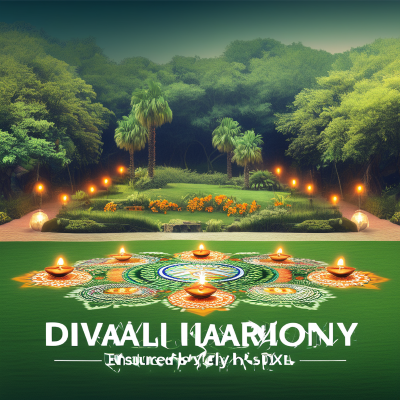}}
\vspace{-3mm}
\end{subfigure}
\begin{subfigure}[b]{0.16\textwidth}
{\includegraphics[width=\textwidth]{img/edit/dalle_4.png}}
\vspace{-3mm}
\end{subfigure} \\
\begin{subfigure}[b]{0.16\textwidth}
\includegraphics[width=\textwidth]{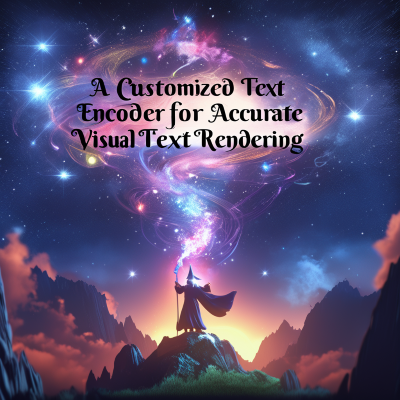}
\caption{\scriptsize{$t_0 = 800$}}
\vspace{-3mm}
\end{subfigure}
\begin{subfigure}[b]{0.16\textwidth}
{\includegraphics[width=\textwidth]{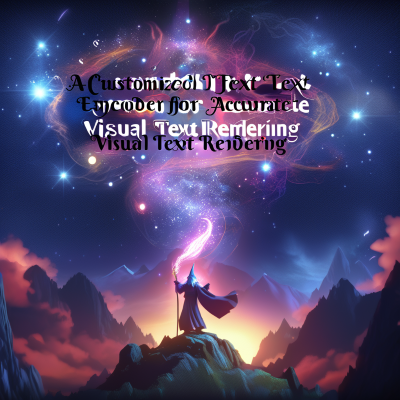}}
\caption{\scriptsize{$t_0 = 700$}}
\vspace{-3mm}
\end{subfigure}
\begin{subfigure}[b]{0.16\textwidth}
{\includegraphics[width=\textwidth]{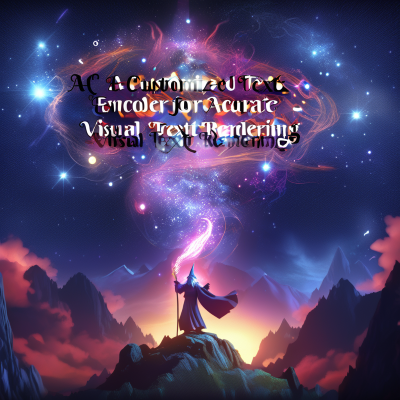}}
\caption{\scriptsize{$t_0 = 600$}}
\vspace{-3mm}
\end{subfigure}
\begin{subfigure}[b]{0.16\textwidth}
{\includegraphics[width=\textwidth]{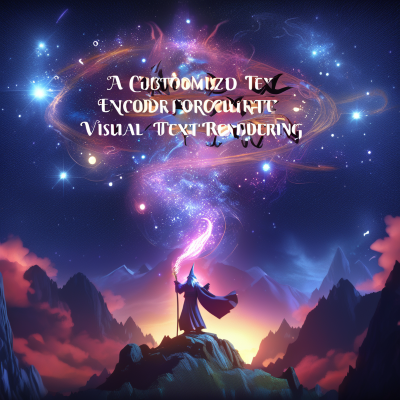}}
\caption{\scriptsize{$t_0 = 500$}}
\vspace{-3mm}
\end{subfigure}
\begin{subfigure}[b]{0.16\textwidth}
{\includegraphics[width=\textwidth]{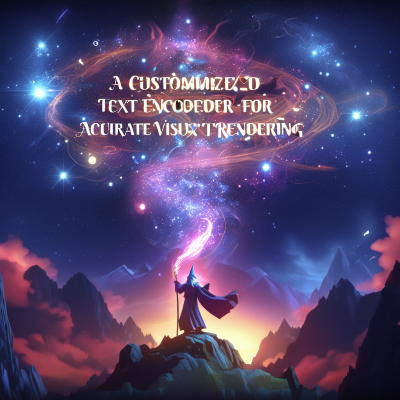}}
\caption{\scriptsize{$t_0 = 400$}}
\vspace{-3mm}
\end{subfigure}
\begin{subfigure}[b]{0.16\textwidth}
{\includegraphics[width=\textwidth]{img/edit/dalle_1.png}}
\caption{\scriptsize{original}}
\vspace{-3mm}
\end{subfigure} 
\caption{\small{Effect of different choices of $t_0$ for Region-wise SDEdit.}}
\label{fig:sdedit_effect_t0}
\end{minipage}

\begin{minipage}[t]{1\linewidth}
\centering
\begin{subfigure}[b]{0.16\textwidth}
\includegraphics[width=\textwidth]{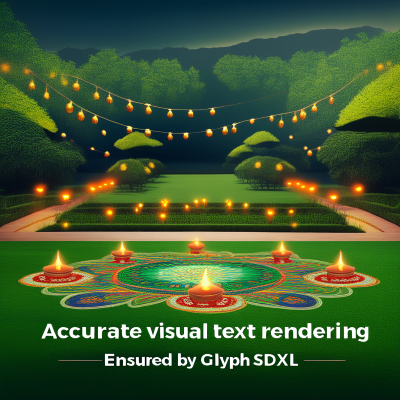}
\vspace{-3mm}
\end{subfigure}
\begin{subfigure}[b]{0.16\textwidth}
{\includegraphics[width=\textwidth]{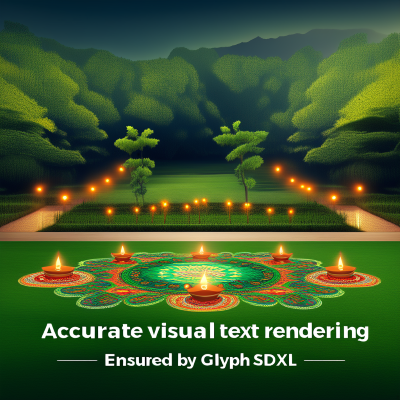}}
\vspace{-3mm}
\end{subfigure}
\begin{subfigure}[b]{0.16\textwidth}
{\includegraphics[width=\textwidth]{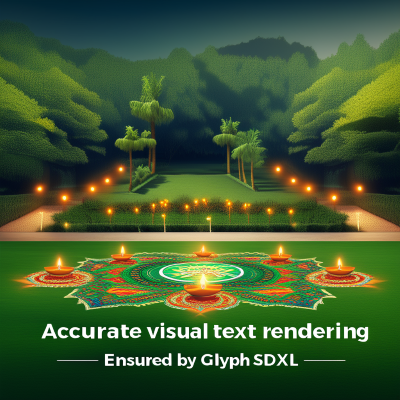}}
\vspace{-3mm}
\end{subfigure}
\begin{subfigure}[b]{0.16\textwidth}
{\includegraphics[width=\textwidth]{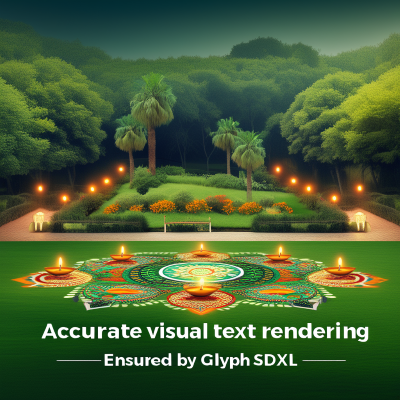}}
\vspace{-3mm}
\end{subfigure}
\begin{subfigure}[b]{0.16\textwidth}
{\includegraphics[width=\textwidth]{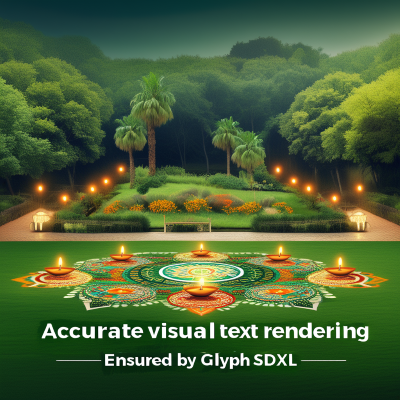}}
\vspace{-3mm}
\end{subfigure}
\begin{subfigure}[b]{0.16\textwidth}
{\includegraphics[width=\textwidth]{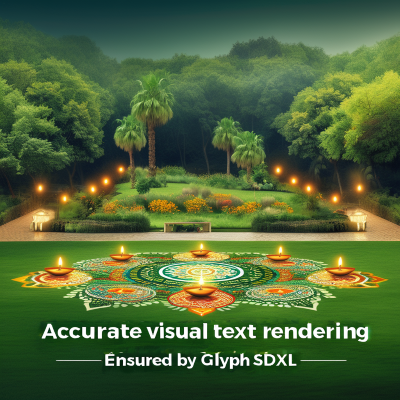}}
\vspace{-3mm}
\end{subfigure} \\
\begin{subfigure}[b]{0.16\textwidth}
\includegraphics[width=\textwidth]{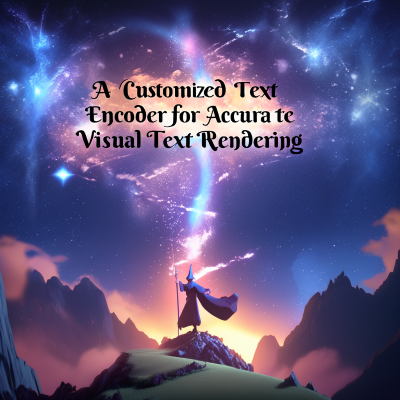}
\caption{\scriptsize{$t_1 = 600$}}
\vspace{-3mm}
\end{subfigure}
\begin{subfigure}[b]{0.16\textwidth}
{\includegraphics[width=\textwidth]{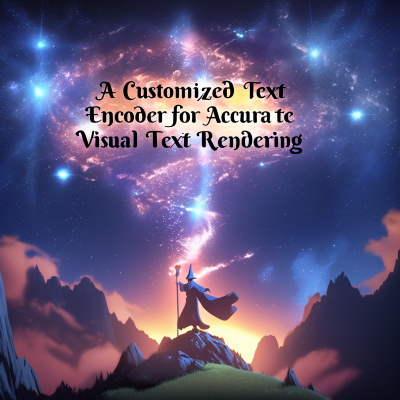}}
\caption{\scriptsize{$t_1 = 500$}}
\vspace{-3mm}
\end{subfigure}
\begin{subfigure}[b]{0.16\textwidth}
{\includegraphics[width=\textwidth]{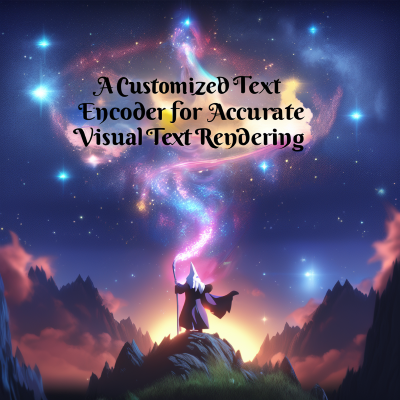}}
\caption{\scriptsize{$t_1 = 400$}}
\vspace{-3mm}
\end{subfigure}
\begin{subfigure}[b]{0.16\textwidth}
{\includegraphics[width=\textwidth]{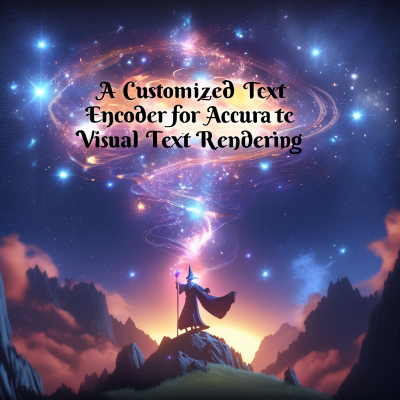}}
\caption{\scriptsize{$t_1 = 300$}}
\vspace{-3mm}
\end{subfigure}
\begin{subfigure}[b]{0.16\textwidth}
{\includegraphics[width=\textwidth]{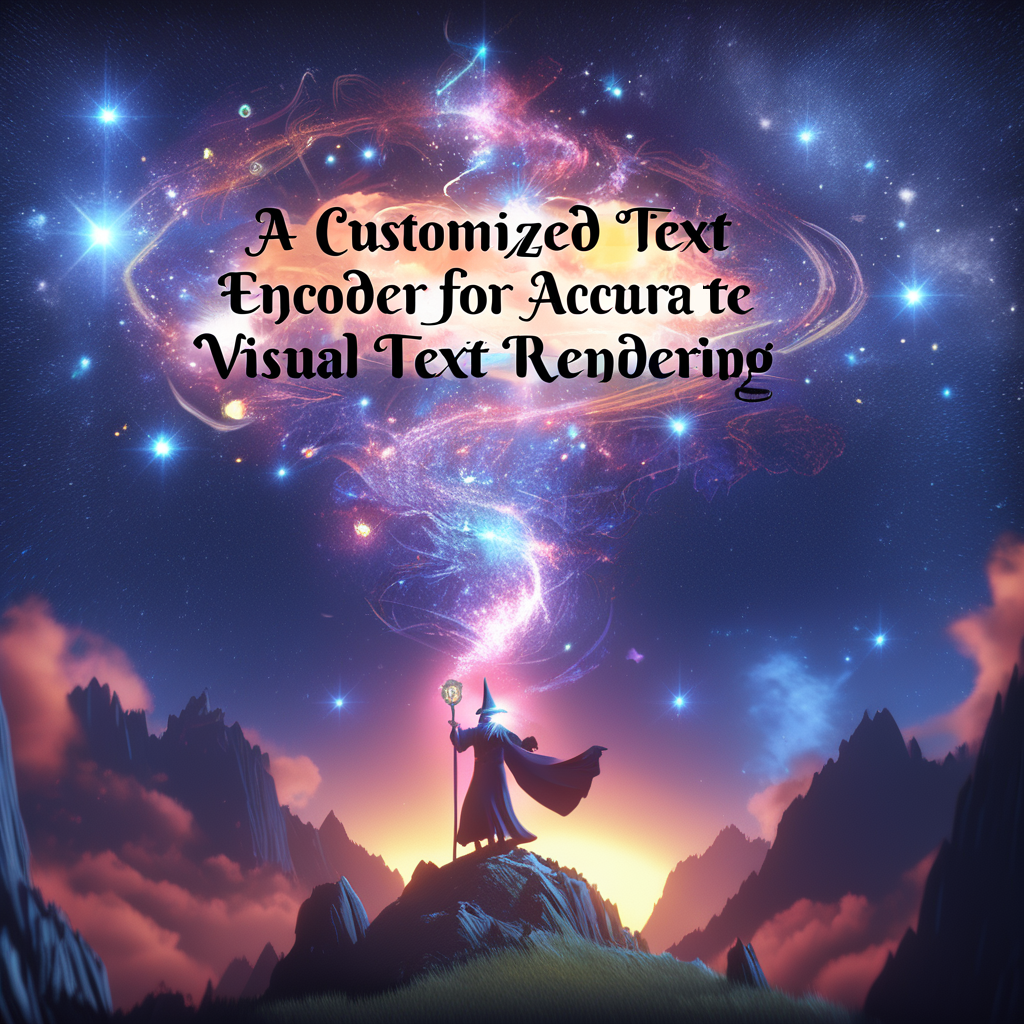}}
\caption{\scriptsize{$t_1 = 200$}}
\vspace{-3mm}
\end{subfigure}
\begin{subfigure}[b]{0.16\textwidth}
{\includegraphics[width=\textwidth]{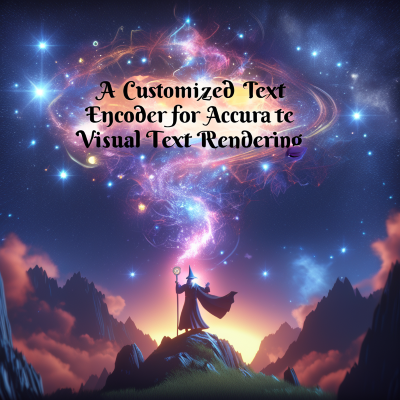}}
\caption{\scriptsize{$t_1 = 100$}}
\vspace{-3mm}
\end{subfigure} 
\caption{\small{Effect of different choices of $t_1$ for Region-wise SDEdit.}}
\label{fig:sdedit_effect_t1}
\end{minipage}
\end{figure*}

\vspace{1mm}
\noindent \textbf{Region-wise SDEdit Scheme:} For any given input image, $t_0$ steps of noise are initially added. Beginning at $t_0$, the Glyph-SDXL model is employed iteratively on the noised image to perform denoising. To ensure modifications are confined exclusively to glyph pixels—thereby keeping background pixels untouched—only glyph areas undergo denoising throughout this phase. The process progresses until timestep $t_1$, at which point the entire image undergoes denoising to guarantee overall coherence. Figure~\ref{fig:region_wise_sdedit} depicts the framework of our region-wise SDEdit scheme. 

\noindent \textbf{Effect of parameter choice:} We study the effect of different choices for $t_0$ and $t_1$.

We first fix $t_1 = 300$ and study the effect of different $t_0$. As illustrated in Figure \ref{fig:sdedit_effect_t0}, a large $t_0$ is crucial to ensure that the glyph latents are fully edited. Smaller $t_0$ keeps a larger proportion of the original latents, resulting in conflicts and degrading performance. 

Furthermore, we fix $t_0 = 800$ and study the effect of different choices of $t_1$. As illustrated in Figure \ref{fig:sdedit_effect_t1}, larger $t_1$ ensures better coherence between the pixels inside and outside the glyph boxes, but significantly changes the background image. Smaller $t_1$, on the contrary, maintains the background image while sacrificing coherence.

\subsection*{D. Ablation on the Design-to-Scene Alignment}

\begin{figure*}[t]
\begin{minipage}[t]{1\linewidth}
\centering
\begin{subfigure}[b]{0.16\textwidth}
\includegraphics[width=\textwidth]{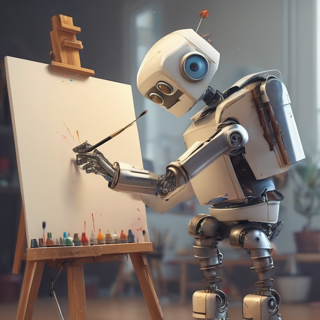}
\vspace{-3mm}
\end{subfigure}
\begin{subfigure}[b]{0.16\textwidth}
\includegraphics[width=\textwidth]{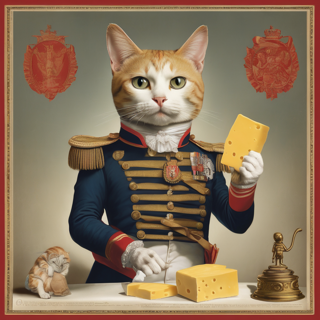}
\vspace{-3mm}
\end{subfigure}
\begin{subfigure}[b]{0.16\textwidth}
\includegraphics[width=\textwidth]{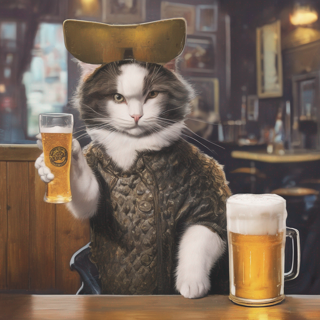}
\vspace{-3mm}
\end{subfigure}
\begin{subfigure}[b]{0.16\textwidth}
\includegraphics[width=\textwidth]{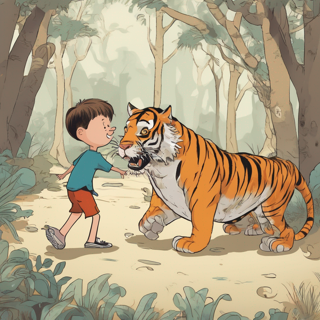}
\vspace{-3mm}
\end{subfigure}
\begin{subfigure}[b]{0.16\textwidth}
\includegraphics[width=\textwidth]{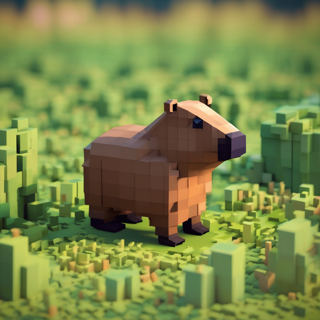}
\vspace{-3mm}
\end{subfigure}
\begin{subfigure}[b]{0.16\textwidth}
\includegraphics[width=\textwidth]{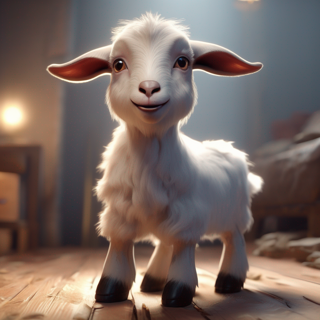}
\vspace{-3mm}
\end{subfigure}\\

\begin{subfigure}[b]{0.16\textwidth}
\includegraphics[width=\textwidth]{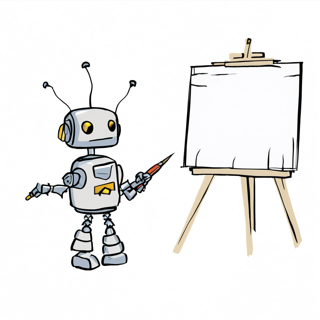}
\vspace{-3mm}
\end{subfigure}
\begin{subfigure}[b]{0.16\textwidth}
\includegraphics[width=\textwidth]{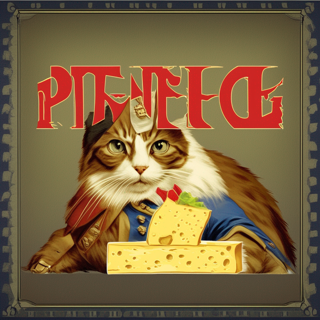}
\vspace{-3mm}
\end{subfigure}
\begin{subfigure}[b]{0.16\textwidth}
\includegraphics[width=\textwidth]{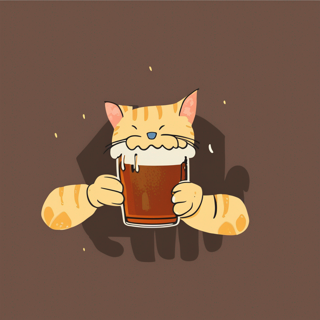}
\vspace{-3mm}
\end{subfigure}
\begin{subfigure}[b]{0.16\textwidth}
\includegraphics[width=\textwidth]{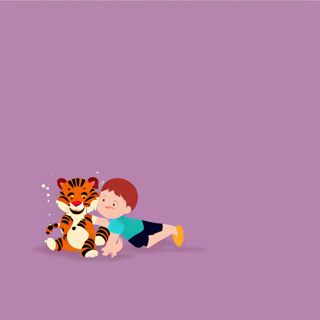}
\vspace{-3mm}
\end{subfigure}
\begin{subfigure}[b]{0.16\textwidth}
\includegraphics[width=\textwidth]{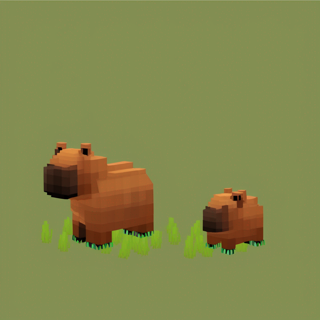}
\vspace{-3mm}
\end{subfigure}
\begin{subfigure}[b]{0.16\textwidth}
\includegraphics[width=\textwidth]{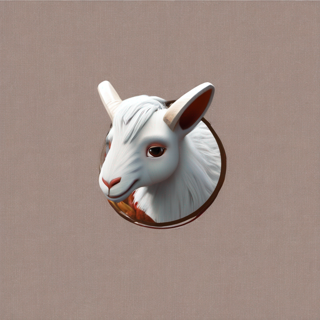}
\vspace{-3mm}
\end{subfigure}\\

\begin{subfigure}[b]{0.16\textwidth}
\includegraphics[width=\textwidth]{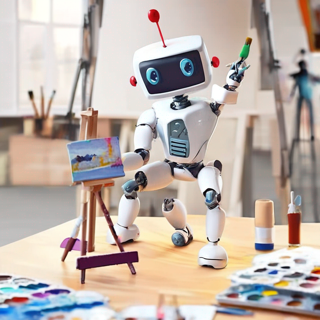}
\vspace{-3mm}
\end{subfigure}
\begin{subfigure}[b]{0.16\textwidth}
\includegraphics[width=\textwidth]{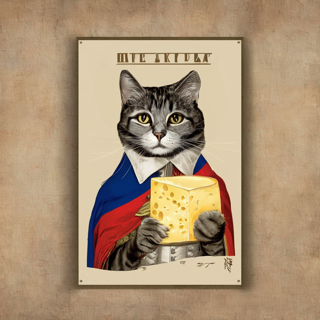}
\vspace{-3mm}
\end{subfigure}
\begin{subfigure}[b]{0.16\textwidth}
\includegraphics[width=\textwidth]{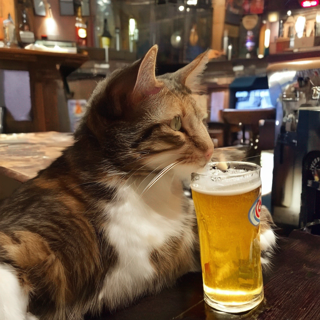}
\vspace{-3mm}
\end{subfigure}
\begin{subfigure}[b]{0.16\textwidth}
\includegraphics[width=\textwidth]{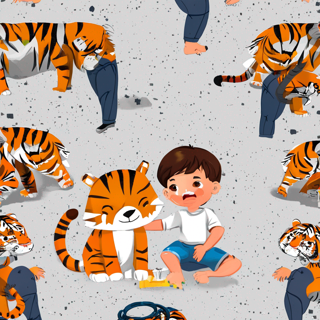}
\vspace{-3mm}
\end{subfigure}
\begin{subfigure}[b]{0.16\textwidth}
\includegraphics[width=\textwidth]{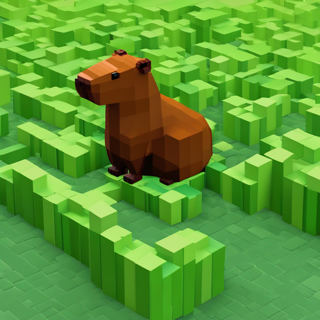}
\vspace{-3mm}
\end{subfigure}
\begin{subfigure}[b]{0.16\textwidth}
\includegraphics[width=\textwidth]{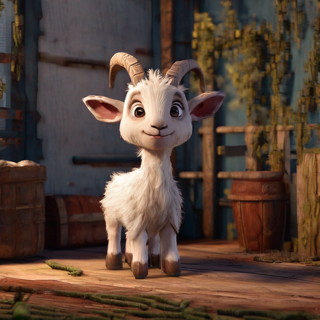}
\vspace{-3mm}
\end{subfigure}\\

\begin{subfigure}[b]{0.16\textwidth}
\includegraphics[width=\textwidth]{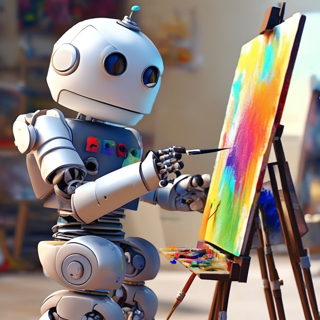}
\vspace{-3mm}
\end{subfigure}
\begin{subfigure}[b]{0.16\textwidth}
\includegraphics[width=\textwidth]{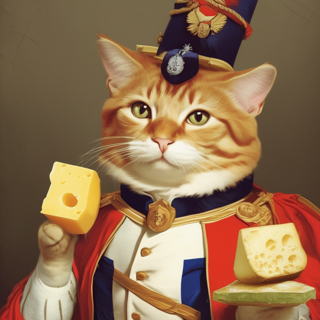}
\vspace{-3mm}
\end{subfigure}
\begin{subfigure}[b]{0.16\textwidth}
\includegraphics[width=\textwidth]{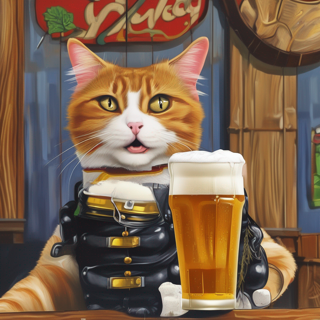}
\vspace{-3mm}
\end{subfigure}
\begin{subfigure}[b]{0.16\textwidth}
\includegraphics[width=\textwidth]{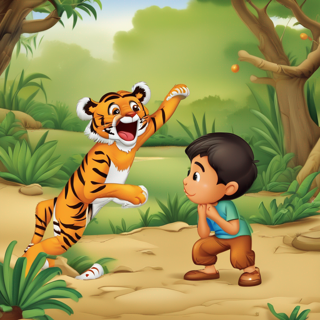}
\vspace{-3mm}
\end{subfigure}
\begin{subfigure}[b]{0.16\textwidth}
\includegraphics[width=\textwidth]{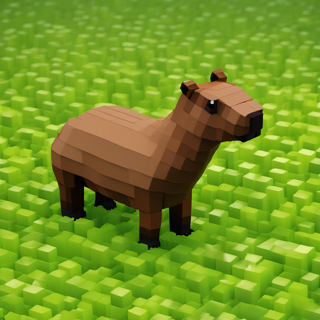}
\vspace{-3mm}
\end{subfigure}
\begin{subfigure}[b]{0.16\textwidth}
\includegraphics[width=\textwidth]{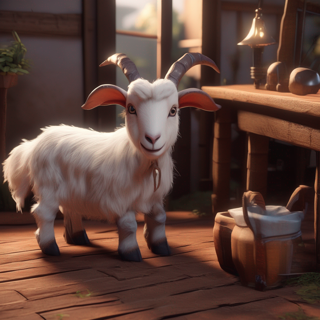}
\vspace{-3mm}
\end{subfigure}
\vspace{-3mm}
\caption{\small{\textbf{Illustrating the impact of incorporating SDXL-generated images in the design-to-scene fine-tuning process.} Displayed in sequence are the images generated by: the original SDXL on the first row, Glyph-SDXL on the second row, Glyph-SDXL-Scene fine-tuned without SDXL-generated images on the third row, and finally, Glyph-SDXL-Scene utilizing SDXL-generated images on the last row.}}
\label{fig:sdxl_cmp}
\vspace{2mm}
\end{minipage}

\begin{minipage}[t]{1\linewidth}
\centering
\begin{subfigure}[b]{0.16\textwidth}
\includegraphics[width=\textwidth]{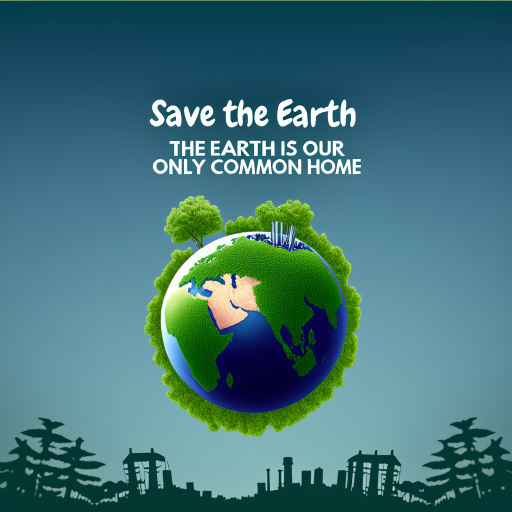}
\vspace{-3mm}
\end{subfigure}
\begin{subfigure}[b]{0.16\textwidth}
\includegraphics[width=\textwidth]{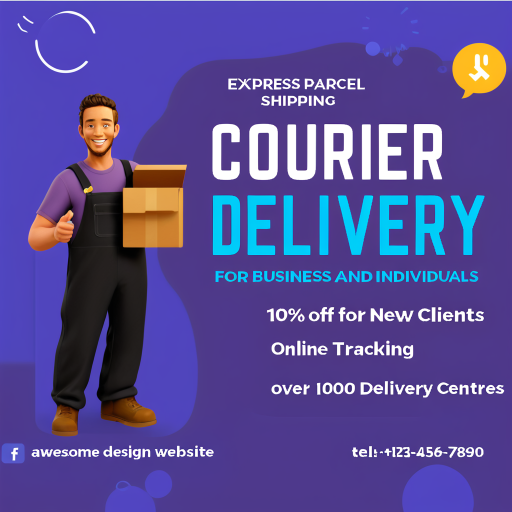}
\vspace{-3mm}
\end{subfigure}
\begin{subfigure}[b]{0.16\textwidth}
\includegraphics[width=\textwidth]{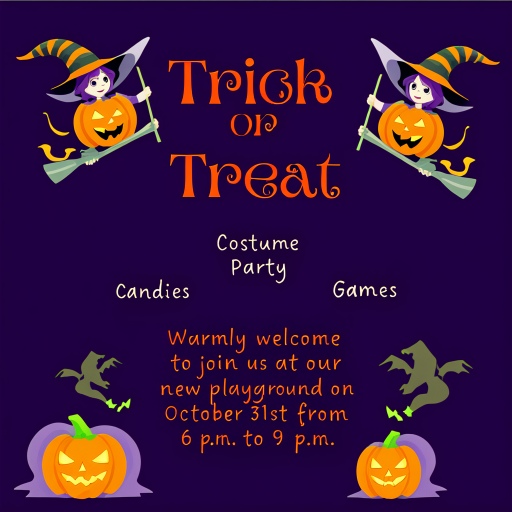}
\vspace{-3mm}
\end{subfigure}
\begin{subfigure}[b]{0.16\textwidth}
\includegraphics[width=\textwidth]{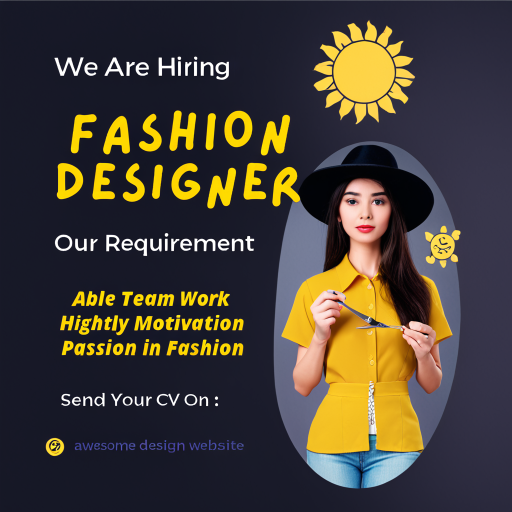}
\vspace{-3mm}
\end{subfigure}
\begin{subfigure}[b]{0.16\textwidth}
\includegraphics[width=\textwidth]{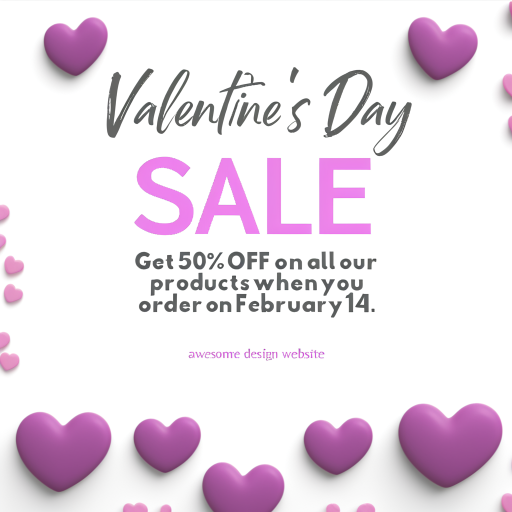}
\vspace{-3mm}
\end{subfigure}
\begin{subfigure}[b]{0.16\textwidth}
\includegraphics[width=\textwidth]{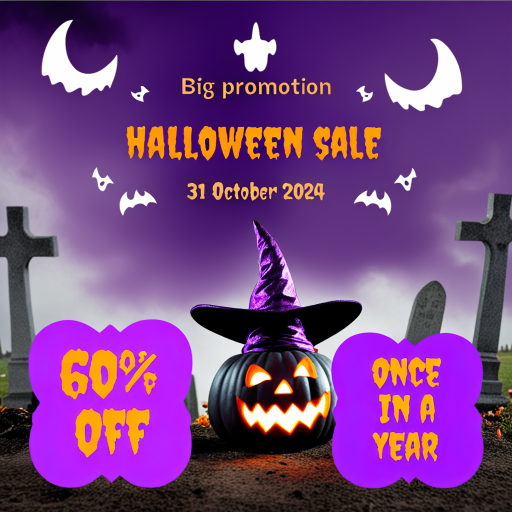}
\vspace{-3mm}
\end{subfigure}\\

\begin{subfigure}[b]{0.16\textwidth}
\includegraphics[width=\textwidth]{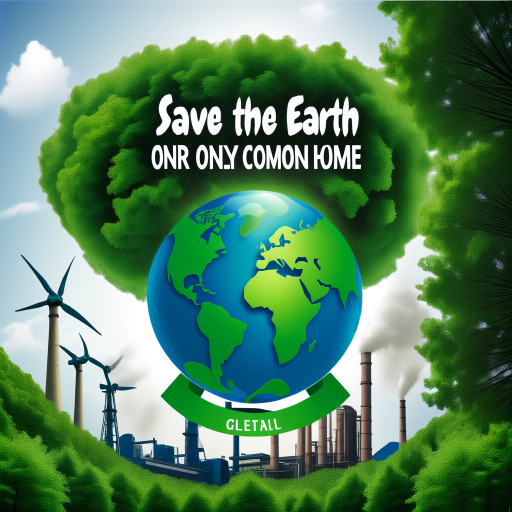}
\vspace{-3mm}
\end{subfigure}
\begin{subfigure}[b]{0.16\textwidth}
\includegraphics[width=\textwidth]{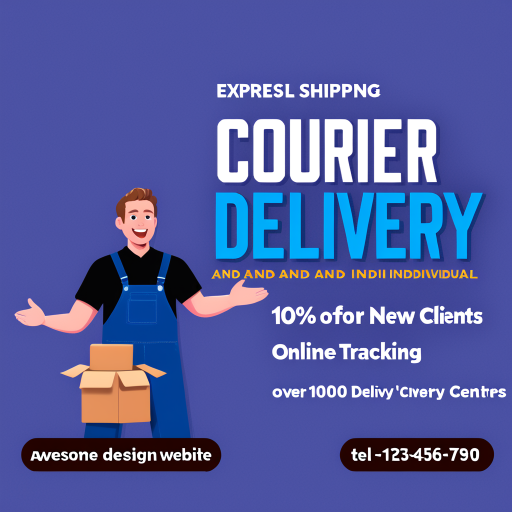}
\vspace{-3mm}
\end{subfigure}
\begin{subfigure}[b]{0.16\textwidth}
\includegraphics[width=\textwidth]{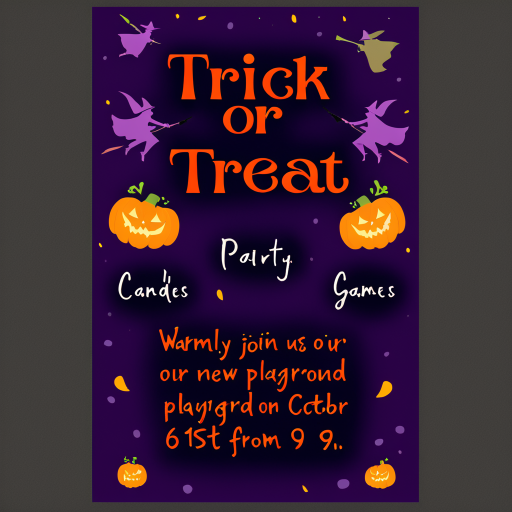}
\vspace{-3mm}
\end{subfigure}
\begin{subfigure}[b]{0.16\textwidth}
\includegraphics[width=\textwidth]{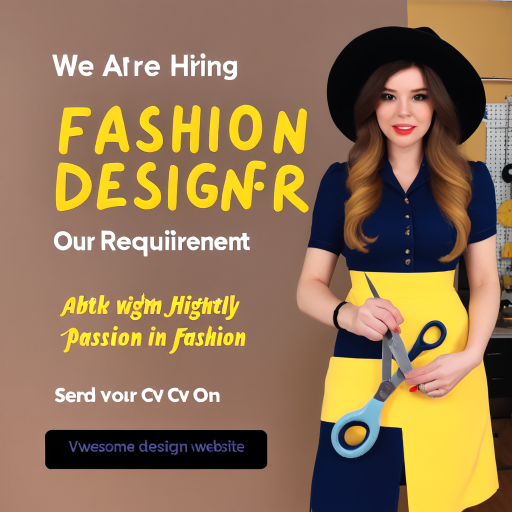}
\vspace{-3mm}
\end{subfigure}
\begin{subfigure}[b]{0.16\textwidth}
\includegraphics[width=\textwidth]{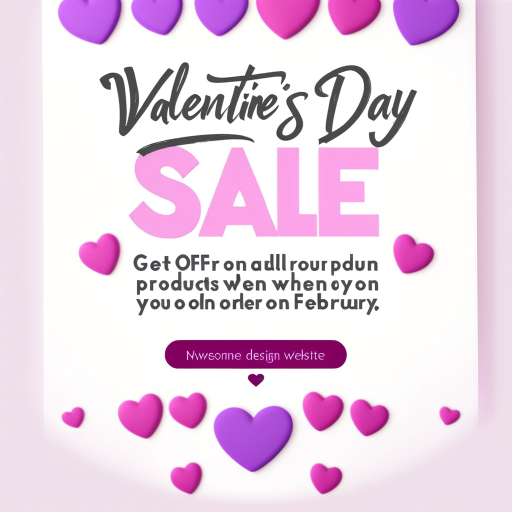}
\vspace{-3mm}
\end{subfigure}
\begin{subfigure}[b]{0.16\textwidth}
\includegraphics[width=\textwidth]{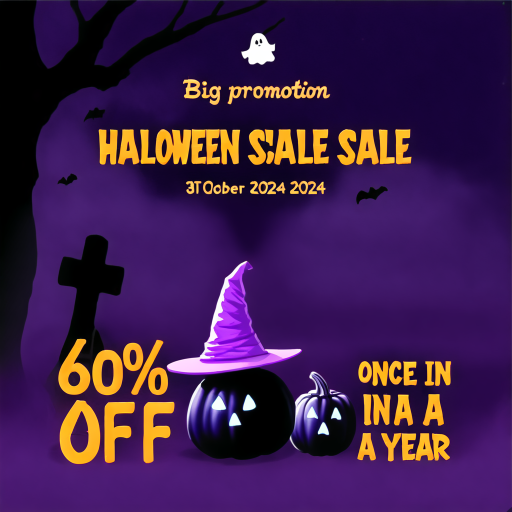}
\vspace{-3mm}
\end{subfigure}\\

\begin{subfigure}[b]{0.16\textwidth}
\includegraphics[width=\textwidth]{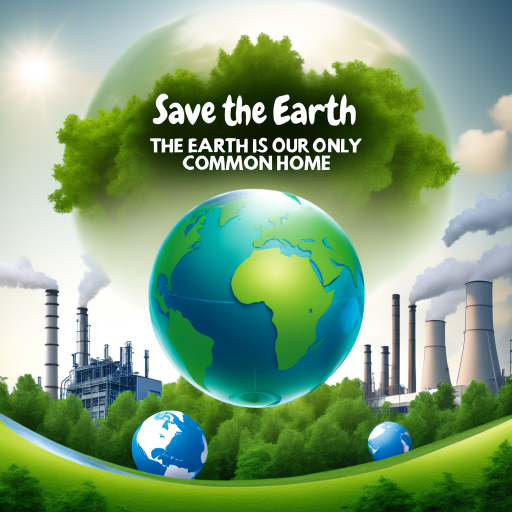}
\vspace{-3mm}
\end{subfigure}
\begin{subfigure}[b]{0.16\textwidth}
\includegraphics[width=\textwidth]{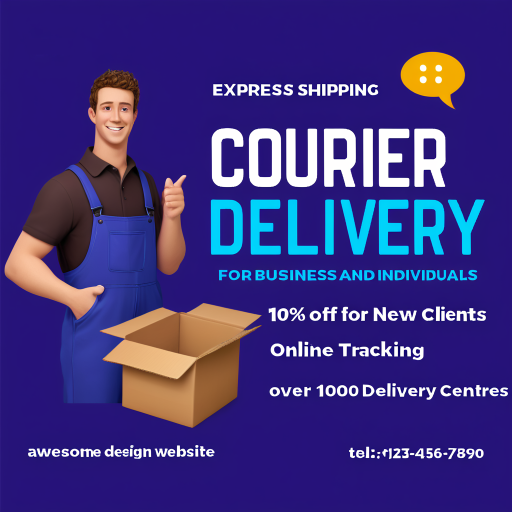}
\vspace{-3mm}
\end{subfigure}
\begin{subfigure}[b]{0.16\textwidth}
\includegraphics[width=\textwidth]{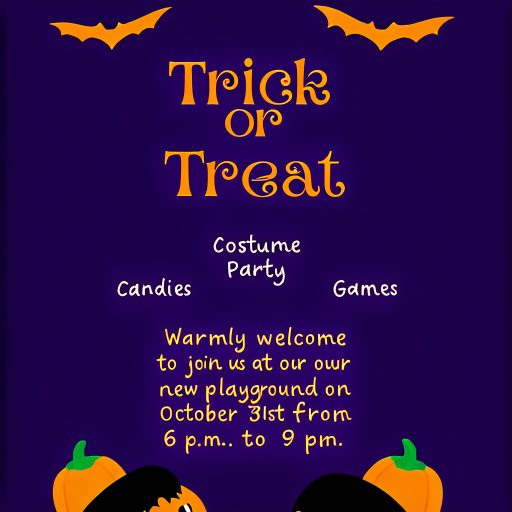}
\vspace{-3mm}
\end{subfigure}
\begin{subfigure}[b]{0.16\textwidth}
\includegraphics[width=\textwidth]{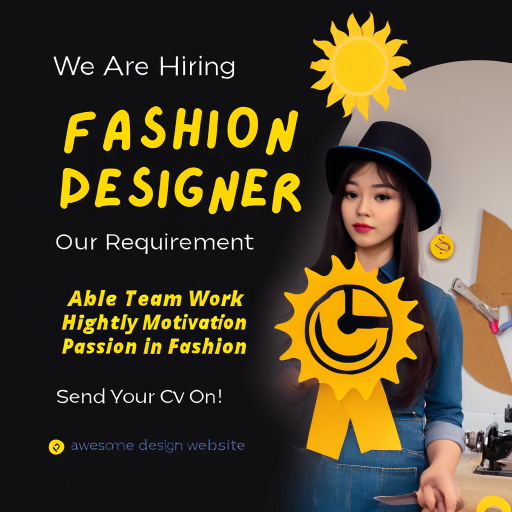}
\vspace{-3mm}
\end{subfigure}
\begin{subfigure}[b]{0.16\textwidth}
\includegraphics[width=\textwidth]{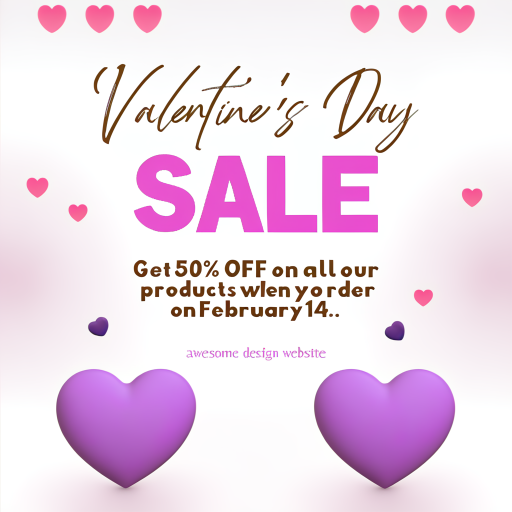}
\vspace{-3mm}
\end{subfigure}
\begin{subfigure}[b]{0.16\textwidth}
\includegraphics[width=\textwidth]{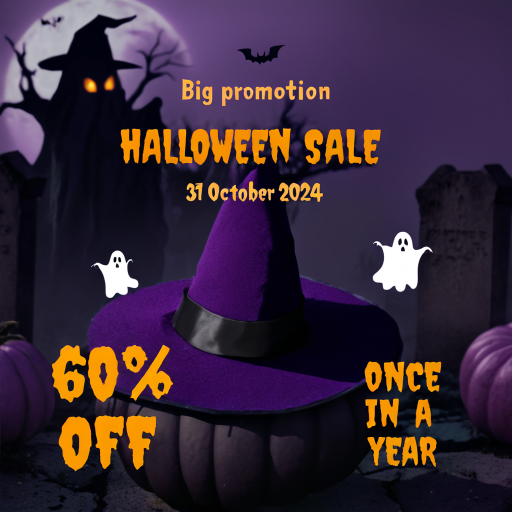}
\vspace{-3mm}
\end{subfigure}
\vspace{-3mm}
\caption{\small{\textbf{Illustrating the impact of incorporating graphic design images in the design-to-scene fine-tuning process.} Displayed in sequence are the images generated by: Glyph-SDXL on the first row, Glyph-SDXL-Scene fine-tuned without graphic design images on the second row, and finally, Glyph-SDXL-Scene utilizing graphic design images on the last row.}}
\label{fig:design_cmp}
\end{minipage}
\end{figure*}

\begin{figure*}[t]
\begin{minipage}[t]{1\linewidth}
\centering
\begin{subfigure}[b]{0.16\textwidth}
\includegraphics[width=\textwidth]{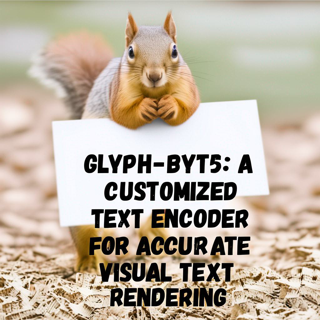}
\vspace{-3mm}
\end{subfigure}
\begin{subfigure}[b]{0.16\textwidth}
\includegraphics[width=\textwidth]{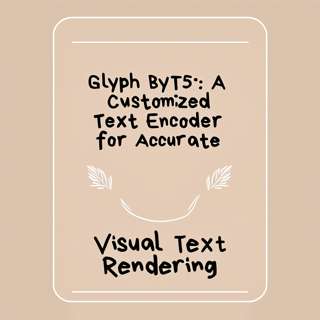}
\vspace{-3mm}
\end{subfigure}
\begin{subfigure}[b]{0.16\textwidth}
\includegraphics[width=\textwidth]{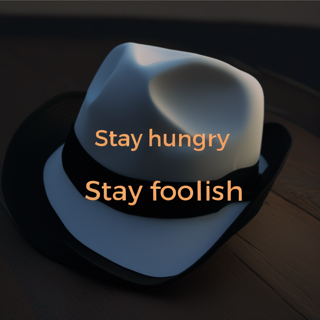}
\vspace{-3mm}
\end{subfigure}
\begin{subfigure}[b]{0.16\textwidth}
\includegraphics[width=\textwidth]{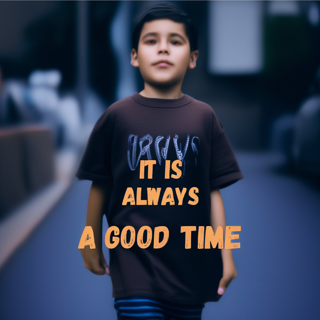}
\vspace{-3mm}
\end{subfigure}
\begin{subfigure}[b]{0.16\textwidth}
\includegraphics[width=\textwidth]{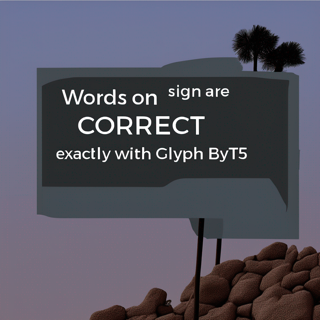}
\vspace{-3mm}
\end{subfigure}
\begin{subfigure}[b]{0.16\textwidth}
\includegraphics[width=\textwidth]{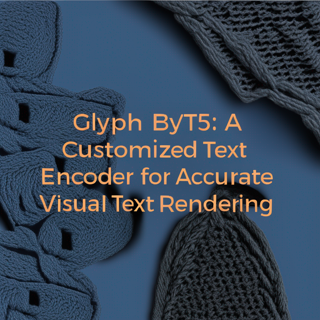}
\vspace{-3mm}
\end{subfigure}\\

\begin{subfigure}[b]{0.16\textwidth}
\includegraphics[width=\textwidth]{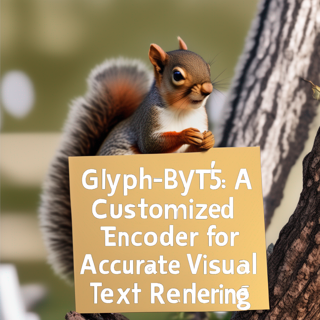}
\vspace{-3mm}
\end{subfigure}
\begin{subfigure}[b]{0.16\textwidth}
\includegraphics[width=\textwidth]{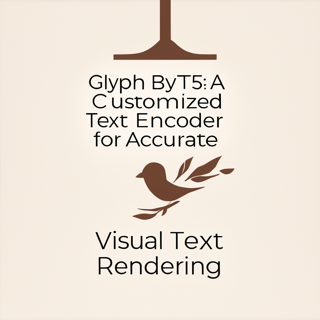}
\vspace{-3mm}
\end{subfigure}
\begin{subfigure}[b]{0.16\textwidth}
\includegraphics[width=\textwidth]{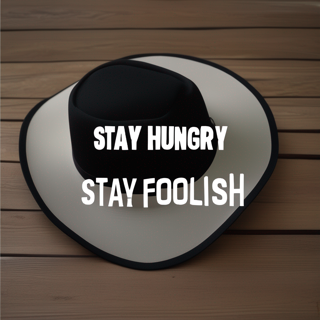}
\vspace{-3mm}
\end{subfigure}
\begin{subfigure}[b]{0.16\textwidth}
\includegraphics[width=\textwidth]{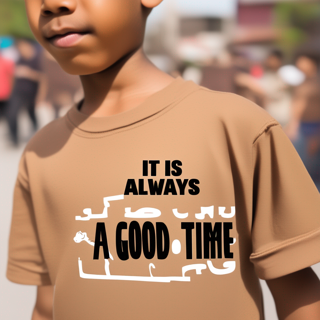}
\vspace{-3mm}
\end{subfigure}
\begin{subfigure}[b]{0.16\textwidth}
\includegraphics[width=\textwidth]{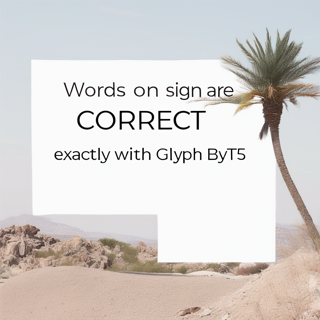}
\vspace{-3mm}
\end{subfigure}
\begin{subfigure}[b]{0.16\textwidth}
\includegraphics[width=\textwidth]{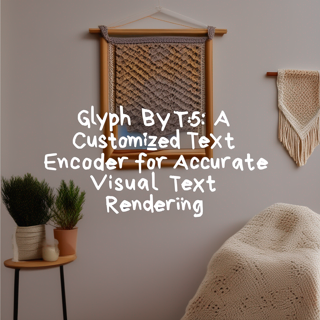}
\vspace{-3mm}
\end{subfigure}\\

\begin{subfigure}[b]{0.16\textwidth}
\includegraphics[width=\textwidth]{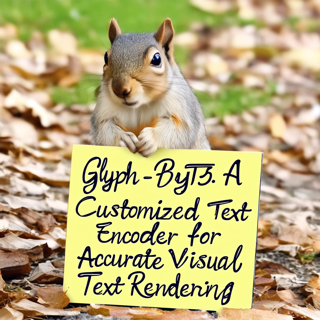}
\vspace{-3mm}
\end{subfigure}
\begin{subfigure}[b]{0.16\textwidth}
\includegraphics[width=\textwidth]{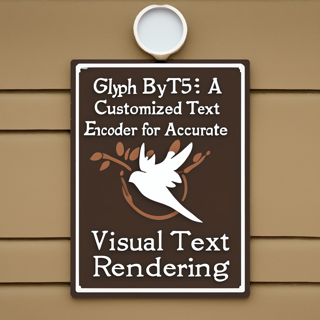}
\vspace{-3mm}
\end{subfigure}
\begin{subfigure}[b]{0.16\textwidth}
\includegraphics[width=\textwidth]{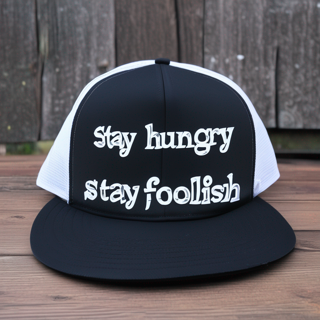}
\vspace{-3mm}
\end{subfigure}
\begin{subfigure}[b]{0.16\textwidth}
\includegraphics[width=\textwidth]{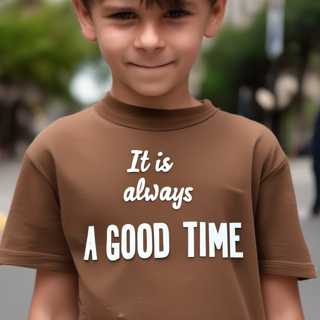}
\vspace{-3mm}
\end{subfigure}
\begin{subfigure}[b]{0.16\textwidth}
\includegraphics[width=\textwidth]{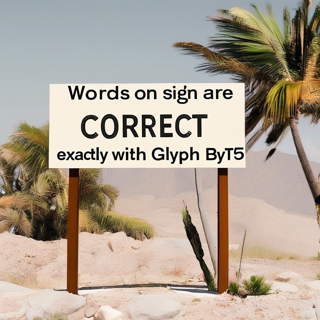}
\vspace{-3mm}
\end{subfigure}
\begin{subfigure}[b]{0.16\textwidth}
\includegraphics[width=\textwidth]{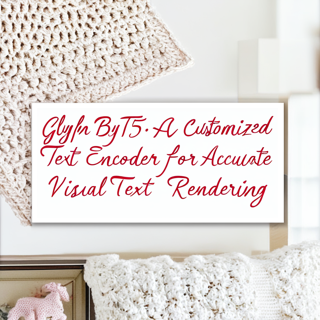}
\vspace{-3mm}
\end{subfigure}
\caption{\small{\textbf{Illustrating the impact of incorporating scene-text images in the design-to-scene fine-tuning process.} Displayed in sequence are the images generated by: Glyph-SDXL on the first row, Glyph-SDXL-Scene fine-tuned without TextSeg images on the second row, and finally, Glyph-SDXL-Scene utilizing TextSeg images on the last row.}}
\label{fig:scene_cmp}
\end{minipage}
\end{figure*}

We examine the impact of employing three types of high-quality data, as shown in Figure~\ref{fig:sdxl_cmp}, Figure~\ref{fig:design_cmp}, and Figure~\ref{fig:scene_cmp}. 

First, Figure~\ref{fig:sdxl_cmp} presents a comparison that confirms the importance of fine-tuning Glyph-SDXL with synthetic images created by SDXL. This process significantly \emph{mitigates the 'language drift' phenomenon observed when fine-tuning solely with graphic design data.} Furthermore, our analysis reveals that fine-tuning with a combined dataset of TextSeg and graphic design images is beneficial, even in the absence of images generated by SDXL.

Then, we illustrate the impact of incorporating graphic design images in Figure~\ref{fig:design_cmp}, \emph{highlighting their importance for ensuring accurate font rendering}. This is evident from comparing the results displayed in the second and third rows.
Last, in Figure~\ref{fig:scene_cmp}, we confirm the critical role of incorporating the TextSeg dataset in \emph{generating scene text that seamlessly blends with background objects}. For instance, without the TextSeg dataset, the placement of scene text often appears illogical, occasionally even situated outside intended boards or sign placeholders.

\subsection*{E. More ablation experiments}

\vspace{1mm}
\noindent\textbf{Comparison with ControlNet-style Model} To validate the effectiveness of our customized text encoder against other forms of feature guidance, we fine-tune a representative ControlNet-style SDXL model, i.e., GlyphControl~\cite{yang2023glyphcontrol}, and report results in Table \ref{tab:compare_controlnet}.
We can see that our Glyph-SDXL significantly performs better, and the performance gap increases with a larger number of characters. This indicates that the ControlNet-style model, conditioned on rendered text masks, suffers from fundamental limitations when handling paragraph-level dense visual text.

\vspace{1mm}
\noindent\textbf{Comparison with Aligning Directly with SDXL Latent Space} We compare our approach with directly aligning with SDXL's latent space, i.e. using SDXL-VAE as the visual encoder. 
As indicated in Table~\ref{tab:compare_vae}, SDXL-VAE is weak at extracting reliable visual text features and lags behind DINOv2 significantly.

\vspace{1mm}
\noindent\textbf{Ablation on other Text Encoder Fusion Scheme} One might question the effectiveness of a basic approach that concatenates the text embeddings from different text encoders. The comparison results are detailed in Table \ref{tab:effect_regionwise}.
Empirically, we find that this baseline underperforms significantly due to the substantial discrepancies among the text encoders.

\subsection*{F. Font Type Blending}

\begin{figure*}[t]
\begin{minipage}[t]{1\linewidth}
\centering
\begin{subfigure}[b]{0.19\textwidth}
{\includegraphics[width=\textwidth]{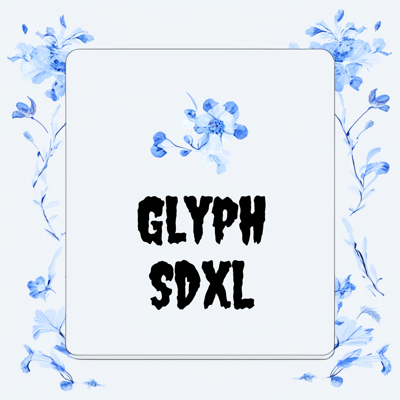}}
\vspace{-3mm}
\end{subfigure}
\begin{subfigure}[b]{0.19\textwidth}
{\includegraphics[width=\textwidth]{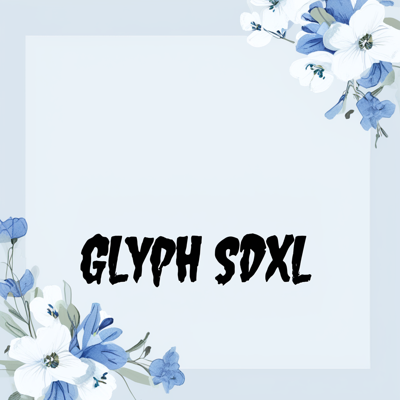}}
\vspace{-3mm}
\end{subfigure}
\begin{subfigure}[b]{0.19\textwidth}
{\includegraphics[width=\textwidth]{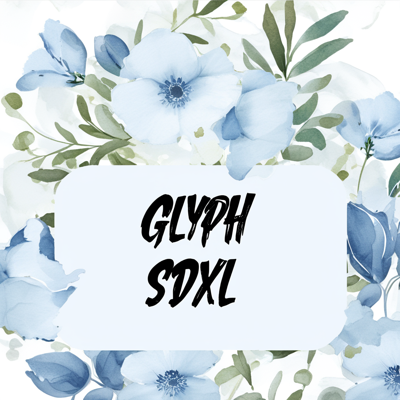}}
\vspace{-3mm}
\end{subfigure}
\begin{subfigure}[b]{0.19\textwidth}
{\includegraphics[width=\textwidth]{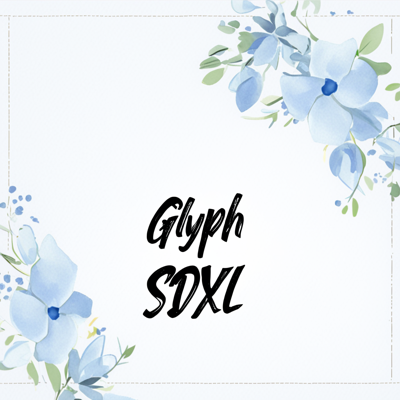}}
\vspace{-3mm}
\end{subfigure}
\begin{subfigure}[b]{0.19\textwidth}
\includegraphics[width=\textwidth]{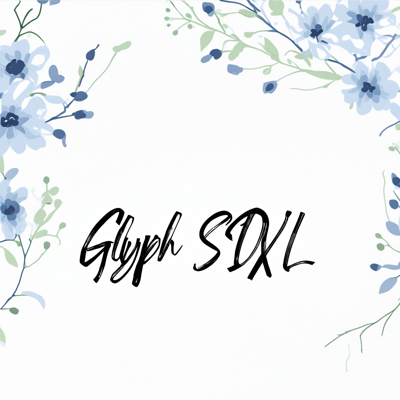}
\vspace{-3mm}
\end{subfigure}
\caption{\small{Illustrating the effect of font type blending.}}
\label{fig:font_type_blending}
\end{minipage}
\end{figure*}

We illustrate the effect of blending different font types to create new unseen font types in order to demonstrate the extrapolation abilities of our Glyph-SDXL model. As illustrated in Figure \ref{fig:font_type_blending}, we interpolate the embeddings of the italic \textit{Brightwall-Italic} font type with  the distinctive \textit{Creepster-Regular} font type to create a new blended font type that is italic and contains the special effect of \textit{Creepster-Regular}.

\subsection*{G. Detailed Prompt List}

We illustrate the detailed prompts for generated images shown in Figure 1 and Figure 5 in Table~\ref{tab:prompt_list}.

\begin{table*}[htbp]
\vspace{-3mm}
\begin{minipage}[t]{1\linewidth}  
\centering  
\tablestyle{1pt}{1.2}  
\resizebox{1.0\linewidth}{!}  
{  
\begin{tabular}{l|>{\centering\arraybackslash}m{16cm}}  
Image & Prompt \\  
\shline  
\scriptsize{Fig 1, Row 1, Col1} & \scriptsize{Background: Cards and invitations. The image features a white card adorned with blue flowers and greenery. Tags: blue, white, modern, simple, elegant, floral, illustration, professional, aesthetic, announcement. Text: Text "It was the best of times, it was the worst of times. It was the age of wisdom, it was the age of foolishness." in $<$color-1$>$, $<$font-421$>$.} \\ \hline

\scriptsize{Fig 1, Row 1, Col2} & \scriptsize{Background: Cards and invitations. The image features a white card adorned with blue flowers and greenery. Tags: blue, white, modern, simple, elegant, floral, illustration, professional, aesthetic, announcement. Text: Text "It was the best of times, it was the worst of times. It was the age of wisdom, it was the age of foolishness." in $<$color-1$>$, $<$font-469$>$.
} \\ \hline

\scriptsize{Fig 1, Row 1, Col3} & \scriptsize{Background: Cards and invitations. The image features an endless lush green forest. Tags: elegant, illustration, professional, aesthetic, announcement. Text: Text "It was the best of times, it was the worst of times." in $<$color-36$>$, $<$font-126$>$. Text "It was the age of wisdom, it was the age of foolishness." in $<$color-19$>$, $<$font-42$>$.
} \\ \hline

\scriptsize{Fig 1, Row 1, Col4} & \scriptsize{Background: Blue Modern Stars Bookmark. The image features the stary universe with saturn, mars and other planets in aesthetic oil painting style. Tags: elegant, illustration, professional, aesthetic. Text: Text "It was the best of times, it was the worst of times." in $<$color-0$>$, $<$font-47$>$. Text "It was the age of wisdom, it was the age of foolishness." in $<$color-38$>$, $<$font-420$>$.
} \\ \hline

\scriptsize{Fig 1, Row 2, Col1} & \scriptsize{Background: Instagram Posts. The image features a stack of pancakes with syrup and strawberries on top. The pancakes are arranged in a visually appealing manner, with some pancakes placed on top of each other. The syrup is drizzled generously over the pancakes, and the strawberries are scattered around, adding a touch of color and freshness to the scene. The overall presentation of the pancakes is appetizing and inviting. Tags: brown, peach, grey, modern, minimalist, simple, colorful, illustration, Instagram post, instagram, post, national pancake day, international pancake day, happy pancake day, pancake day, pancake, sweet, cake, discount, sale. Text: Text "Get 50\% Discount for your first order" in $<$color-3$>$, $<$font-97$>$. Text "Have a try!" in $<$color-0$>$, $<$font-97$>$. Text "Celebrate with Pancakes!" in $<$color-4$>$, $<$font-97$>$.
} \\ \hline

\scriptsize{Fig 1, Row 2, Col2} & \scriptsize{Background: Cards and invitations. The image features a large gray elephant sitting in a field of flowers, holding a smaller elephant in its arms. The scene is quite serene and picturesque, with the two elephants being the main focus of the image. The field is filled with various flowers, creating a beautiful and vibrant backdrop for the elephants. Tags: Light green, orange, Illustration, watercolor, playful, Baby shower invitation, baby boy shower invitation, baby boy, welcoming baby boy, koala baby shower invitation, baby shower invitation for baby shower, baby boy invitation, background, playful baby shower card, baby shower, card, newborn, born, Baby Shirt Baby Shower Invitation. Text: Text "RSVP to +123-456-7890" in $<$color-18$>$, $<$font-100$>$. Text "John Doe" in $<$color-99$>$, $<$font-210$>$. Text "Baby Shower" in $<$color-53$>$, $<$font-210$>$. Text "Please Join Us For a" in $<$color-18$>$, $<$font-100$>$. Text "In Honoring" in $<$color-18$>$, $<$font-100$>$. Text "01 July, 2024 | 00:00 PM Grand Central Hotel" in $<$color-18$>$, $<$font-100$>$.
} \\ \hline

\scriptsize{Fig 1, Row 2, Col3} & \scriptsize{Background: Flyers. The image features a purple background with a witch flying on a broomstick, surrounded by several pumpkins. The pumpkins are scattered throughout the scene, with some positioned closer to the witch and others further away. The combination of the purple background, the witch, and the pumpkins creates a festive and spooky atmosphere. Tags: purple, orange, colorful, illustration, creative, fun, dark, bold, playful, cute, cartoon, flyer, halloween, trick or treat, costume, party, spooky, pumpkin, trick, event. Text: Text "Games" in $<$color-27$>$, $<$font-197$>$. Text "Costume Party" in $<$color-27$>$, $<$font-197$>$. Text "Candies" in $<$color-27$>$, $<$font-197$>$. Text "Warmly welcome to join us at our new playground on October 31st from 6 p.m. to 9 p.m." in $<$color-51$>$, $<$font-197$>$. Text "Treat" in $<$color-51$>$, $<$font-371$>$. Text "or" in $<$color-51$>$, $<$font-371$>$. Text "Trick" in $<$color-51$>$, $<$font-371$>$.
} \\ \hline

\scriptsize{Fig 1, Row 2, Col4} & \scriptsize{Background: Instagram Posts. The image features a purple witch's hat on a pumpkin, which is placed in front of a graveyard. The pumpkin is positioned in the center of the scene, and the hat is slightly tilted to the left. There are three ghosts in the background, with one on the left side, one on the right side, and another one in the middle. The ghosts are positioned at different heights, with the one on the left being the tallest, the one in the middle being the shortest, and the one on the right being slightly taller than the middle ghost. Tags: purple, orange, yellow, illustration, halloween, halloween day, halloween party, happy halloween, pumpkins, trick or treats, spooky, haunted, event, party, festive, witch, monster, scary, ghost, instagram post. Text: Text "Big promotion" in $<$color-14$>$. Text "31 October 2024" in $<$color-14$>$. Text "HALLOWEEN SALE" in $<$color-57$>$, $<$font-252$>$. Text "ONCE IN A YEAR" in $<$color-57$>$, $<$font-252$>$. Text "60\% OFF" in $<$color-57$>$, $<$font-252$>$.
} \\ \hline

\scriptsize{Fig 1, Row 3, Col1} & \scriptsize{Background: A photo of a cute squirrel holding a sign, 4k, dslr. Text: Text "Glyph-ByT5: A Customized Text Encoder for Accurate Visual Text Rendering".
} \\ \hline

\scriptsize{Fig 1, Row 3, Col2} & \scriptsize{Background: A man standing in the midst of a vibrant sunflower field with a mountain range in the background under a blue sky, holding a sign that reads "Glyph-ByT5: A Customized Text Encoder for Accurate Visual Text Rendering" Vincent van Gogh style. Text: Text "Glyph-ByT5: A Customized Text Encoder for Accurate Visual Text Rendering".
} \\ \hline

\scriptsize{Fig 1, Row 3, Col3} & \scriptsize{Background: An intriguing scene of a blank sign standing amidst a rocky landscape, with a backdrop of a clear sky and a palm tree. Text: Text "Words on sign are". Text "CORRECT". Text "exactly with Glyph ByT5".
} \\ \hline

\scriptsize{Fig 1, Row 3, Col4} & \scriptsize{Background: The image shows a sign with a stylized design, featuring a bird and branches. The sign is hanging from a ceiling, and it appears to be located outside a building. The design is simple and modern, with a limited color palette that includes shades of brown and white. The bird and branches are depicted in a minimalist style, with clean lines and a lack of detail that gives the sign a contemporary feel. The sign is likely intended to provide information or direction to passersby, but the specific content of the sign is not visible in the image. Text "Glyph ByT5: A Customized Text Encoder for Accurate". Text "Visual Text Rendering".
} \\ \hline

\scriptsize{Fig 5, Row 1, Col1} & \scriptsize{Background: The image features a decorative frame with a floral design, showcasing a variety of flowers. The frame is adorned with a combination of pink, yellow, and white flowers, creating a visually appealing and colorful display. The flowers are arranged in a way that fills the frame, giving the impression of a vibrant and lively scene. Text: Text " "Marriage does not guarantee you will be together forever, it’s only paper. It takes love, respect, trust, understanding, friendship, and faith in your relationship to make it last."" in $<$color-10$>$, $<$font-358$>$.
} \\ \hline

\scriptsize{Fig 5, Row 1, Col2} & \scriptsize{Background: The image features a white background with a few plants and flowers scattered across it. There are three main plants in the scene, with one located on the left side, another in the middle, and the third on the right side. Additionally, there are two smaller plants in the upper part of the image. The plants are of various sizes and shapes, adding a sense of diversity to the scene. Text: Text "Give yourself the same amount of attention and warmth you selflessly give to others." in $<$color-7$>$, $<$font-196$>$.
} \\ \hline

\scriptsize{Fig 5, Row 1, Col3} & \scriptsize{Background: Instagram Posts. The image features a woman sitting in a lotus position, also known as a yoga pose, with her legs crossed and her hands resting on her knees. She is surrounded by a serene environment, with trees in the background and a sun in the sky above her. The woman appears to be meditating or practicing yoga in a peaceful outdoor setting. Tags: WHITE, BROWN, BLUE, MODERN, meditation, exercise, fitness, yoga day, poster, health, illustration, international, position, concept, relaxation, yoga, woman. Text: Text "" let's get moving for a healthy body!"" in $<$color-2$>$, $<$font-30$>$. Text "BEAUTY OF YOGA" in $<$color-2$>$, $<$font-30$>$. Text "Experience" in $<$color-2$>$, $<$font-500$>$.
} \\ \hline

\scriptsize{Fig 5, Row 1, Col4} & \scriptsize{Background: The image features a white background with a circular frame made of colored pencils. The frame is filled with a variety of colored pencils, creating a visually appealing and artistic design. The pencils are arranged in different positions, with some overlapping and others standing alone. The combination of colors and the circular shape of the frame make the image a unique and creative piece of art. Text: Text "I TOTALLY REMEMBERED YOUR BIRTHDAY DEAR FRIEND!" in $<$color-14$>$, $<$font-101$>$.
} \\ \hline

\scriptsize{Fig 5, Row 1, Col5} & \scriptsize{Background: Facebook Post. The image features a man wearing a black shirt and blue overalls, holding a brown box. He is standing in front of a blue background, which has a few speech bubbles scattered around. The man appears to be smiling, possibly indicating that he is happy or excited about the box he is holding. Tags: Violet, purple, illustration, illustrated, corporate, professional, Courier, delivery, parcel, package, fast, free, express, shipping, vehicle, transportation, pickup, centre, man, character. Text: Text "EXPRESS PARCEL SHIPPING" in $<$color-0$>$, $<$font-4$>$. Text "awesome design website" in $<$color-0$>$, $<$font-4$>$. Text "tel.: +123-456-7890" in $<$color-0$>$, $<$font-4$>$. Text "over 1000 Delivery Centres" in $<$color-0$>$, $<$font-4$>$. Text "Online Tracking" in $<$color-0$>$, $<$font-4$>$. Text "10\% off for New Clients" in $<$color-0$>$, $<$font-4$>$. Text "FOR BUSINESS AND INDIVIDUALS" in $<$color-123$>$, $<$font-4$>$. Text "DELIVERY" in $<$color-123$>$, $<$font-54$>$. Text "COURIER" in $<$color-0$>$, $<$font-54$>$.
} \\ 
\end{tabular}  
}
\caption{  
\footnotesize{Detailed prompt for generated images in Figure 1 and Figure 5.}}
\label{tab:prompt_list}
\vspace{3mm}
\end{minipage}
\end{table*}

\subsection*{H. Typography Layout Planning with GPT-4}

To reduce reliance on manually provided typography layouts, such as target text boxes, we utilize the visual planning capabilities of GPT-4 to automate layout generation and subsequently display images based on these layouts. Additionally, we employ the layout prediction capabilities of TextDiffuser-2's LLM to determine target text boxes. Following these predictions, we implement our Glyph-SDXL model to generate visual text images, as shown in Figure \ref{fig:gpt4}. The results indicate that GPT-4's layout capabilities are significantly more reliable than those of TextDiffuser-2's layout LLM.

\begin{figure*}[t]
\begin{minipage}[t]{1\linewidth}
\centering
\begin{subfigure}[b]{0.16\textwidth}
\includegraphics[width=\textwidth]{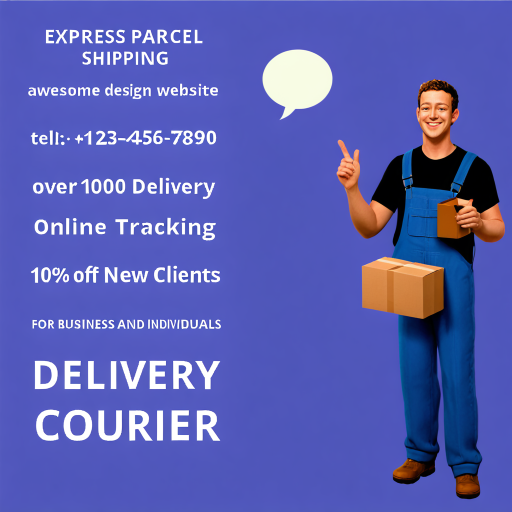}
\caption*{\scriptsize{GPT4}}
\vspace{-3mm}
\end{subfigure}
\begin{subfigure}[b]{0.16\textwidth}
{\includegraphics[width=\textwidth]{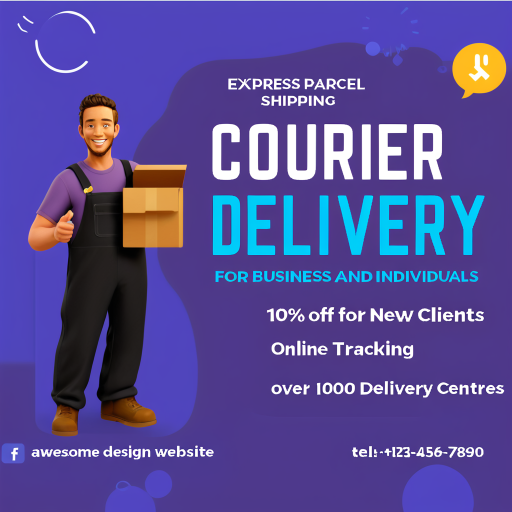}}
\caption*{\scriptsize{Human}}
\vspace{-3mm}
\end{subfigure}
\begin{subfigure}[b]{0.16\textwidth}
{\includegraphics[width=\textwidth]{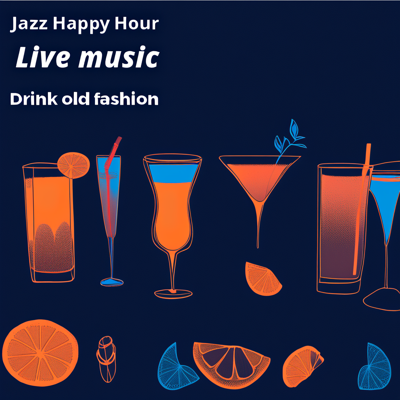}}
\caption*{\scriptsize{GPT4}}
\vspace{-3mm}
\end{subfigure}
\begin{subfigure}[b]{0.16\textwidth}
{\includegraphics[width=\textwidth]{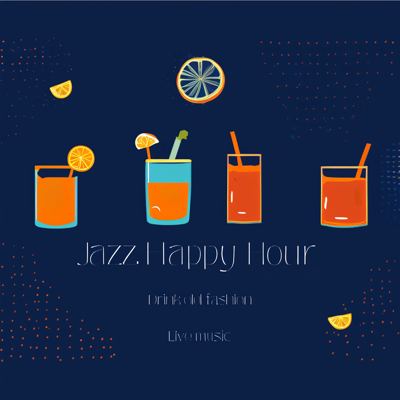}}
\caption*{\scriptsize{Human}}
\vspace{-3mm}
\end{subfigure}
\begin{subfigure}[b]{0.16\textwidth}
{\includegraphics[width=\textwidth]{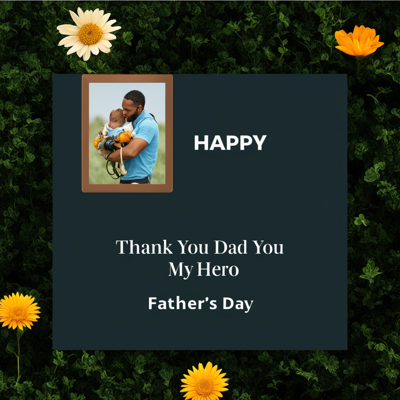}}
\caption*{\scriptsize{GPT4}}
\vspace{-3mm}
\end{subfigure}
\begin{subfigure}[b]{0.16\textwidth}
{\includegraphics[width=\textwidth]{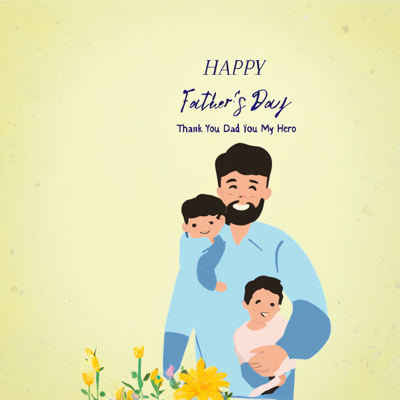}}
\caption*{\scriptsize{Human}}
\vspace{-3mm}
\end{subfigure}
\vspace{3mm}
\caption{\small{GPT-4 layout planning failure cases. Results generated using the layout predicted with GPT-and human designers are showcased in Col 1, 3, 5 and Col 2, 4, 6 respectively.}}
\label{fig:gpt4_failure_cases}
\end{minipage}
\end{figure*}

\begin{figure*}[t]
\begin{minipage}[t]{1\linewidth}
\centering
\begin{subfigure}[b]{0.19\textwidth}
\includegraphics[width=\textwidth]{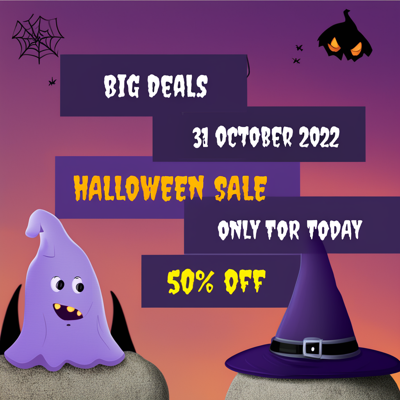}
\vspace{-3mm}
\end{subfigure}
\begin{subfigure}[b]{0.19\textwidth}
{\includegraphics[width=\textwidth]{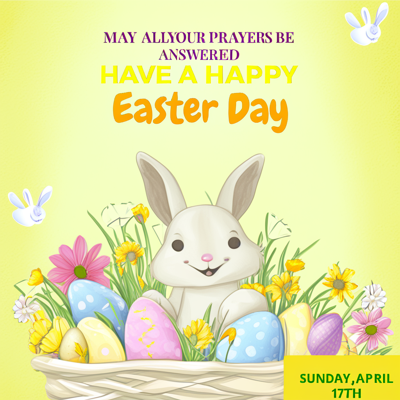}}
\vspace{-3mm}
\end{subfigure}
\begin{subfigure}[b]{0.19\textwidth}
{\includegraphics[width=\textwidth]{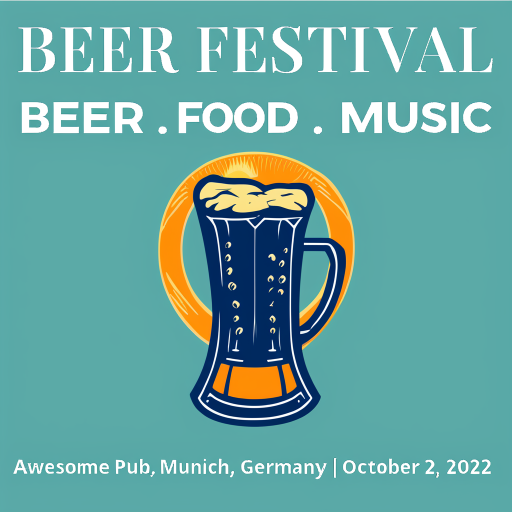}}
\vspace{-3mm}
\end{subfigure}
\begin{subfigure}[b]{0.19\textwidth}
{\includegraphics[width=\textwidth]{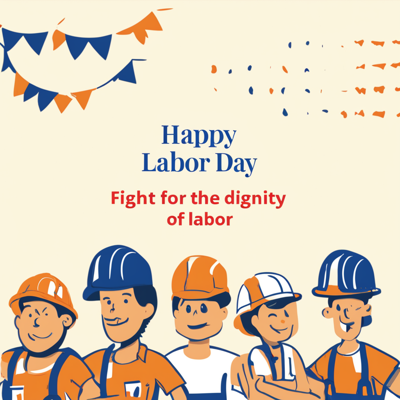}}
\vspace{-3mm}
\end{subfigure}
\begin{subfigure}[b]{0.19\textwidth}
{\includegraphics[width=\textwidth]{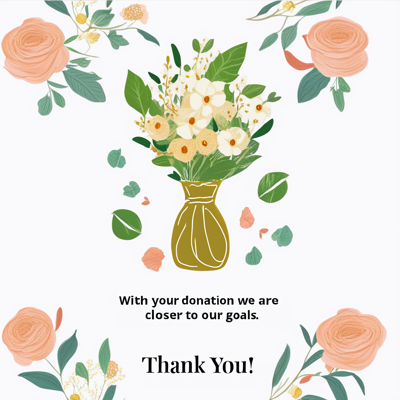}}
\vspace{-3mm}
\end{subfigure}
\begin{subfigure}[b]{0.19\textwidth}
\includegraphics[width=\textwidth]{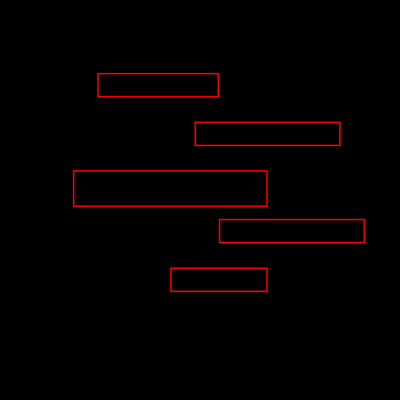}
\vspace{-3mm}
\end{subfigure}
\begin{subfigure}[b]{0.19\textwidth}
{\includegraphics[width=\textwidth]{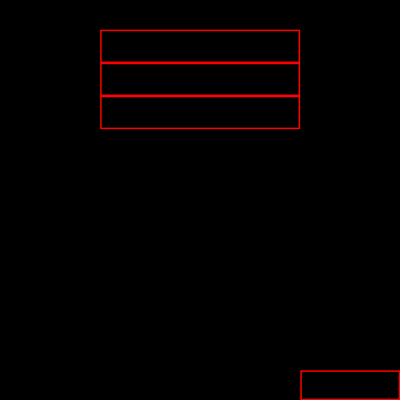}}
\vspace{-3mm}
\end{subfigure}
\begin{subfigure}[b]{0.19\textwidth}
{\includegraphics[width=\textwidth]{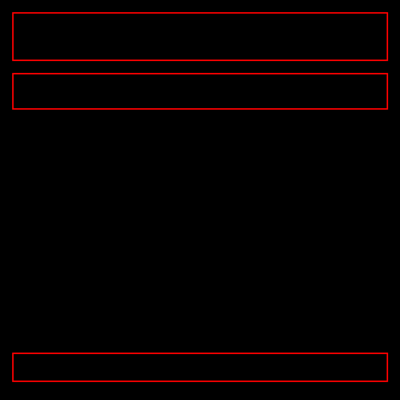}}
\vspace{-3mm}
\end{subfigure}
\begin{subfigure}[b]{0.19\textwidth}
{\includegraphics[width=\textwidth]{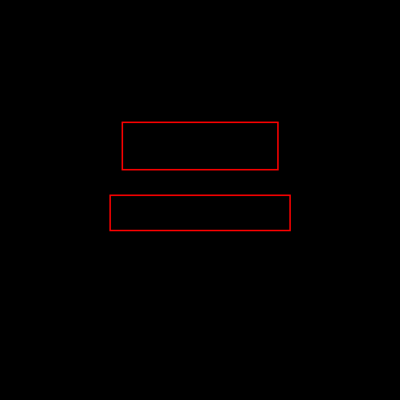}}
\vspace{-3mm}
\end{subfigure}
\begin{subfigure}[b]{0.19\textwidth}
{\includegraphics[width=\textwidth]{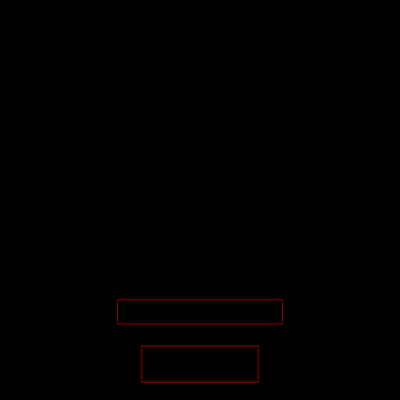}}
\vspace{-3mm}
\end{subfigure}
\begin{subfigure}[b]{0.19\textwidth}
\includegraphics[width=\textwidth]{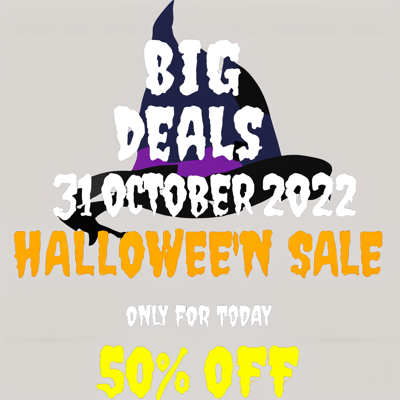}
\vspace{-3mm}
\end{subfigure}
\begin{subfigure}[b]{0.19\textwidth}
{\includegraphics[width=\textwidth]{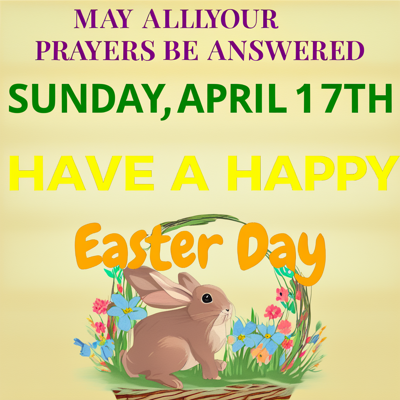}}
\vspace{-3mm}
\end{subfigure}
\begin{subfigure}[b]{0.19\textwidth}
{\includegraphics[width=\textwidth]{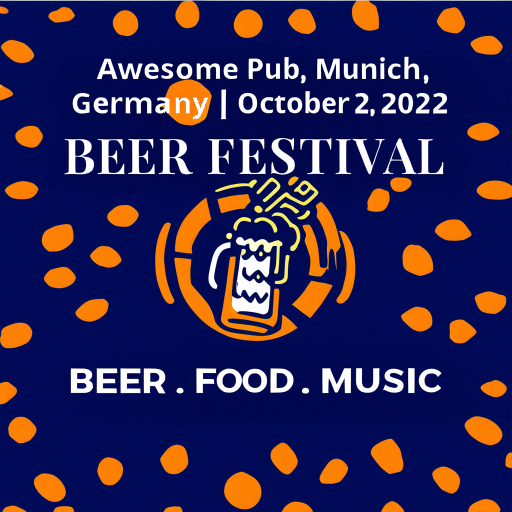}}
\vspace{-3mm}
\end{subfigure}
\begin{subfigure}[b]{0.19\textwidth}
{\includegraphics[width=\textwidth]{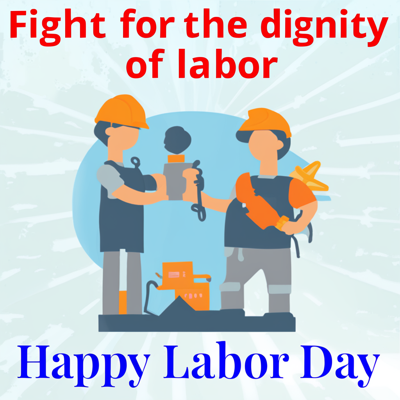}}
\vspace{-3mm}
\end{subfigure}
\begin{subfigure}[b]{0.19\textwidth}
{\includegraphics[width=\textwidth]{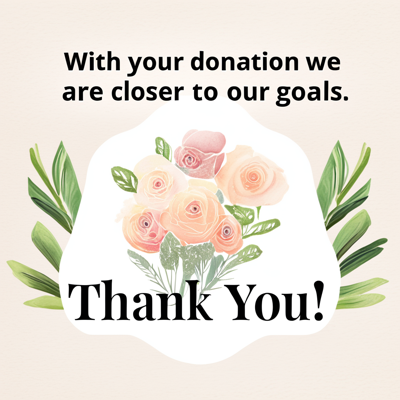}}
\vspace{-3mm}
\end{subfigure}
\begin{subfigure}[b]{0.19\textwidth}
\includegraphics[width=\textwidth]{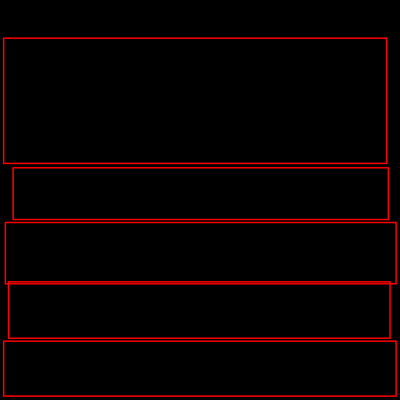}
\vspace{-3mm}
\end{subfigure}
\begin{subfigure}[b]{0.19\textwidth}
{\includegraphics[width=\textwidth]{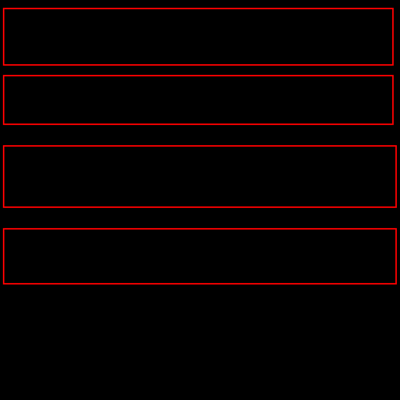}}
\vspace{-3mm}
\end{subfigure}
\begin{subfigure}[b]{0.19\textwidth}
{\includegraphics[width=\textwidth]{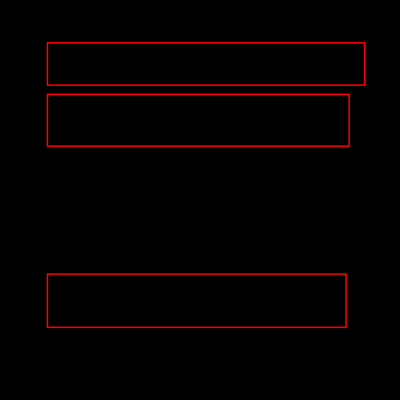}}
\vspace{-3mm}
\end{subfigure}
\begin{subfigure}[b]{0.19\textwidth}
{\includegraphics[width=\textwidth]{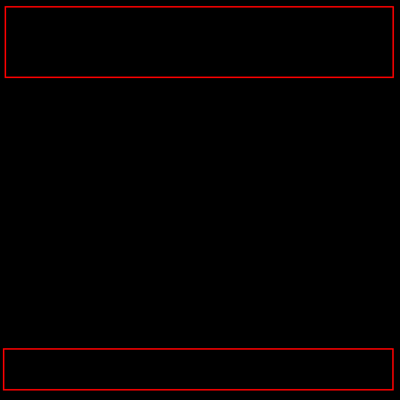}}
\vspace{-3mm}
\end{subfigure}
\begin{subfigure}[b]{0.19\textwidth}
{\includegraphics[width=\textwidth]{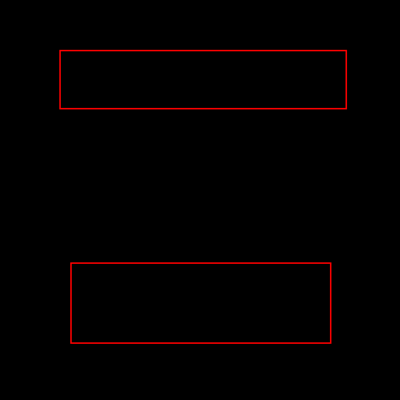}}
\vspace{-3mm}
\end{subfigure}
\caption{\small{Illustrating images and layout examples generated by integrating GPT-4 (Row 1 \& 2) and TextDiffuser-2 (Row 3 \& 4) (as TextDiffuser-2 is not trained for planning font type and colors, we use the styles recommended by GPT-4) as a planning prior, respectively.}}
\label{fig:gpt4}
\end{minipage}
\end{figure*}

Moreover, we present several typical failure cases encountered with GPT-4 in Figure~\ref{fig:gpt4_failure_cases}. Notably, GPT-4 tends to uniformly distribute all text boxes within the same column (Columns 1 \& 2), cluster text boxes into one corner (Columns 3 \& 4), or overlook layout constraints implied by text semantics, such as placing "Happy" and "Father" together (Columns 5 \& 6).

\subsection*{I. Visual Quality}

To assess the visual quality, we report the FID scores of Glyph-SDXL on two benchmarks: the COCO benchmark, consisting of 5K text-image pairs, and the \textsc{VisualParagraphy} benchmark, consisting of 1K design-oriented text-image pairs that contain visual text in each image. The FID scores are $28.06$ and $47.96$, respectively.
While our method's FID score on COCO images is slightly higher than the $26.48$ achieved by SDXL, it is important to note that the FID metric inherently favors natural images over design images. This is due to the FID's reliance on Inception-v3, which is pre-trained on the natural image dataset ImageNet.

\subsection*{J. Statistics over multiple runs}

We report the mean+std of OCR word-level precision over five runs in Table \ref{tab:ocr_mean_std}. Additionally, we also define a successful sample as an image with word-level precision greater than a threshold $\theta$. We report the mean+std of the success rate over different thresholds in Table~\ref{tab:success_mean_std}.
The $23\%$ success rate noted at the bottom right of the table indicates that on average, one out of every four attempts will result in a $100\%$ accurate glyph generation, even for texts exceeding 100 characters (equivalent to 20 English words). In comparison, \dalle fails to produce any fully correct glyph images on the same setting. Furthermore, \dalle's success rates at $\theta=100\%$ are $20.75\%$, $19.25\%$, and $0\%$ for the three shorter text lengths.

\begin{table}[h]
\begin{minipage}[h]{\linewidth}
\centering
\tablestyle{10pt}{1}
\resizebox{\linewidth}{!}
{
\setlength{\tabcolsep}{15pt}
\begin{tabular}{l|cccc}
\multirow{2}{*}{Method} &  \multicolumn{4}{c}{Precision ($\%$)}  \\\cline{2-5}
& $\le$20 chars & $\le$20-50 chars & $\le$50-100 chars & $\ge$100 chars \\
\shline
Glyph-SDXL & $93.20 \pm 0.64 $ & $93.19 \pm 0.67$ & $91.23 \pm 0.60$ & $89.98 \pm 0.51$
\end{tabular}
}
\caption{
\footnotesize{Mean+std of word-level prediction over five runs.}}
\label{tab:ocr_mean_std}
\end{minipage}
\begin{minipage}[h]{\linewidth}
\centering
\resizebox{\linewidth}{!}
{
\setlength{\tabcolsep}{10pt}
\begin{tabular}{c|cccc}
\multirow{2}{*}{OCR Precision threshold $\theta$} &  \multicolumn{4}{c}{Success rate ($\%$)}  \\\cline{2-5}
& $\le$20 chars & $\le$20-50 chars & $\le$50-100 chars & $\ge$100 chars \\
\shline
90\% & $79.76 \pm 0.78 $ & $65.68 \pm 1.07$ & $68.64 \pm 1.33$ & $62.32 \pm 1.65$ \\
95\% & $79.76 \pm 0.78 $ & $64.96 \pm 1.32$ & $50.40 \pm 1.01$ & $47.20 \pm 1.41$ \\
100\% & $79.76 \pm 0.78 $ & $64.96 \pm 1.32$ & $45.04 \pm 1.90$ & $23.12 \pm 1.67$
\end{tabular}
}
\caption{
\footnotesize{Mean+std of success rate over five runs.}}
\label{tab:success_mean_std}
\end{minipage}
\end{table}